\documentclass{article}
\usepackage{jfrExamplee,times}

\usepackage{bigfoot}
\DeclareNewFootnote[plain]{default}

\usepackage{graphicx}
\usepackage{apalike}
\usepackage{setspace}

\usepackage{array} 
\usepackage{multirow}
\usepackage[usenames, dvipsnames]{color}

\usepackage{amssymb} 
\usepackage{amsmath} 
\usepackage{pifont} 
\usepackage{caption}
\usepackage{subcaption}
\usepackage{footnote}
\usepackage{xfrac} 
\usepackage{gensymb} 

\usepackage[title]{appendix}

\usepackage{hyperref}

\usepackage{lastpage}
\usepackage{fancyhdr}
\newcommand{\specialcell}[2][c]{%
  \begin{tabular}[#1]{@{}c@{}}#2\end{tabular}}
  
\newcommand{\revised}[1]{{#1}}

\newcommand{\overlap}{IoU}

\newcommand{\detection}{D}
\newcommand{\gtLabel}{L}

\newcommand{\ours}{MFC}
\newcommand{\oursO}{$\ours{}^{O}$}
\newcommand{\oursA}{$\ours{}^{A}$}
\newcommand{\oursC}{$\ours{}^{C}$}
\newcommand{\oursOA}{$\ours{}^{Std}$} 
\newcommand{\oursOAC}{$\ours{}^{OAC}$}

\newcommand{\RPNBF}{$RPN{+}BF^{Std}$}
\newcommand{\DetectNet}{$DetectNet^{Std}$}
\newcommand{\MSCNN}{$MSCNN^{Std}$}
\newcommand{\nApproachesText}{three} 
\newcommand{\nApproachesNum}{4} 

\newcommand{\Baseline}{Baseline}

\title{Comparing Apples and Oranges: \\Off-Road Pedestrian Detection on the NREC Agricultural Person-Detection Dataset}

\author{
Zachary Pezzementi\thanks{National Robotics Engineering Center, Carnegie Mellon University, Pittsburgh, PA 15201} \\
\texttt{pez@nrec.ri.cmu.edu} \\
\And
Trenton Tabor$^*$ \\
\texttt{ttabor@nrec.ri.cmu.edu} \\
\And
Peiyun Hu\thanks{Robotics Institute, Carnegie Mellon University, Pittsburgh, PA 15201} \\
\texttt{peiyunh@cs.cmu.edu} \\
\And
Jonathan K. Chang$^*$ \\
\texttt{jchang@nrec.ri.cmu.edu} \\
\And
Deva Ramanan$^\dagger$ \\
\texttt{deva@cs.cmu.edu} \\
\And
Carl Wellington$^*$ \\
\texttt{carl@carlwellington.com} \\
\And
Benzun P. Wisely Babu$^*$ \\
\texttt{bpwiselybabu@wpi.edu} \\
\And
Herman Herman$^*$ \\
\texttt{herman@nrec.ri.cmu.edu} \\
}

\begin{document}

\maketitle

\begin{abstract}
Person detection from vehicles has made rapid progress recently with the advent of multiple high-quality datasets of urban and highway driving, yet no large-scale benchmark is available for the same problem in off-road or agricultural environments. Here we present the NREC Agricultural Person-Detection Dataset to spur research in these environments. It consists of labeled stereo video of people in orange and apple orchards taken from two perception platforms (a tractor and a pickup truck), along with vehicle position data from RTK GPS. We define a benchmark on part of the dataset that combines a total of 76k labeled person images and 19k sampled person-free images.
The dataset highlights several key challenges of the domain, including varying environment, substantial occlusion by vegetation, people in motion and in non-standard poses, and people seen from a variety of distances; meta-data are included to allow targeted evaluation of each of these effects.
Finally, we present baseline detection performance results for \nApproachesText{} leading approaches from urban pedestrian detection and our own convolutional neural network approach that benefits from the incorporation of additional image context. We show that the success of existing approaches on urban data does not transfer directly to this domain.
\end{abstract}

\section{Introduction}

\begin{figure*}
    \centering
    \begin{subfigure}[t]{0.6\textwidth}
    \includegraphics[width=0.49\linewidth]{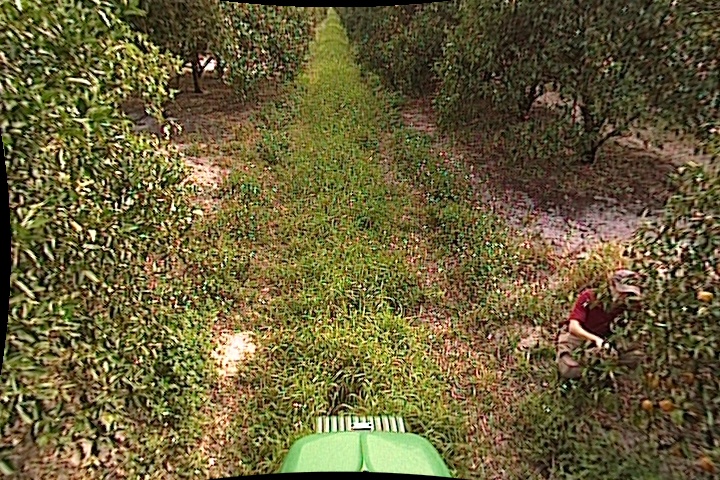}
    \includegraphics[width=0.49\linewidth]{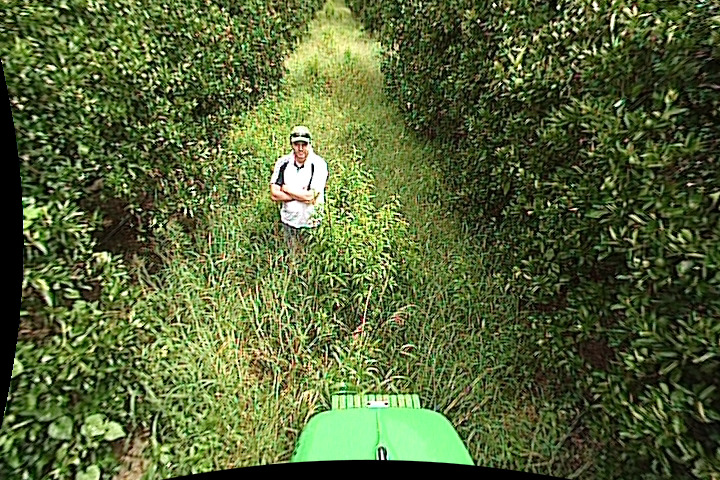}
    \caption{Orchard}
    \end{subfigure}
    \begin{subfigure}[t]{0.3\textwidth}
    \includegraphics[width=0.871\linewidth]{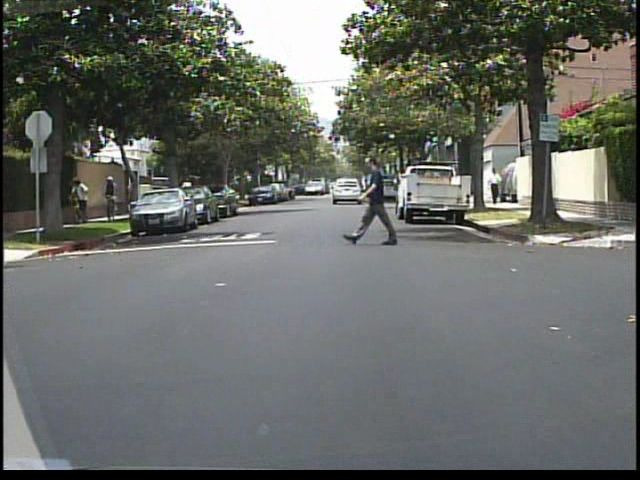}
    \caption{Urban}
    \end{subfigure}
    \caption{Example poses and occlusions commonly seen in the orchard, compared to a typical urban scene from \cite{Dollar2012PAMI:Caltech}.}
    \label{fig:poses_ag_vs_urban}
\end{figure*}

Reliable detection of people in real-world environments is an essential component to any autonomous system that needs to be able to operate with or around people, and working farms are no exception. For systems that are expected to interact with humans, the need is obvious. Even when no interaction is intended, the type of heavy machinery common in many agricultural applications presents a potential safety hazard to people around them. Even heavily automated operations typically still have human workers on site, with loose security perimeters. \revised{Although workers should generally stay clear of automated tractor operations, this does not always happen in practice, and any fully automated system needs to be able to reliably detect people.} It therefore remains an important problem for the foreseeable future.

There has been considerable work to date on detecting people from vehicles in urban environments, motivated primarily by the strong push for autonomous driving. Several good datasets have been published in this domain (discussed further in Section~\ref{sec:dataset_review}), galvanizing research progress in the field.
\revised{Although previous systems have relied heavily on laser for detecting obstacles, LIDAR alone can not distinguish between people and some plants that should be driven over. For modern commercial systems, there is a push for the use of stereo cameras where possible, to reduce cost and moving parts. Moreover, in the largest person detection benchmark that includes laser data, the current top performers do not make use of it~\cite{geiger2013vision}. We therefore focus our attention on visual methods.}
Automated driving of agricultural equipment on private land has long been commonplace, with GPS-guided driving going back two decades~\cite{OConnor96automaticsteering}, yet it has received much less attention in the perception community. 
Perhaps this is motivated by an assumption that advances in urban person detection will be directly \revised{transferable} to off-road domains. We test that hypothesis in this work.

Although specialty crops make up a small portion ($<1\%$) of the acreage devoted to agriculture in the United States, they account for $49\%$ of the hired labor on farms, $23\%$ specifically in orchards\revised{~\cite{agCensus2012}.} \revised{
A survey of tractor-related fatalities in the industry found ``farm tractors were the leading source of death within the farming industry'', and nearly one fourth of these fatalities were due to run-over incidents~\cite{myers1998statistics}.
Although national injury rate statistics are not available specifically for specialty crops, the large portion of the workforce suggests many of these injuries would be sustained there. 
Additionally, in a Korean survey of per-capita injury rates for different categories of farming, orchard farming was highest, about 50\% higher than the next-highest category~\cite{chae2014estimated}}. 
These are therefore operations where person detection is particularly relevant both for safety and for potential interaction. Orchards in particular are a venue that highlights some of the biggest challenges of off-road domains, including tall weeds and tall crops with dense foliage.
They are also operations with many opportunities for tasks that can be automated, both during the growing season and during harvest. In this work, we focus on two crops with active research in automating operations: apples~\cite{zhangintelligent,Wang_2012_7240} and oranges~\cite{subramanian2006development,Moorehead12:Orchard,freitas2012practical,hamner2012specialty}. While these two environments are much more similar to one another in structure than either is to highway or urban environments, they also have some significant differences in appearance. The two therefore provide opportunities to compare the transferability of knowledge learned between environments within the same domain versus across domains.

In this work we present an extensive dataset of videos taken from perception systems mounted on two different vehicle testbeds in two different agricultural environments. The dataset is available for public download from our project website \footnote{\href{http://www.nrec.ri.cmu.edu/projects/usdapersondetection/dataset/}{http://www.nrec.ri.cmu.edu/projects/usdapersondetection/dataset/}}. Each frame is annotated with labels of the locations of visible people. We aim to provide a benchmark that exposes the unique and challenging aspects of off-road robotics, particularly in agricultural environments, and capture people in a large variety of conditions that an automated system operating in these environments might have to deal with. These include several challenges that are seldom seen in urban data: since people in agricultural settings are often working, they may take on a great variety of poses; occlusion from vegetation is much more variable and may be extensive; highly-textured foliage, often set into motion by the wind, makes up much of the background. The result is a much larger corpus of labeled person stereo videos than any previous person detection benchmarks and, to our knowledge, the first public benchmark in an agricultural environment, or even in any off-road environment.

We also present some initial results on the benchmark, using \nApproachesText{} state-of-the-art detectors from the urban pedestrian detection benchmarks, as well as our own convolutional neural network (CNN) approach. These existing approaches' strong performance does not transfer directly to this new environment. Our approach, Multiscale Foveal Context (\ours{}), substantially out-performs other methods, particularly on small instances, despite instances of this size being much more prevalent in the urban data. These results highlight the importance of evaluating on and developing specifically for this domain.

Prior work on parts of this dataset is described in~\cite{tabor2015people}. The work described here represents an order of magnitude more labeled data, an additional environment, a new detection approach, and considerably more analysis and experimentation.

\subsection{Related Datasets}\label{sec:dataset_review}

Historically speaking, datasets have played a profound role in spurring progress in computer vision, both in terms of rigorously evaluating the state-of-the-art and in terms of providing training data for machine-learning algorithms. Indeed, a growing observation in many circles is that progress is limited not by algorithms but by data~\cite{zhu2015we,halevy2009unreasonable}.
Most recently, large-scale datasets such as ImageNet~\cite{deng2009imagenet} have enabled the success of data-hungry learning architectures, such as deep networks~\cite{krizhevsky2012imagenet}. Large-scale empirical evaluations date back at least to iconic work on image classification~\cite{nene1996columbia,fei2006one} and object detection~\cite{everingham2010pascal}. Benchmarks also play vital roles in low-level vision, including stereo~\cite{scharstein2002taxonomy,seitz2006comparison}, optical-flow~\cite{baker2011database}, and boundary detection~\cite{martin2001database}. 
Similar datasets for off-road robotic systems have been less common. The Marulan dataset~\cite{Peynot-IJRR-2010} provides multi-sensor data on several off-road conditions, but it does not have any associated benchmark, nor does it contain people.
Most related to us are benchmark datasets for pedestrian detection, discussed further below.

The importance and difficulty of dataset construction is now readily appreciated~\cite{ponce2006dataset}. One challenge is that of dataset bias~\cite{torralba2011unbiased}, a phenomena where algorithms overfit to the idiosyncrasies of a particular dataset. This is a particular concern for robotic applications, where scene-specific context (such as knowledge of the geometry of the scene) is often assumed. We take care to collect diverse data across multiple agricultural scenes, times of day, people, clothing, etc., and perform extensive {\em cross-dataset} experiments to verify the generality of our conclusions. A related concept that we explore is that of cross-dataset transfer learning~\cite{tian2015pedestrian}, where data from source domains with different biases (e.g., urban scenes) are used to potentially improve results on target domains of interest (agricultural scenes). \revised{While previous work has explored transfer of off-the-shelf deep features to a target field robotics setting~\cite{bewley2016imagenet}, we show that adaptation of deep features (through fine-tuning) significantly improves results in the target domain}.
A second difficulty, particularly with that of video, is that of annotation cost~\cite{volkmer2005web}. Most contemporary datasets make heavy use of crowdsourcing to produce ground-truth annotations~\cite{sorokin2008utility}, though quality assurance is a significant concern, particularly for structured annotations, such as object tracks in video sequences~\cite{vondrick2013efficiently}. We make use of trained, in-house annotators to ensure high-quality bounding-box annotations in stereo video streams.

\begin{table*}[th]
\centering
\begin{tabular}{c|rrrccccc}
Dataset & \specialcell{\revised{Training}\\Images} & \specialcell{Validation\\Images} & \specialcell{Test\\Images} & 
\specialcell{Includes\\Stereo} & \specialcell{Includes\\ Video} &  \specialcell{Includes\\Vehicle\\Position} & \specialcell{\#~Benchmarked\\Algorithms} & Environment \\ \hline
KITTI\footnotemark[2]{}   & 7,481\,\,\,  & --\footnotemark[6]{}        & 7,518\,\,\,  & $\checkmark$   & $\checkmark$     & -- & 70       & Urban       \\
Caltech\footnotemark[3]{} & 34,893\footnotemark[5]{}  & 784\footnotemark[5]{}      & 4,025\footnotemark[5]{}  & --              & $\checkmark$     & -- & 57       & Urban       \\
Daimler\footnotemark[4]{} & 22,789\,\,\,  & --\footnotemark[6]{}        & 21,790\,\,\,  & $\checkmark$\footnotemark[7]{}   & $\checkmark$\footnotemark[7]{} & -- & N/A      & Urban       \\
Ours        & 48,370\,\,\,  & 23,577\,\,\,     & 23,950\,\,\,  & $\checkmark$   & $\checkmark$     & $\checkmark$ & \nApproachesNum{} & \specialcell{Off-Road,\\Agricultural}
\end{tabular}
\caption{Comparison of sizes and contents of publicly-available person detection datasets.}
\label{tab:bench}
\end{table*}

With the focus on safety of autonomous vehicles around humans, our dataset is most similar to datasets created for developing and evaluating self-driving car technology. This has been a very active research area, with many datasets released over the last twenty years. In Table~\ref{tab:bench}, we summarize some characteristics of the datasets that are currently influential or similar to our own. Both Caltech~\cite{Dollar2012PAMI:Caltech} and KITTI~\cite{geiger2013vision} include online rankings, which have led to robust competition between research groups. Daimler~\cite{keller2011new} is a larger dataset and includes stereo, but their lack of an online ranking makes it hard to estimate penetration through the community. All of these datasets chose to release just two partitions, a \revised{training} and test set. By current convention, the Caltech dataset sequence 5 is commonly used for development testing, or machine learning validation. Additionally, current convention for Caltech modifies the subsampling rate\revised{$^5$}. We take some inspiration from these datasets, using tools from Caltech for labeling and evaluation and using the KITTI format for releasing additional camera and vehicle position information. We also choose to specify three full splits of our data to enable effective, independent evaluation of candidate detectors before evaluation on the test set, an uncommon practice in pedestrian detection.

\stepcounter{footnote}\footnotetext-{\protect\cite{geiger2013vision}}
\stepcounter{footnote}\footnotetext-{\protect\cite{Dollar2012PAMI:Caltech}}
\stepcounter{footnote}\footnotetext-{\protect\cite{keller2011new}}
\stepcounter{footnote}\footnotetext{Caltech includes 30 Hz video but recommends subsampling by 30x for training and evaluation. Recent work~\protect\cite{hosang2015taking,zhang2016faster} has only subsampled by 3x for training, so we report the larger number for training here.}
\stepcounter{footnote}\footnotetext{These benchmarks release a single combined set for training and validation, leaving that subdivision to users.}
\stepcounter{footnote}\footnotetext{Daimler includes stereo and video for test set only.}

\subsection{Challenges in Agricultural Environments}
\label{sec:ag-env} 

There are several important differences between off-road, agricultural environments and urban ones in the context of pedestrian detection, which have not been well captured by existing datasets.
\subsubsection{Color and Texture}
One important difference between the domains is with respect to the color composition of the background. Though there is certainly variation across factors like the exact agricultural application and time of day/year, the color of the background is much more predictable, dominated by greens and browns. The colors represented by people in these environments are just as variable as in the urban setting, though, making them often easier to distinguish on the basis of color.

The texture of regions surrounding people is also highly influenced by the ample vegetation. As shown in~\cite{tabor2015people}, the typical gradient orientation alignment is drastically different between the two environments.
 
\subsubsection{Poses}
Urban pedestrian detection benchmarks are normally dominated by views of standing and walking people. While these poses are also common in agricultural settings, so are a number of different and more challenging poses that this work aims to cover. People are expected to be doing work, so in addition to standard standing poses, they may be crouched or bent over or climbing on ladders. They may also be moving from place to place and transitioning between these poses, e.g., with limbs extended in ways that would not otherwise be commonly seen. Finally, these environments have mud, weeds hiding the placement of feet or legs, and complex geometry; humans may fall and be isolated without other help to get up.

\subsubsection{Occlusion}
Because of the prevalence of vegetation, it is common for people to be not fully visible.
In this orchard setting, for instance, people are often working in the trees, so the most common types of occlusion are from the side by tree branches and from below by undergrowth in the rows. Autonomous systems frequently need to push through tree branches, so it can be very safety-relevant to be able to detect people in these partial occlusion situations. A sample of these poses and occlusions is shown in Figure~\ref{fig:poses_ag_vs_urban}.

\subsubsection{Natural Factors}
Certain other effects, while also present in urban settings, have a more profound effect on agricultural environments. While we do not aim to address these corner cases in this work, they deserve mention due to their importance to robust, long-running autonomous systems: Seasonal changes lead to much more dramatic appearance variation than in typical urban settings. The absence of leaves or other vegetation can shift the entire color and texture distribution of the environment, and it also changes the occlusion characteristics of the scene. While urban settings are composed primarily of rigid objects, agricultural environments often see strong effects from wind, potentially causing dramatic changes for any approaches that incorporate motion information.

\subsection{Related Detection Algorithm Work}
Detection performance has had immense growth in recent years, spurred by excitement in applications, useful datasets, and algorithmic advances. \revised{While classic approaches tend to make use of visual templates defined on handcrafted features~\cite{felzenszwalb2010object}, recent advances make use of data-driven features learned with deep neural networks~\cite{lecun2015deep}. While such networks were originally designed for object classification, they have been successfully applied to object detection}. \cite{sermanet2013overfeat}'s Overfeat began by sliding a fixed window across the image and computing features, not unlike the previous rigid style detectors, and then classifying/regressing the location.  The R-CNN detector from \cite{girshick2014rich}, on the other hand, utilized a CNN as a feature extractor in a more standard pipeline involving region proposals, bounding box regressors, and SVMs.  Although spatial pyramid pooling (SPP) removed redundant calculations, R-CNN was computationally intensive. \cite{DBLP:journals/corr/Girshick15}'s Fast R-CNN improved computational efficiency more by sharing computation of convolutional features across proposals. In \cite{ren2015faster}, Faster-RCNN combined the region proposal generation into a single neural network, which led to significant speed improvements.  

\revised{In person detection (as in most detection applications), positive instances are also exceedingly rare, vastly outnumbered by the number of detections that should not be produced, even ignoring images that contain no people. This has implications both on which metrics are appropriate for evaluating performance and on the performance requirements for real-world systems, and it must also be taken into account during training, where the data are highly imbalanced.}

When performing detection of only a single class, false positives are mostly due to confusing hard background instances with difficult small scale objects, which are important in safety and surveillance applications. 
On small objects, the Region-of-Interest (RoI) pooling employed by the R-CNN family can cause features to be not discriminative.  To address these issues, \cite{zhang2016faster}'s RPN+BF adds a boosted forest (BF) classifier on top of Faster R-CNN's region proposal network (RPN). Also, feature resolutions are expanded with \textit{\`a trous} convolutions~\cite{mallat1999wavelet} and concatenated at multiple resolution levels across the RPN's network before being fed to the boosted forest.  In \cite{cai2016unified}, 
MS-CNN tackles these issues within one training framework by adapting multiple resolutions to its multi-scale (MS) region proposals, while adding feature up-sampling, hard negative mining, and relative-sized context embedding. Both RPN+BF and MS-CNN perform strongly on urban pedestrian detection benchmarks, but it remains an open question whether such methods are also appropriate for other applications.
\revised{DetectNet}~\cite{detectNetBlog} is a widely used framework due to its availability from NVIDIA and serves as a baseline, generic, off-the-shelf method.
We evaluate these algorithms to determine how well the success of methods on urban detection transfers to off-road or agricultural domains. 
We also explore our own novel person detector, described in detail in Section~\ref{sec:alg}, which incorporates additional surrounding image context that helps with very small pedestrians.

\subsection{Evaluation}

All of the major urban pedestrian detection benchmarks use evaluation metrics based on bounding box overlap in a single image~\cite{geiger2013vision,Dollar2012PAMI:Caltech,keller2011new}. Not all research on perception for autonomous vehicles evaluates object detection performance independently though. A chain of research from \cite{pomerleau1989alvinn} to \cite{bojarski2016end} tries to learn a direct mapping from sensor data to control signals, and other work \revised{focuses} on quantities that only indirectly depend on detections, such as safe speed~\cite{dima2011:PVS} or safe distance in current and adjacent lanes of travel~\cite{chen2015deepdriving}. Although these methods may allow the best evaluation (and possibly learning) for the given system on a particular application, we prefer to remain agnostic to the specifics of the robotic system or how it may apply the results of person detection. We therefore follow the prior datasets and opt for a bounding-box-based training and evaluation approach.

\section{The NREC Agricultural Pedestrian Detection Dataset}\label{sec:dataset}

\subsection{Log Sets and Annotation}\label{sec:logSets}
In order to produce a large scale person detection dataset, we combined data from four field data collections over three years, each completed using similar hardware and collection methodologies. 
Consecutive video sequences (called ``logs'') are collected with the same person in the same outfit in a variety of poses, referred to as a ``log set''. 

We categorize the poses to execute during collection as static, with the person standing or crouching in one location; moving, with a person walking or in the act of standing up; and unusual, which includes a person lying on the ground or falling into view. Each of these categories of poses is consistent within a data collection and should exist in each log set from that collection. Changes in procedure mean that each data collection has slightly different sets of poses though. Since each data collection spans several days, it naturally captures images at various times of day and in different weather conditions.

The outfit, pose, and current weather are recorded as meta-data for the video sequence. These annotations are included in the directory path for each sequence in the dataset and can be used to search or partition the dataset into categories. Each sequence also has associated normalized camera position and projective calibration information.

\subsection{Data Collection Process}\label{sec:collection}

\begin{figure*}[t!]
    \centering
    \begin{subfigure}[t]{0.3\linewidth}
    \includegraphics[width=\linewidth]{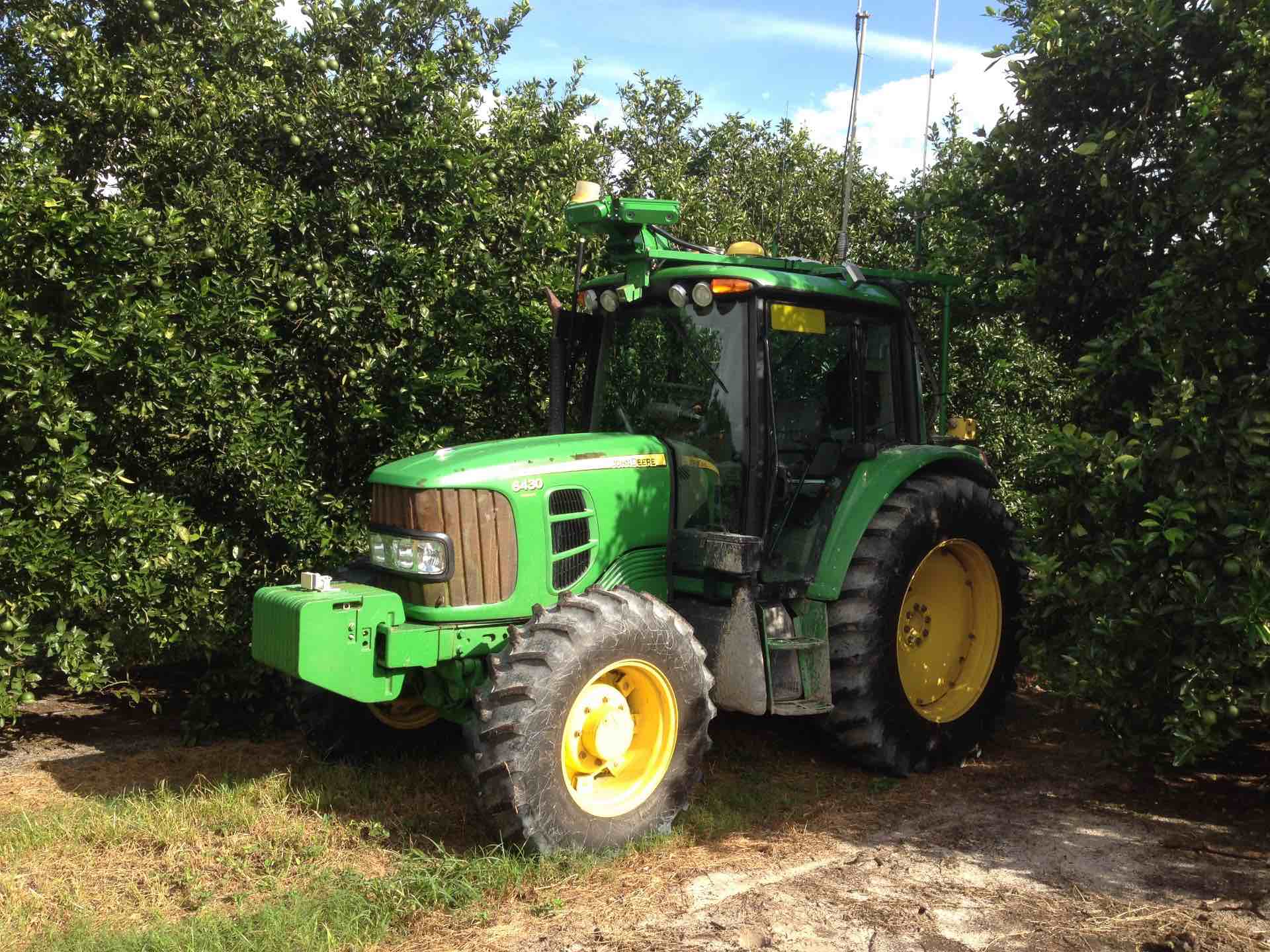}
    \caption{Orange grove tractor}
    \label{fig:platform_orange}
    \end{subfigure}
    \begin{subfigure}[t]{0.169\linewidth}
    \includegraphics[width=\linewidth]{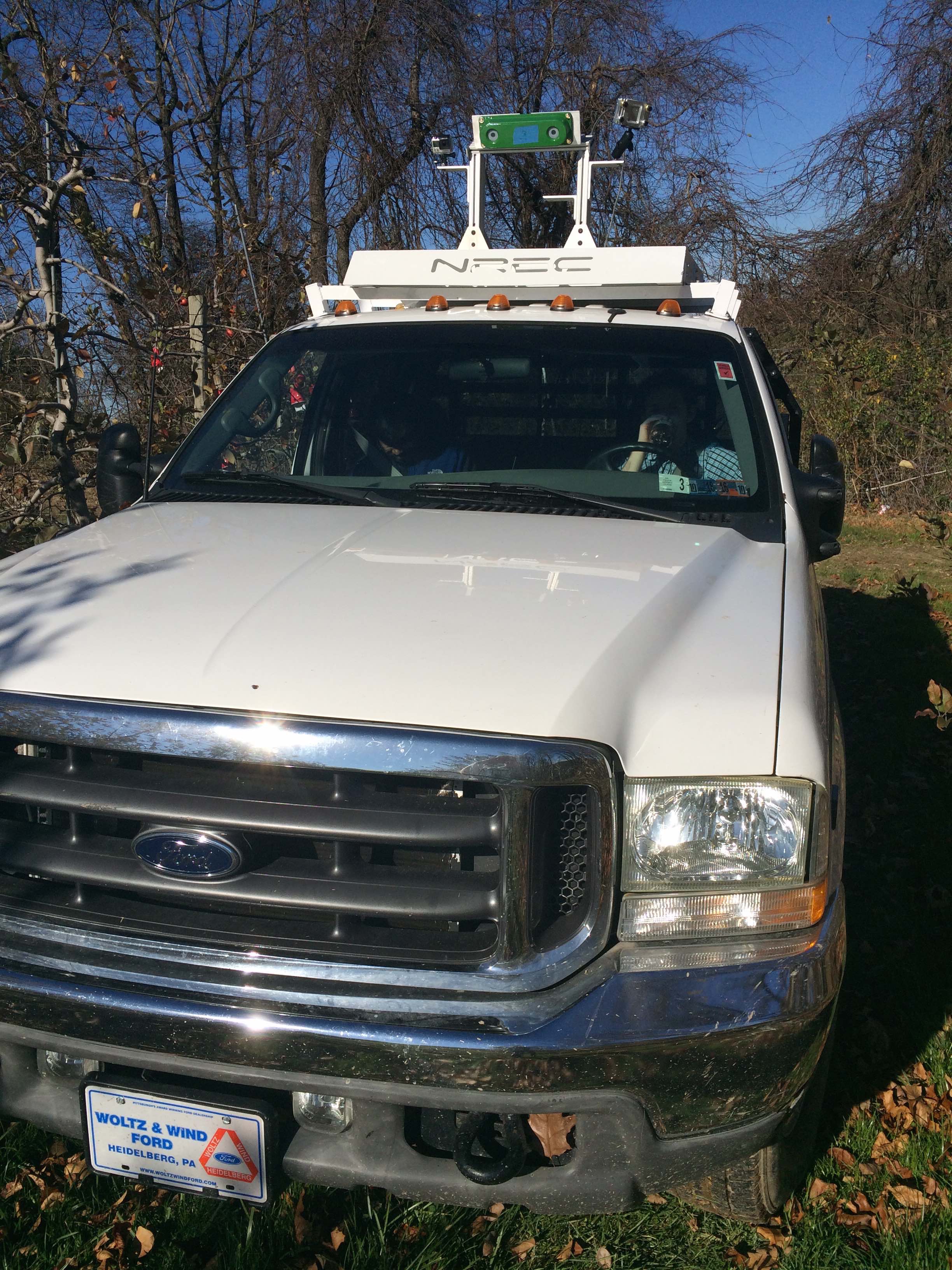}
    \caption{Apple orchard truck}
    \label{fig:platform_apple}
    \end{subfigure}
    \begin{subfigure}[t]{0.3\linewidth}
    \includegraphics[width=\linewidth]{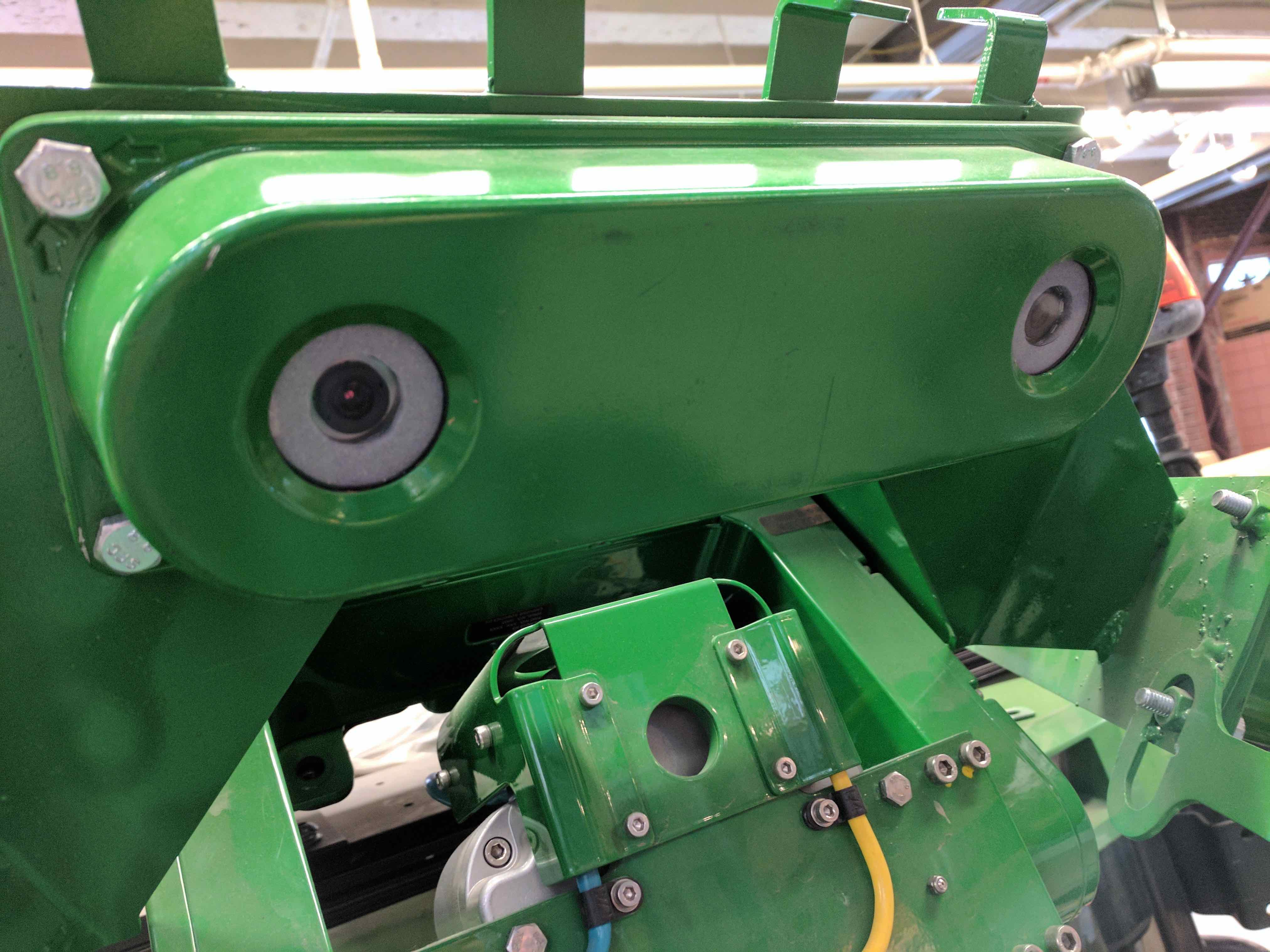}
    \caption{Camera hardware}
    \label{fig:platform_camera}
    \end{subfigure}
    \caption{Data collection platforms. The camera is mounted 3m above the ground and tilted down about 30$\degree$ from horizontal on the tractor and at a height of 2.7m with a tilt of ~26$\degree$ on the pickup truck.}
    \label{fig:platform}
\end{figure*}

The dataset consists of four different data collections, from 2013 to 2015, designed to cover a variety of conditions and two different environments.
For all collections, images were acquired with a \revised{stereo} pair of custom high-dynamic-range cameras (shown in Figure~\ref{fig:platform_camera}\revised{)}, separated by a 20cm baseline, capturing $720 \times 480$ images at a frame rate of 7.5 Hz. \revised{With the very strong sun and deep shadows common in these environments, we required extreme dynamic range, which limited our choice of imager. The limited resolution and lower frame-rate should not be a restriction given the relatively low speeds of the vehicle in these operations.} Vehicle position was simultaneously recorded using an RTK GPS. 
Intrinsic and extrinsic calibration of the camera system was also performed for each collection for inclusion with the data. On both platforms, the camera was mounted at the front of the vehicle\revised{'s} roof, looking forward, and angled down to place the image top edge near the horizon. More information about provided camera calibration vehicle position measurements are in Appendix~\ref{sec:appCalib}.

\subsubsection{Orange Grove}

Three data collections occurred in a commercial orange grove operated by an industry partner in central Florida. Previous work in this orange grove includes autonomous tractor development described in~\cite{Moorehead12:Orchard} and system-level performance evaluation of autonomous tractors described in~\cite{dima2011:PVS}. The same autonomous tractor research platform from those projects (shown in Figure~\ref{fig:platform_orange}) which has done over 1,500 km. of useful work in this grove, was used for these data collections.

The scale of the orange grove is shown in Figure~\ref{fig:orange_orchard}. Our data do not cover the full extent of the grove, though we use different regions during different collections, to avoid seeing the same section in multiple log sets. An important feature of the grove is that alternating gaps between lines of trees have differing geometry and weed removal strategies. These two types of gaps are called ``beds'' and ``swales''. The beds are at the same height as the ground under the trees and have limited weed growth, resulting in a clearer view of the ground and possible obstacles. In contrast, swales are sunk one meter into the ground, are used for drainage, and support fast weed growth. The concave geometry, muddy terrain, and tall weeds in swales result in more occlusion of obstacles. This collection site is very flat, allowing simplifying assumptions in mapping and making driving at consistent speeds of approximately 5 mph possible with a fixed throttle setting.

\subsubsection{Apple Orchard}

\begin{figure*}[th]
    \centering
    \begin{subfigure}[b]{0.4\linewidth}
    \begin{tabular}{rc}
    \rotatebox[origin=l]{90}{\hspace{1cm}3~miles}&\rotatebox{90}{\includegraphics[height=.8\linewidth]{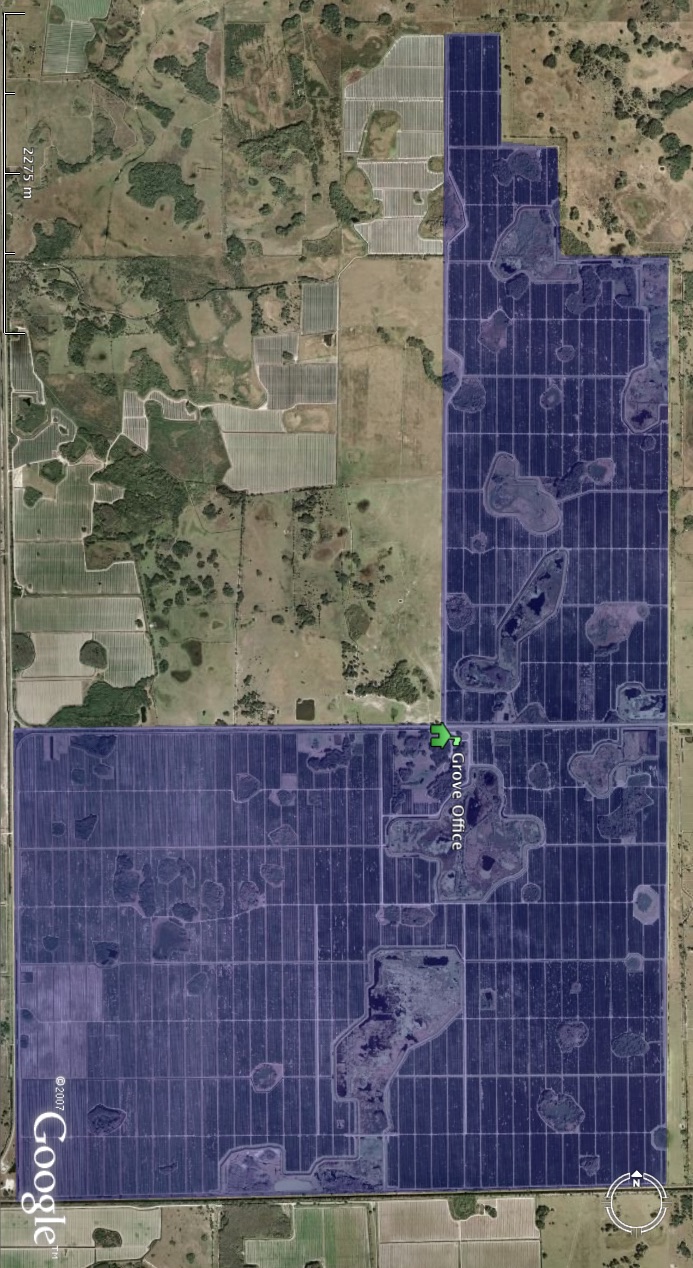}}\\
    &4.5 miles
    \end{tabular}
    \caption{Orange grove}
    \label{fig:orange_orchard}
    \end{subfigure}
    \begin{subfigure}[b]{0.4\linewidth}
    \begin{tabular}{rc}
    \rotatebox[origin=l]{90}{\hspace{.9cm}$\sim$0.25~miles}&\includegraphics[width=.8\linewidth]{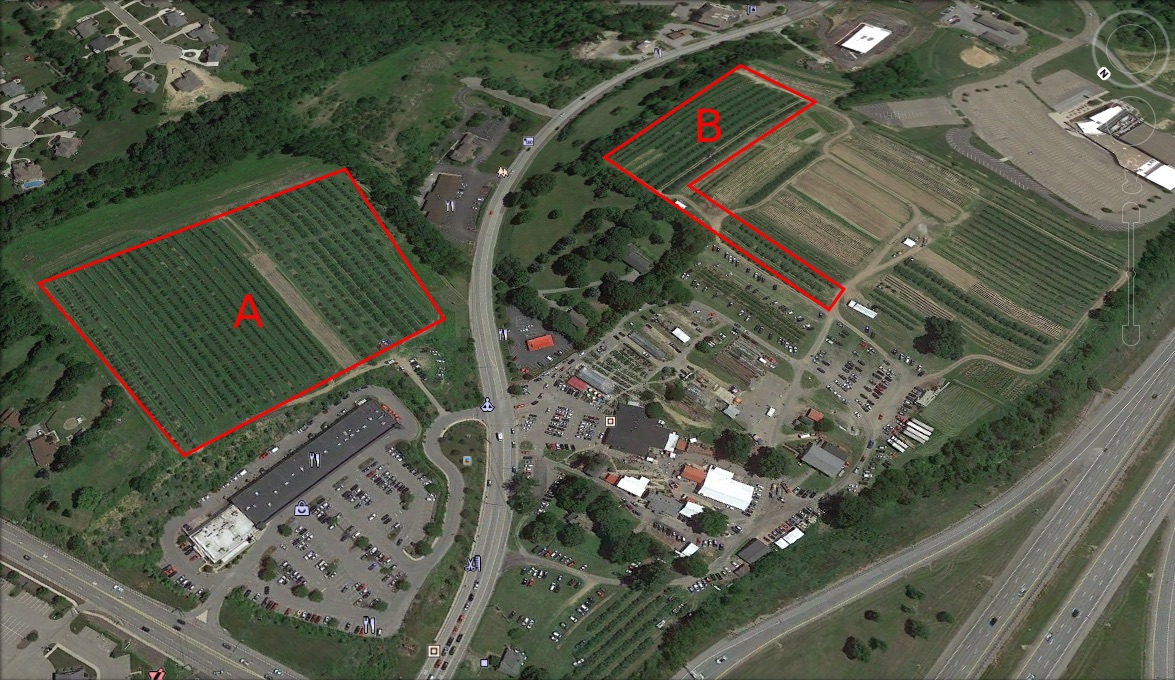}\\
    &$\sim$0.5 miles
    \end{tabular}
    \caption{Apple orchard}
    \label{fig:apple_orchard}
    \end{subfigure}
    \caption{Overhead view of the Devil's Garden Grove orange orchard and Soergel's apple orchard (courtesy of Google Earth) with annotation of where data were collected in this work.}
    \label{fig:orchards}
\end{figure*}

One data collection was performed in an apple orchard at Soergel Orchards in Wexford, PA. As can be seen in the overhead view of the orchard in Figure~\ref{fig:apple_orchard}, this site was much smaller than the orange grove, necessitating several passes through the same rows. We consequently approached each row from both directions and at different times of the day to try to capture as much variety as possible. 

These data were collected using a pickup truck platform shown in Figure~\ref{fig:platform_apple}. In this collection we tried to maintain a 5 mph fixed speed, but using a gas pedal and driving on hilly terrain led to more variation in vehicle speed. This collection also includes some higher frame rate data not found in orange and not used in the benchmark, which is described in Appendix~\ref{sec:appUnlabeled}.

\subsubsection{Static Person}
Across all of our data collection trips, we consistently collect video sequences where just the tractor is moving and the subjects stay in place. These videos still exhibit frames with substantial appearance variation as the vehicle approaches the person, most noticeably in scale, but also in occlusion. These sequences are especially useful for profiling vehicle safety, such as in~\cite{dima2011:PVS}, since it can be difficult to estimate the distance to a moving person in the general case. These logs include people standing and crouching.

\begin{figure*}[th]
\centering
\begin{tabular}{cccccc}
\multirow{3}{*}{\includegraphics[width=4.7cm]{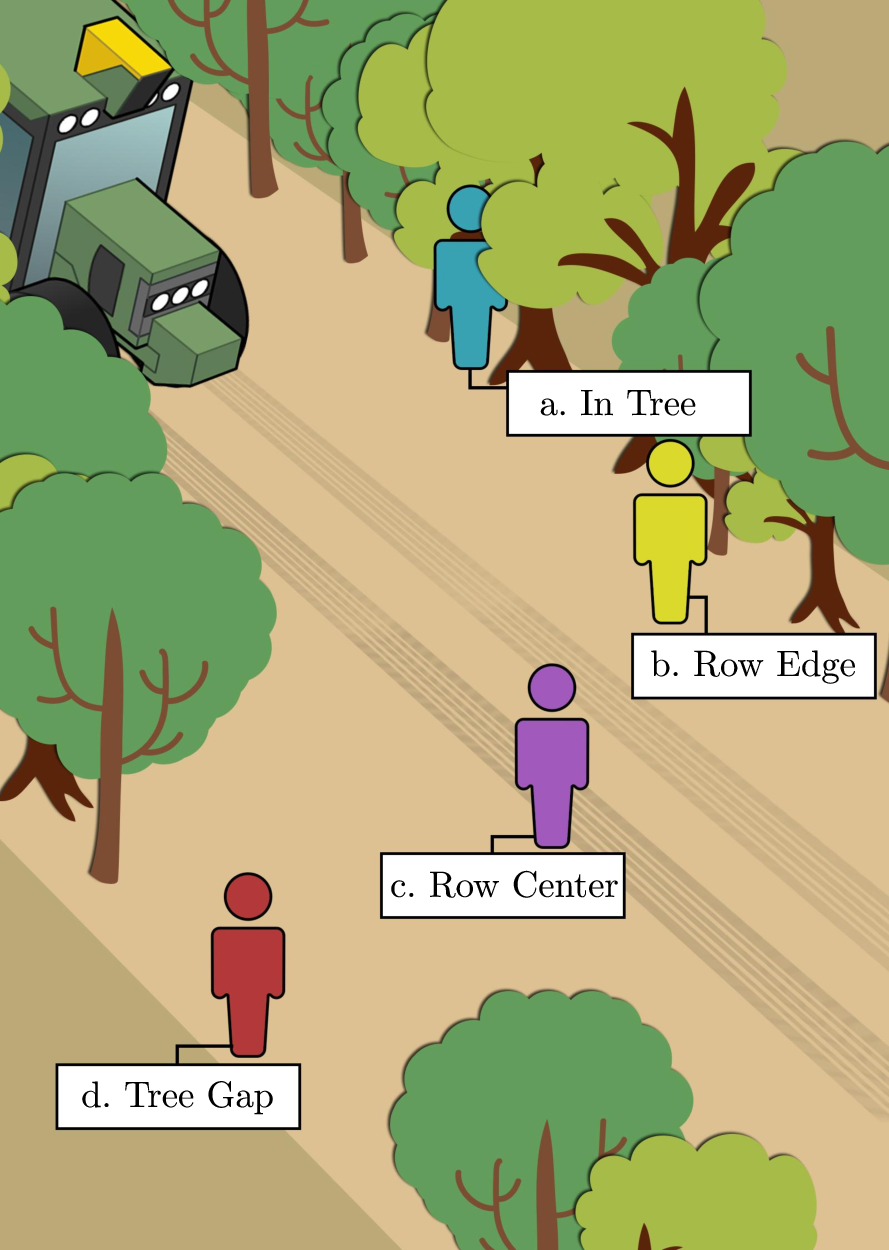}}                   & & (a) In Tree & (b) Row Edge & (c) Row Center & (d) Tree Gap \\
& \rotatebox{90}{\hspace{1cm}Orange} & \includegraphics[height=3cm]{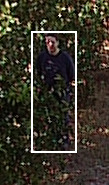}  & \includegraphics[height=3cm]{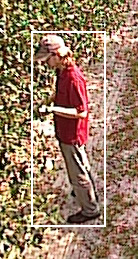} & \includegraphics[height=3cm]{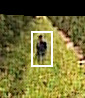}  & \includegraphics[height=3cm]{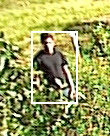}  \\
& \rotatebox{90}{\hspace{1.15cm}Apple} & \includegraphics[height=3cm]{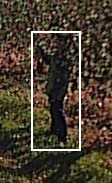}  & \includegraphics[height=3cm]{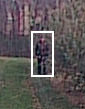} & \includegraphics[height=3cm]{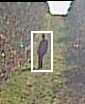}  & \includegraphics[height=3cm]{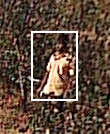}  
\end{tabular}
\caption{Typical locations of static people during data collection. \textbf{a. In Tree} have a person with heavy occlusion to the side and/or from above throughout most the sequence. \textbf{b. Row Edge} have a person with light occlusion from the side for some of the sequence. \textbf{c. Row Center} only includes occlusion from below from weeds or scene geometry. \textbf{d. Tree Gap} has a person standing in a location with a missing tree, so they are farther from the row, but not occluded by trees when the vehicle is close.}
\label{fig:staticPerson}
\end{figure*}

In collecting these data, we strove to include variation in occlusion in a principled and domain-useful way. To that end, we include a set of standard positions in the row for each person-outfit in a data collection. In Figure~\ref{fig:staticPerson}, examples are shown of these standard positions for the final collection. 
In a standard log set, a log was collected with the subject in each of these positions, both standing and crouched. In apple data, we also include examples of subjects on ladders leaned up against the trees, which is a common sight during picking season.

\subsubsection{Moving Person}\label{sec:collect_moving_person}

\begin{figure}[th]
    \centering
    \includegraphics[width=3.25in]{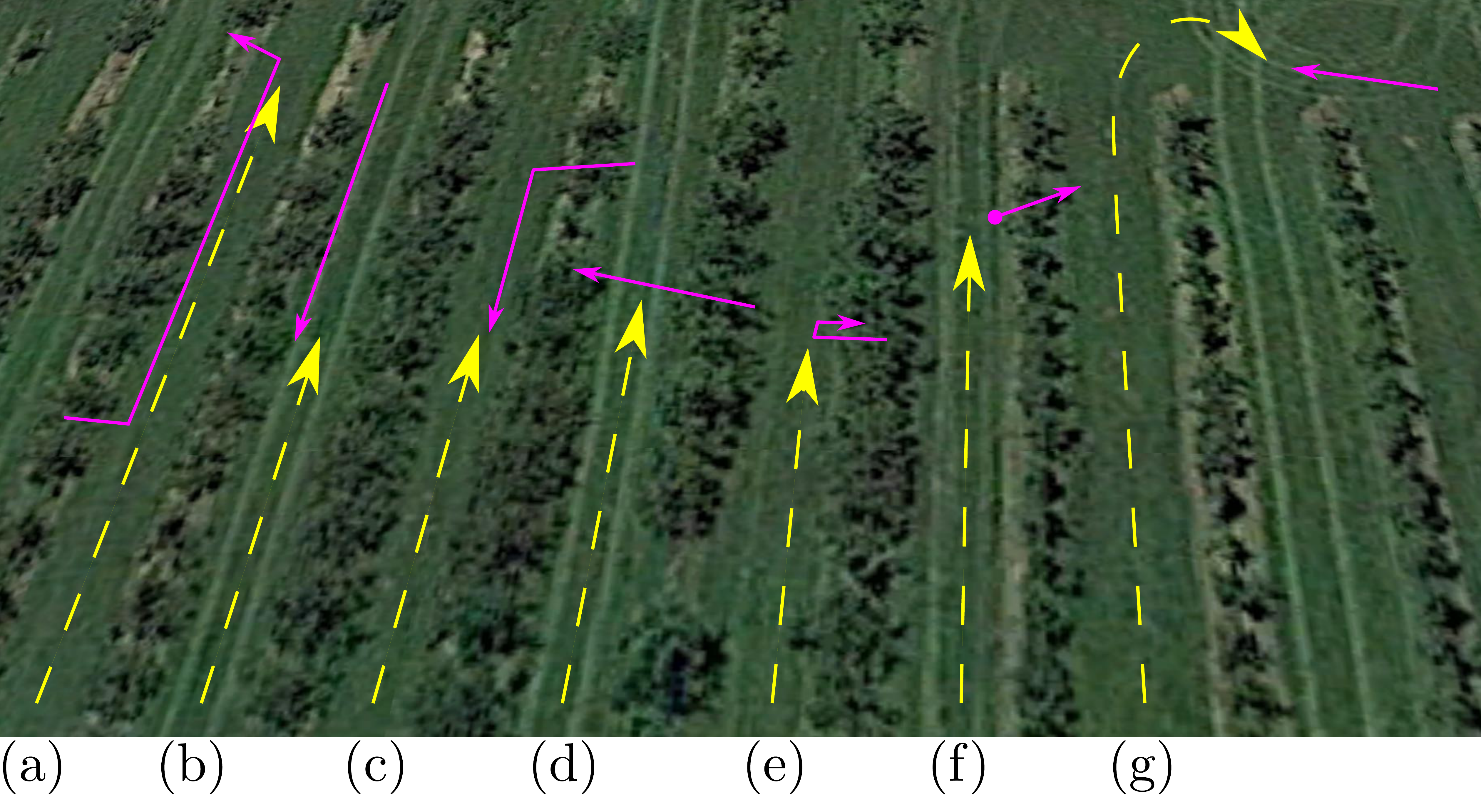}
    \caption{Typical paths for vehicle (yellow dashed) and person (magenta solid) in moving person logs. Row (a) shows an \textbf{Away} sequence, where the vehicle follows behind a person walking along the row. (b) and (c) show the \textbf{Toward} sequences, with the person approaching the moving vehicle along the edge and center of the row respectively. Rows (d) and (e) show a \textbf{Cross} and a \textbf{Step} example, where the subject crosses or steps in front of the vehicle at one of a few distances. Row (f) shows an example of \textbf{get up and leave}, where a seated subject stands up as the vehicle approaches and exits to the side of the row. Finally, row (g) shows a vehicle-person interaction at a row turn.}
    \label{fig:movingPerson}
\end{figure}

\begin{figure}[th]
    \centering
    \includegraphics[width=.48\textwidth]{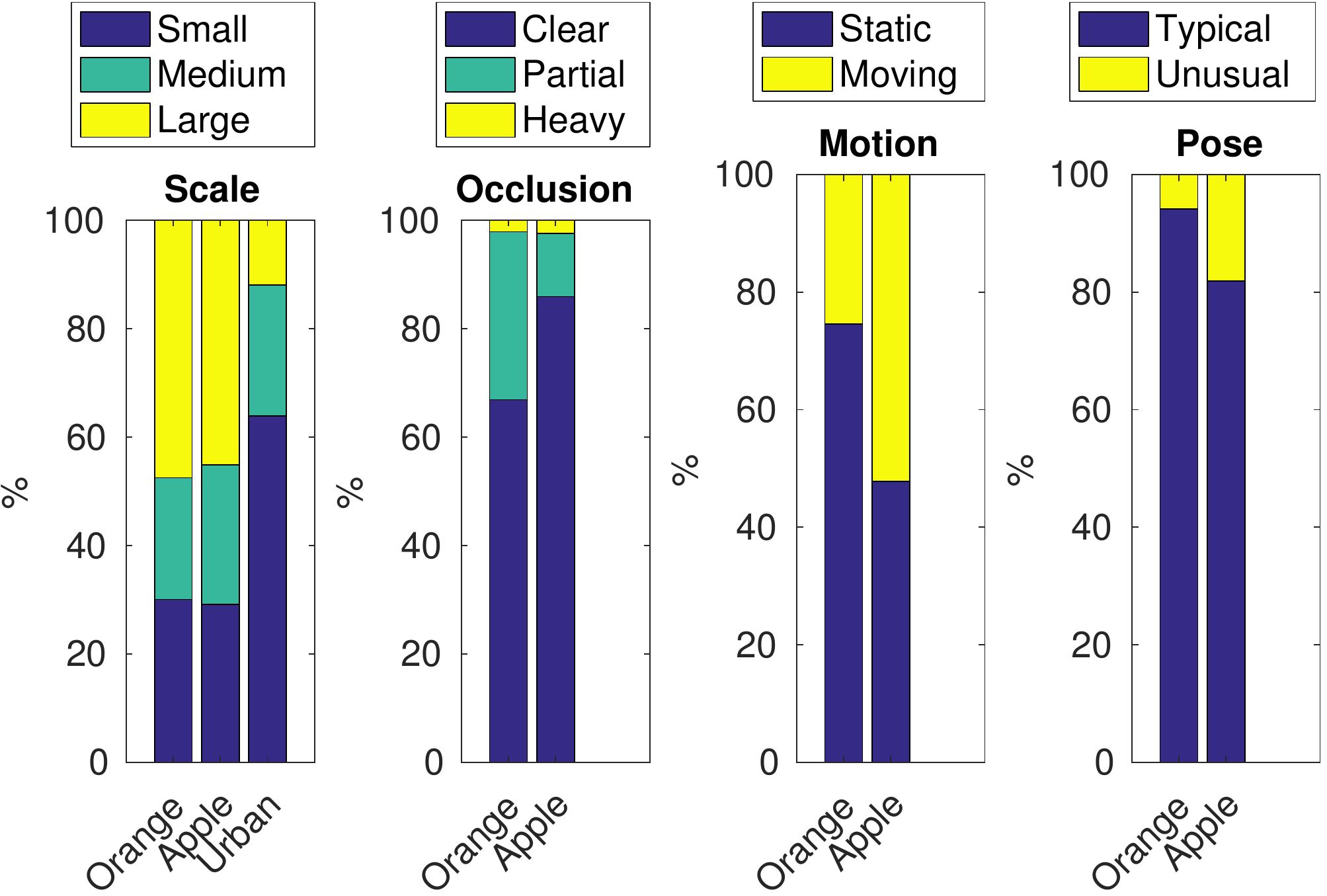}
    
    \caption{\revised{Breakdown of data for each domain into subsets based on bounding box scale, level of occlusion, whether the person is in motion, and the person's pose. Urban values come from the Caltech dataset, \cite{Dollar2012PAMI:Caltech}.}}
    \label{fig:subsets}
\end{figure}

Later data collections focused on including examples of people moving at a comfortable walking pace while the vehicle was parked or in motion. For each person-outfit, we collected sequences for each of the paths shown in Figure~\ref{fig:movingPerson}. 
The paths in which the person's trajectory is perpendicular to that of the vehicle path were repeated multiple times for the motion to occur at different distances from the vehicle (not depicted in the figure).
In particular, paths (b) and (c) \textbf{Toward} and (g) \textbf{Row Turn} all end with the person and vehicle intersecting at a stop at the same location, so they're each collected once. Paths (d) \textbf{Cross} and (e) \textbf{Step} can both be performed at any distance from the vehicle; we choose to collect each at three different distances. Path (f) \textbf{Get Up and Leave} was considered too similar to other paths to collect it more than once for each person-outfit. Additionally, we collected logs with the vehicle parked where the subject was invited to perform whatever actions and assume whatever poses they wished. These were annotated as \textbf{Continuous} or \textbf{Walk Around} sequences.

The counts of labels for the Static People and Moving People categories are shown for each environment in Figure~\revised{\ref{fig:subsets}}. The two groups are of about the same size in apple data, but there are about three times as many static people as moving people in orange data.

\subsubsection{Person in Unusual Pose}\label{sec:collect_unusual_pose}

\begin{figure*}[th]
    \centering
    \begin{subfigure}[t]{0.1\textwidth}
    \includegraphics[width=\linewidth]{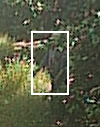}
    \caption{}
    \end{subfigure}
    \begin{subfigure}[t]{0.1\textwidth}
    \includegraphics[width=\linewidth]{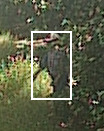}
    \caption{}
    \end{subfigure}
    \begin{subfigure}[t]{0.1\textwidth}
    \includegraphics[width=\linewidth]{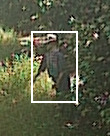}
    \caption{}
    \end{subfigure}
    \begin{subfigure}[t]{0.1\textwidth}
    \includegraphics[width=\linewidth]{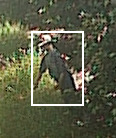}
    \caption{}
    \end{subfigure}
    \begin{subfigure}[t]{0.1\textwidth}
    \includegraphics[width=\linewidth]{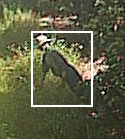}
    \caption{}
    \end{subfigure}
    \begin{subfigure}[t]{0.1\textwidth}
    \includegraphics[width=\linewidth]{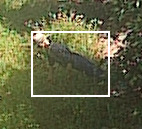}
    \caption{}
    \end{subfigure}
    \begin{subfigure}[t]{0.1\textwidth}
    \includegraphics[width=\linewidth]{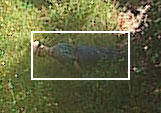}
    \caption{}
    \end{subfigure}
    \begin{subfigure}[t]{0.1\textwidth}
    \includegraphics[width=\linewidth]{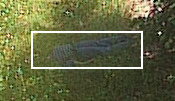}
    \caption{}
    \end{subfigure}
    \begin{subfigure}[t]{0.1\textwidth}
    \includegraphics[width=\linewidth]{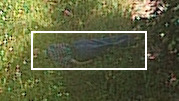}
    \caption{}
    \label{fig:fallen}
    \end{subfigure}
    \caption{Typical \textbf{Fall} sequence captured with Manequin.}
    \label{fig:falling}
\end{figure*}

\revised{We also include some logs of people in poses that are less common to pedestrian detection, but are important to cover here due to their potential safety-critical nature: those where someone is vulnerable (perhaps incapacitated) and unable to move to safety. For each person-outfit, we collected examples at different distances of a person falling into the row. These falls were always simulated with a mannequin to maintain the safety of participants, since they require subjects in positions where they are both vulnerable and also often difficult to see. Labeled images of a sequence showing a person falling are shown in Figure \ref{fig:falling}. We also collected static logs where a mannequin is already lying in the row, either along the row or across it, like in Figure \ref{fig:fallen}. Finally, we include logs of people standing on ladders in the row, since these can be frequently found in orchards.}

\revised{These logs are described as the Unusual Poses group, with all other logs being considered to represent Typical Poses.} As can be seen in Figure~\revised{\ref{fig:subsets}}, Unusual poses make up about $\sfrac{1}{17}$ of the orange data and about $\sfrac{1}{6}$ of the apple data. They are also the primary contributor to the long tail of the distribution of aspect ratios shown in Figure~\ref{fig:aspect_ratios}\revised{, since falling or fallen people's bounding boxes are wider than they are tall. Logs of falling people always begin with the person at the edge of the tree line and therefore exhibit a range of occlusion levels. People lying down, however, are always fully within the row, so their occlusion comes from grass and weeds; they therefore show much more occlusion in the orange data, where grass/weeds are taller.}

\subsubsection{Negative/No Obstacle}\label{sec:negative}

With the controlled farm environment, it is possible to also collect video sequences that do not contain any people. Interspersed with our collection of positive examples, we also collected logs that we expect to be empty as negative examples. This allows us to include more data without requiring more labeling effort. These negative sequences are often much longer than a typical sequence with a person, and it is not clear which sections are most important. \revised{We randomly (to avoid possible aliasing effects) sub-sample each video sequence at a rate of about one image per second to guarantee significant new content / change in viewpoint in each image. This resulted in about $\sfrac{1}{4}$ as many negative images as positive images in each set, balancing between capturing variety from many long logs and avoiding growing the data size unnecessarily; benchmark participants are welcome to experiment with other subsampling strategies though. 

It should also be noted that there is a very asymmetrical relationship between negative and positive training data, even given an equal number of positive and negative images. The vast majority of the aforementioned logs containing people consists of negative data as well, in the form of the non-person region of each positive labeled image. These regions (bounding boxes with no overlap with the person label) were also used as negative training data in our experiments.}

\subsection{Image Labeling Process}
\label{sec:labeling}
Labels of each image containing a person were generated by a dedicated annotator using the Caltech dataset labeling tool~\cite{Dollar2012PAMI:Caltech}. In the Caltech dataset, a person is labeled in each frame with both their visible portion and an estimate of their full extent. The latter requires an annotator to make an assumption about pose information that is not observable though, leaving such labels always inherently uncertain. Additionally, the visible portion of the person is the most relevant for stereo imagery, since it is more readily usable for getting range measurements. In this work, we therefore choose to only label the visible person with a bounding box.

These bounding boxes are first drawn when the subject is at least 20 pixels wide, which we configured as the minimum size on the labeling tool. We never include person labels in the first seven frames of a video sequence. With our frame rate of 7.5 Hz, this means that one second of video is available for initializing motion features before evaluating detectors.

\begin{figure}[ht!]
\centering
\begin{tabular}{ccc|c}
\hline
\multicolumn{3}{c|}{Clear/Partial Occlusion} &  \specialcell{Partial/Heavy\\Occlusion} \\ \hline
\includegraphics[width=1.75cm]{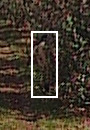} & 
\includegraphics[width=1.75cm]{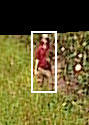} &
\includegraphics[width=1.75cm]{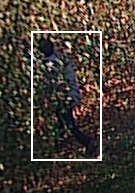} & \includegraphics[width=1.75cm]{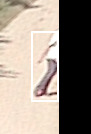} \\ \includegraphics[width=1.75cm]{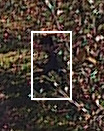} &
\includegraphics[width=1.75cm]{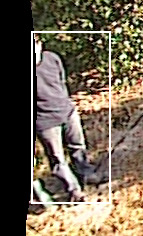} &  \includegraphics[width=1.75cm]{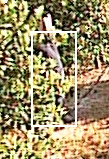} & \includegraphics[width=1.75cm]{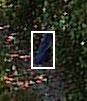} \\
\includegraphics[width=1.75cm]{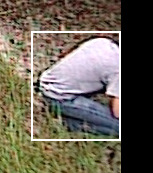} &
\includegraphics[width=1.75cm]{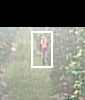} &
\includegraphics[width=1.75cm]{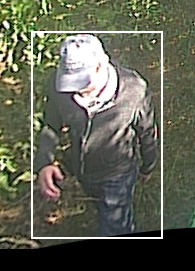} & \includegraphics[width=1.75cm]{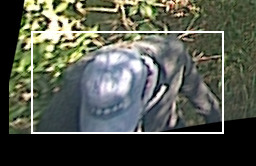} \\ \includegraphics[width=1.75cm]{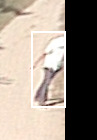} &
\includegraphics[width=1.75cm]{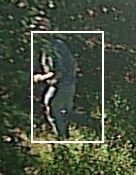} &
\includegraphics[width=1.75cm]{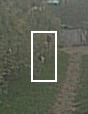} & \includegraphics[width=1.75cm]{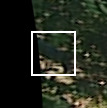} \\
\end{tabular}
\caption{Example occlusion level transitions from the training set. Clear refers to a person labeled as having less than 30\% occlusion. Partial Occlusion refers to a person labeled as having less than 70\% occlusion and more than 30\% occlusion. All images show instances of Partial Occlusion on the boundary. The left columns show instances where the previous or subsequent frame was Clear. The right column shows instances where the previous or subsequent frame was Heavily Occluded.}
\label{fig:occlP}
\end{figure}

In order to still capture which examples are heavily affected by occlusion, we categorize each bounding box into one of three occlusion levels: 
\begin{enumerate}
  \item \textbf{Clean}: Person is more than 70\% visible.
  \item \textbf{Partial Occlusion}: Person is between 30\% and 70\% visible.
  \item \textbf{Heavy Occlusion}: Person is less than 30\% visible. Occurs in few frames, just as the person enters or leaves view. Usually just one body part is visible: arm, leg, body, or head.
\end{enumerate}
Occlusion can either be caused by objects in the scene or by the edge of the image. The difference between these occlusion levels is illustrated in Figure~\ref{fig:occlP}, which shows images during transitions between one occlusion level and another.
These transitions are caused by vehicle and subject motion, especially at the tree line or image edges.

\revised{Due to the prevalence of occlusion in these vegetation-heavy environments, our occlusion categorization is similar to, but more challenging than, that of other pedestrian benchmarks; however, we feel that it is more appropriate to the domain. Caltech~\cite{Dollar2012PAMI:Caltech} has categories for no occlusion (100\% visible), partial occlusion (65\%-100\% visible) and heavy occlusion (20\%-65\% visible). KITTI~\cite{geiger2013vision} also includes categories for ``fully visible'', ``partly occluded'', and ``largely occluded'', but provides no quantitative criteria for the categorization. Daimler~\cite{keller2011new} only evaluates on pedestrians that are fully visible.}

The breakdown of labels by these occlusion categories is shown in Figure~\revised{\ref{fig:subsets}}. There is considerably more partial occlusion in the orange grove than in the apple orchard, because the tree foliage is thicker and remains full all the way to the ground, and there are much taller weeds there, particularly in the swales. Heavy occlusion is fairly rare in both environments, limited to times when a person is just coming in or out of view.

The \Baseline{} set for training and evaluation includes only the Clean and Partial Occlusion labels. The Heavy Occlusion cases are labeled, and we evaluate results on them individually; however, without the context of surrounding images it is very difficult for a human to identify them in many cases. The total counts in those subsets therefore correspond to only Clear and Partial Occlusion cases. \revised{Nonetheless, we hope that including these data in their own subset will help increase the longevity of the benchmark, pushing researchers towards methods that incorporate more context, such as stereo and motion information.} Additionally, counts cover only images containing people, not the negative images that may be present in the same log.

\subsection{Dataset Breakdown}\label{sec:filtering}

The full dataset consists of 76,662 total labeled images containing people, plus 122,395 without any people, covering over 8 hours of video or about half a terabyte of data, divided into 455 individual videos. Eleven different people appear in the images at various times of day, in different seasons, and in two different environments: an orange grove and an apple orchard.

\begin{figure}[th]
    \centering
    \begin{subfigure}[b]{0.32\textwidth}
        \includegraphics[width=\textwidth]{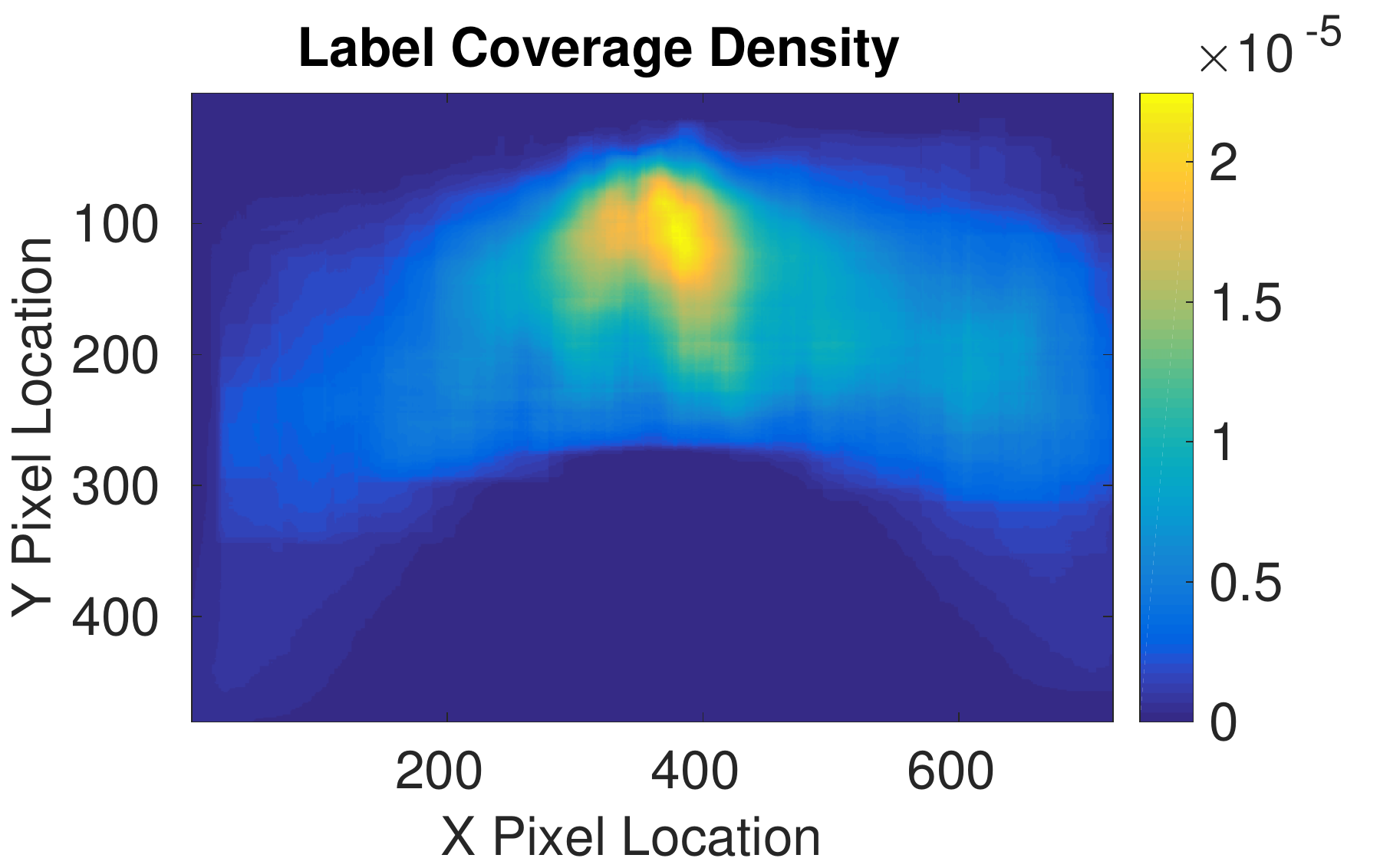}
    \caption{Apple}
    \end{subfigure}
    \begin{subfigure}[b]{0.32\textwidth}
        \includegraphics[width=\textwidth]{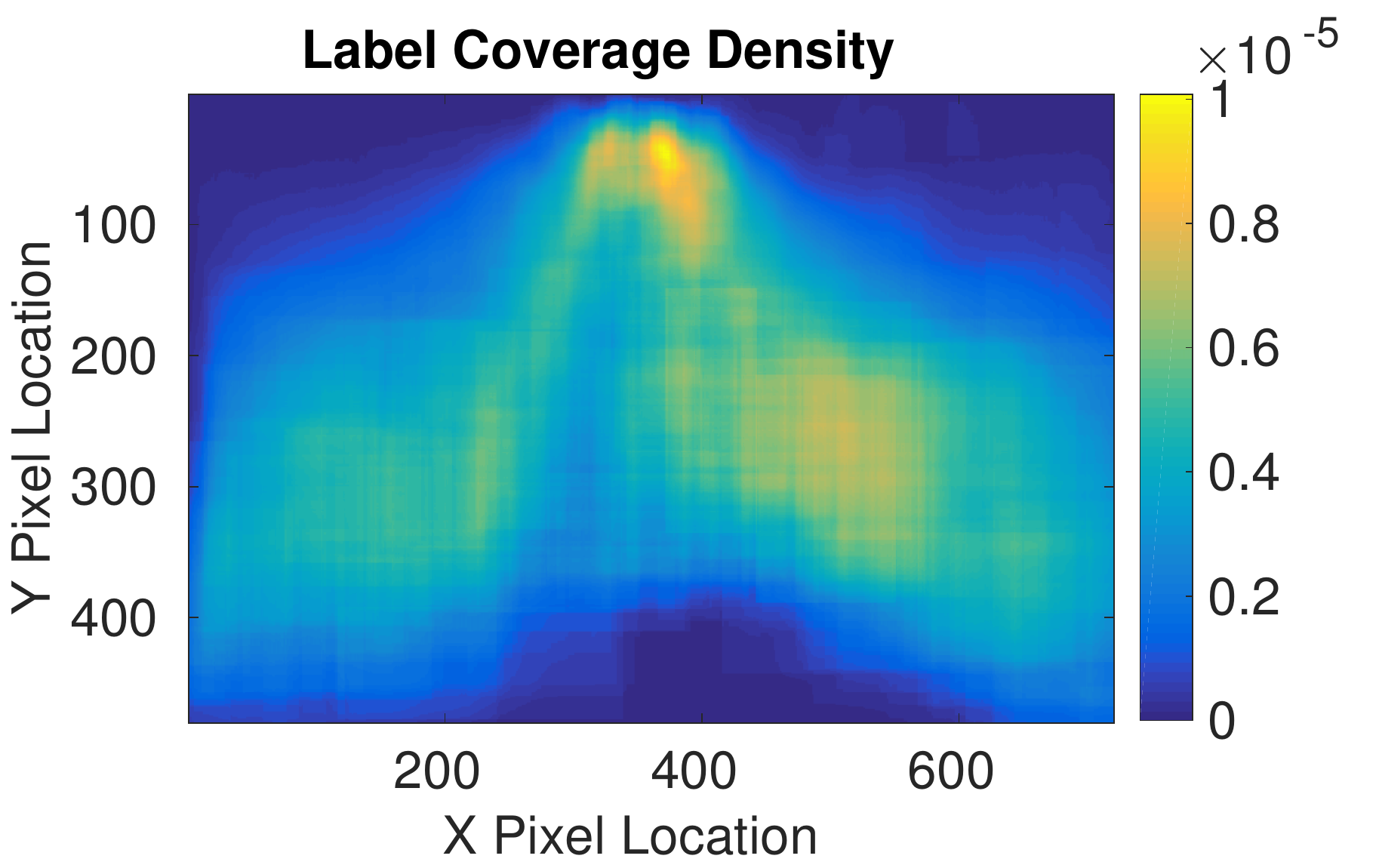}
    \caption{Orange}
    \end{subfigure}
    \caption{Density of label bounding boxes covering each pixel across the image, per dataset. \revised{Shows the frequency with which each image pixel is overlapped by a bounding box.}}
    \label{fig:label_coverage}
\end{figure}

The distribution of label coverage across the image is shown for each environment in Figure~\ref{fig:label_coverage}. Since most logs consist of the vehicle driving down the rows of trees with the people appearing within the row, the geometry of these rows can be seen in both coverage images. Two regions stand out with little to no label coverage: The vehicle hood covers the bottom center of the image, and the upper corners of the image rarely contain people, since a person would have to be at a great distance but not at the row center (only occurring during turns). More of the hood of the pickup is visible in the apple data than of the tractor in the orange data, leading to the larger empty region at the bottom of the apple coverage graph. The greater emphasis on static poses in the orange data also leads to more distinct trails of coverage along where the edge of the treeline projects on either side of the image when looking down the row.

\begin{figure}[th]
    \centering
    \includegraphics[width=.32\textwidth]{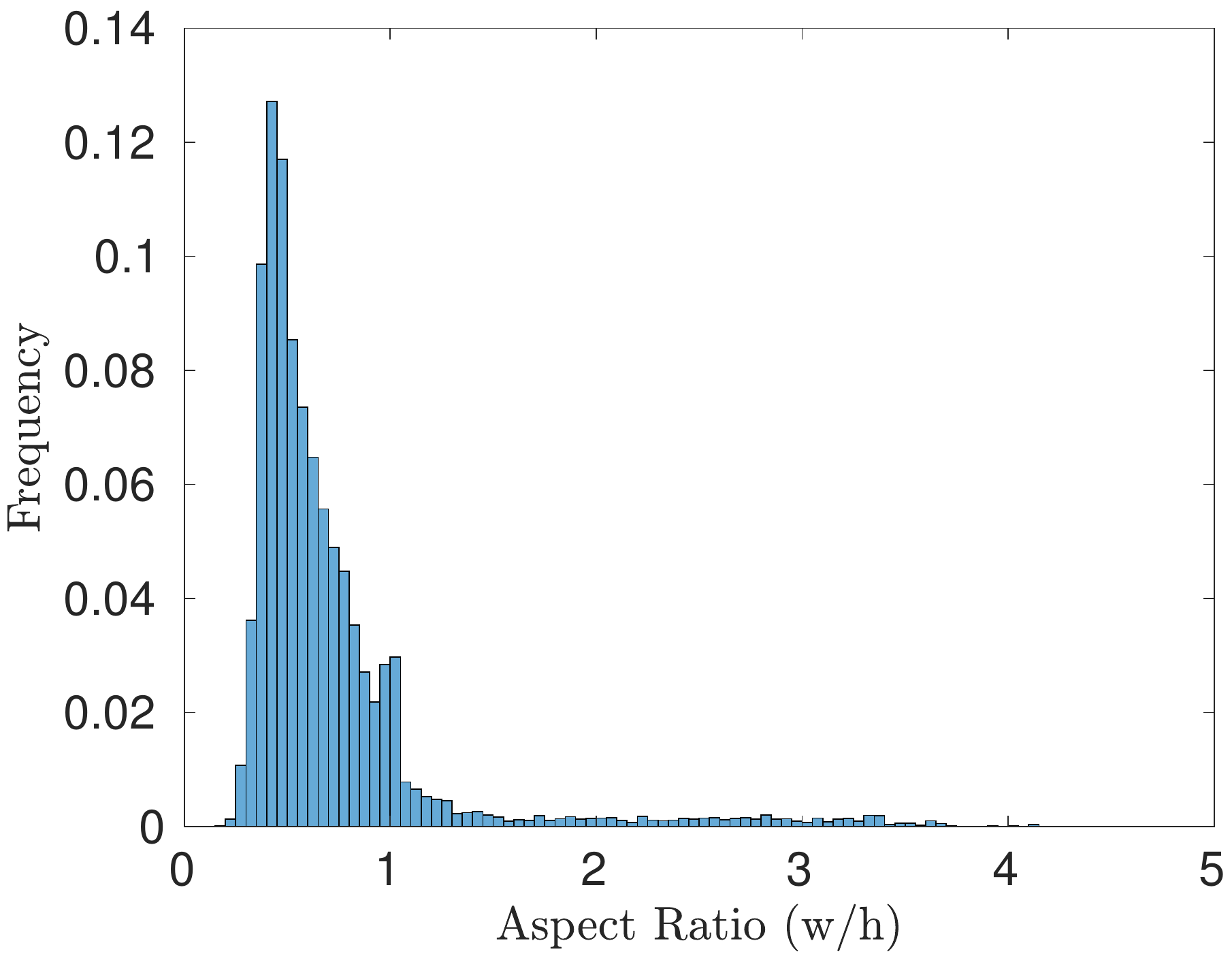}
    \caption{Distribution of \revised{the ratios of width to height of labels} across \Baseline{} dataset. See Figure~\ref{fig:templates} for the set of templates fit to cover this distribution.}
    \label{fig:aspect_ratios}
\end{figure}

The distribution of aspect ratios in the \Baseline{} dataset is shown in Figure~\ref{fig:aspect_ratios}. Though much of the data are clustered around a ratio of 0.5, corresponding approximately to the expected ratio for a fully visible standing person, the distribution is tailed due to the presence of unusual poses and significant occlusion. In evaluation of the Caltech dataset~\cite{Dollar2012PAMI:Caltech}, ground truth bounding boxes are normalized to a common aspect ratio. We choose not to perform such a normalization, since it would heavily skew some challenging examples.

\subsubsection{Training and Testing Data}\label{sec:traintest}

The dataset is divided into three splits: \revised{Training}, Validation and Test. \revised{Training} and Validation are for use in the development and tuning of an algorithm, and only after all such tuning is complete is evaluation on \revised{Test} to be run. In principle, \revised{Training} and Validation can be used in any way  \revised{(such as combining them into a large, joint Training set)}, but Validation is designed to provide a representative pool of data for testing the generalization needed to succeed on the test set, while remaining independent of it. \revised{The Validation and Test logs were chosen simultaneously to meet this goal, each set of logs being required to have an exclusive, diverse set of samples.}

Each log set is classified based on the clothing style of the person in the set. We grouped together sets having similar dress and distributed them between splits, balancing several goals: 
\begin{itemize}
    \item Keep each set of logs (Section~\ref{sec:logSets}) intact, assigned entirely to one split
    \item Target a size ratio of roughly 2:1:1, with the training split being the largest
    \item Maximize variety (dress, subject, pose, motion, weather, time of day) within each split
\end{itemize}
The breakdown of label counts in each split is shown in Table~\ref{tab:label_counts}.

\begin{table*}[th]
    \centering
    \begin{tabular}{c|ccc|c}
        Environment & \revised{Training} (P/N) & Validation (P/N) & Test (P/N) & Total (P/N) \\
        \hline
        Orange & 22,617 / 5,675 & 10,901 / 2,495 & 11,718 / 2,592 & 45,236 / 10,762 \\
        Apple & 15,535 / 4,570 & 8,200 / 1,981 & 7,691 / 1,949 & 31,426 / 8,500 \\
        \hline
        Combined & 38,152 / 10,245 & 19,101 / 4,476 & 19,409 / 4,541 & 76,662 / 19,262
    \end{tabular}
    \caption{Counts of labeled images in each subset for positive (P, containing a person) and negative (N, person-free) images.}
    \label{tab:label_counts}
\end{table*}

\revised{The overall goal is for both Validation and Test to contain a wide variety of data independent of Training, allowing either of them to be a good measure of generalization to unseen data. As a result, performance in generalizing to Validation should be somewhat predictive of performance generalizing to Test.}

\subsection{Data Subsetting}

We defined several filters to divide the data into subsets of interest, to allow more fine-grained analysis of performance under particular conditions. These include the Static vs. Moving people, Typical vs. Unusual poses, and occlusion categories defined in Section~\ref{sec:collection} as well as by environment and scale. Note that these filtering operations can be composed, so that users of the dataset can define very specific subcategories for evaluation; we have limited the subsets shown here for brevity.

\begin{figure}
    \centering
    \begin{tabular}{ccc}
    \hline
    Small/Medium & ... & Medium/Large\\
    \hline
    \includegraphics[width=.089\textwidth]{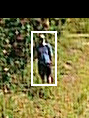} & 
    ... &
    \includegraphics[width=.105\textwidth]{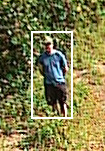}\\
    \includegraphics[width=.088\textwidth]{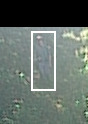} & 
    ... &
    \includegraphics[width=.103\textwidth]{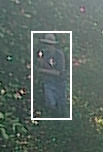}\\
    \end{tabular}
    \caption{Illustration of boundaries between scale categories, where Medium is defined as bounding box area between 1300 and 3500 pixels. All images are of Medium scale. In the left column, the previous frame was Small. In the right column, the subsequent frame was Large.}
    \label{fig:ex_scales}
\end{figure}

\begin{figure*}[th]
    \centering
    \begin{subfigure}[b]{0.7\textwidth}
        \includegraphics[width=\linewidth]{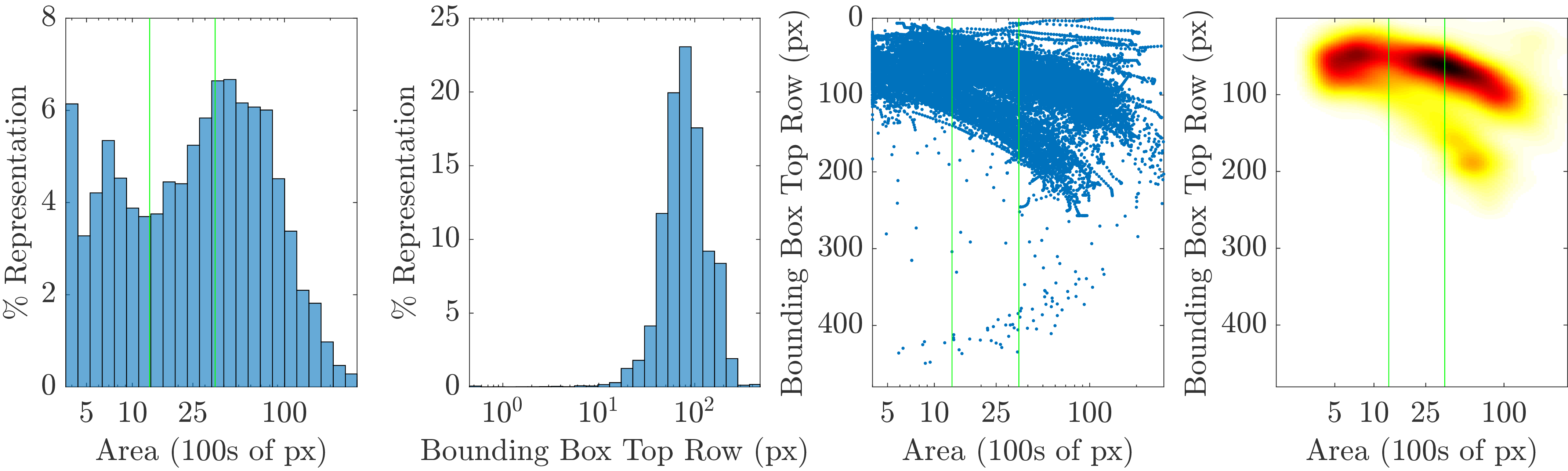}
    \caption{Apple}
    \end{subfigure}
    
    \begin{subfigure}[b]{0.7\textwidth}
        \includegraphics[width=\linewidth]{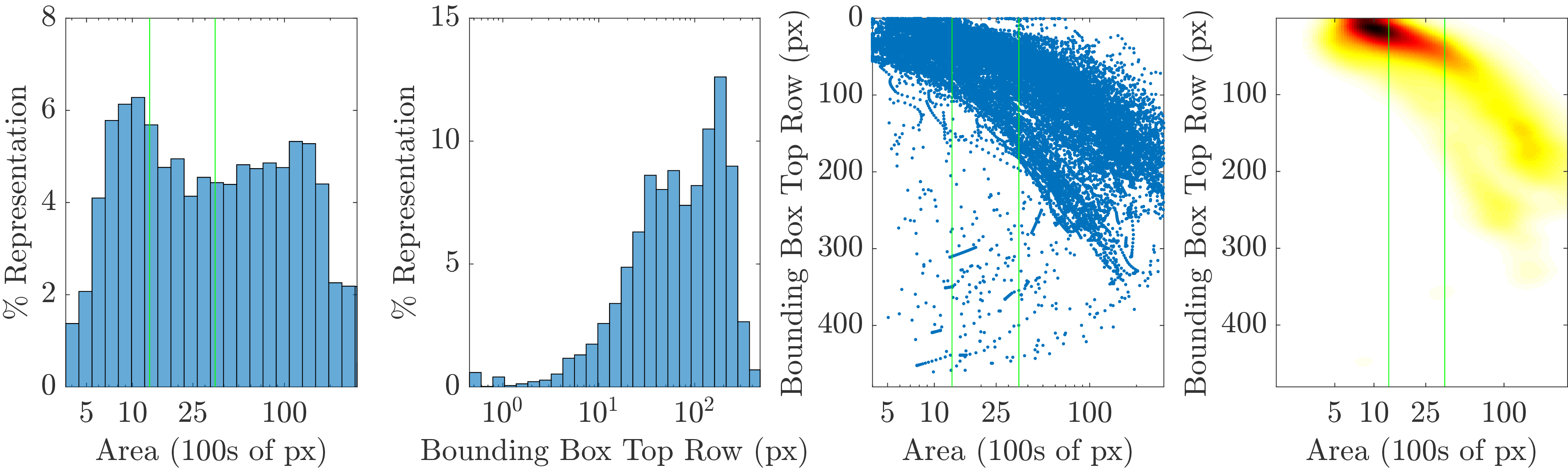}
    \caption{Orange}
    \end{subfigure}
    \caption{Density of bounding box scales, per dataset. From left to right: Two histograms show the univariate distributions of area and the top of the bounding box (both inversely correlated with distance). A scatter plot shows the coverage of the data for each crop, and finally a kernel density estimate shows the log of the density of the points in the scatter plot. A log scale is used for all x-axes, and plots with area are marked with green lines at the boundary between Small, Medium, and Large.}
    \label{fig:label_scales}
\end{figure*}

We divide bounding box scales into categories of small, medium, and large, based on the area of the bounding box, as illustrated in Figure~\ref{fig:ex_scales}. The distribution of scales is shown for each environment in Figure~\ref{fig:label_scales}. Small bounding boxes have area less than 1300 pixels, and large \revised{bounding boxes} have area greater than 3500. Though bounding boxes may grow to be much larger than that, the corresponding distance to the person for these boxes is a fairly small range. The resulting distribution of labels is shown in Figure~\revised{\ref{fig:subsets}}. Many logs begin with the vehicle stationary and a person visible at a great distance and end with the person visible up close when the vehicle comes to a stop, so the concentration of data at these extremes is higher than for the mid-range.

Some estimate of distance categories could be attempted based on the size and location of a bounding box in the image. The correlation between the top of the bounding box and its area is shown in Figure~\ref{fig:label_coverage}. Because of the large variation in pose and occlusion though, these divisions are not clean. We therefore preferred separations based on area, which have consistency in difficulty that are relatively robust to these pose and occlusion changes. The medium scale set should, however, consist of images of people at distances mostly in the range of 10m to 20m.

\section{Evaluation Methodology}\label{sec:eval}

We measure accuracy of detection using the standard bounding box overlap between the detection, $\detection$ and the ground truth bounding box label, $\gtLabel$, given by intersection over union (IoU):
\begin{equation}\label{eq:overlap}
\overlap_{\detection,\gtLabel} = \frac{\detection \cap \gtLabel}{\detection \cup \gtLabel}
\end{equation}
We follow convention in pedestrian detection in ROC curves we present by requiring an IoU score of at least 0.5 to consider a detection correct and plotting miss rate versus false positive rate on a log axis (putting ideal performance in the lower left corner). As discussed in previous work~\cite{tabor2015people}, however, this threshold is overly strict for a number of robotic applications, while it may also be too lenient for others. \revised{There has been some inconsistency in other detection applications for what has been considered an appropriate overlap threshold. Recent work, such as the COCO benchmark~\cite{lin2014microsoftCOCO}, has moved towards averaging accuracies across a set of IoU thresholds.}

\begin{figure}
    \centering
    \includegraphics[width=0.45\textwidth]{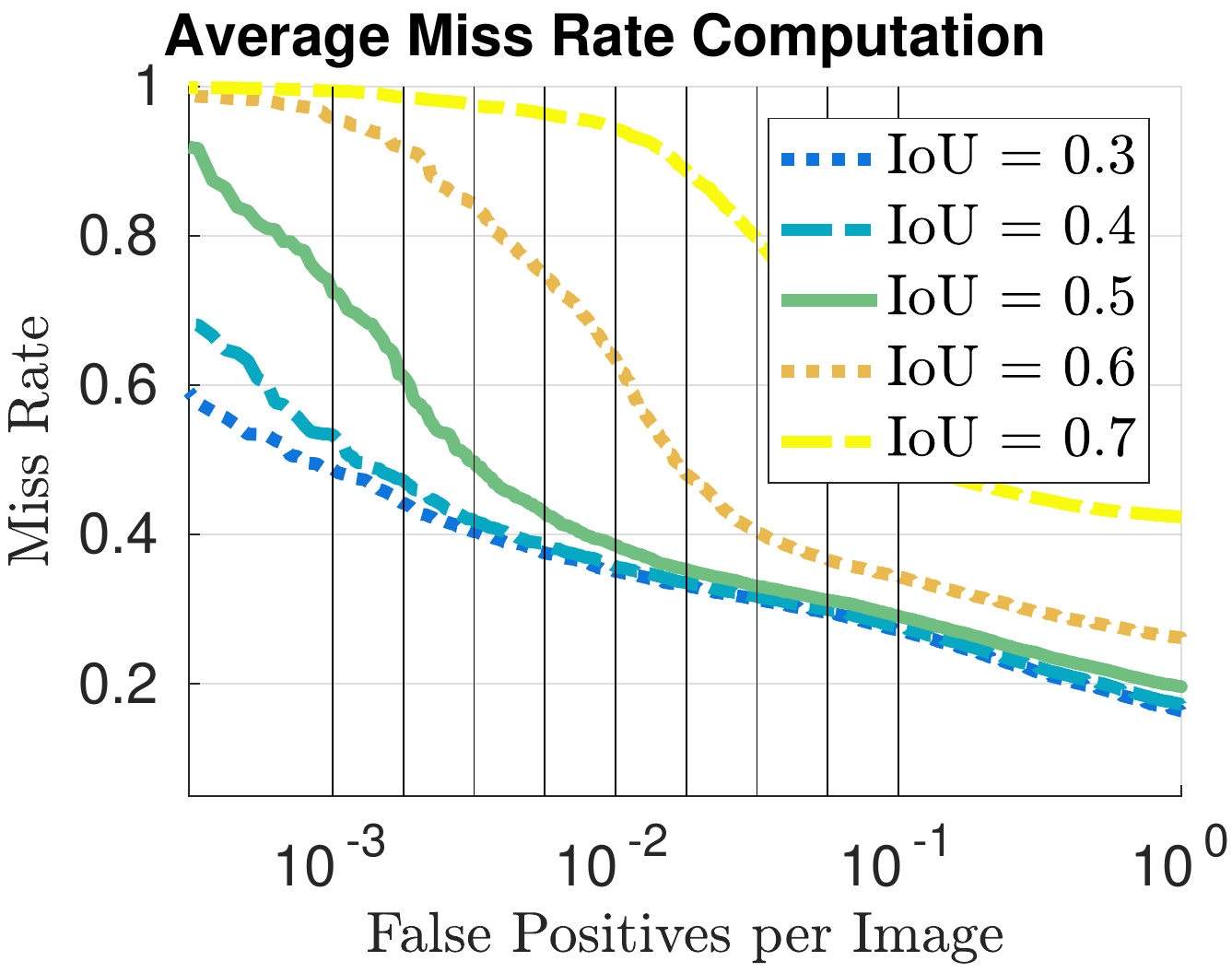}
    \caption{Illustration of the computation of ADR. ROCs are shown for \revised{our proposed method} on \Baseline{} for each IoU step evaluated, and vertical lines indicate the reference FP points sampled for computing the mean. The average miss rate (AMR) is given by the height of every intersection of a vertical line and a colored curve. The final metric, average detection rate is given by 1-AMR.}
    \label{fig:ap_rocs}
\end{figure}

\revised{For overall ranking, we therefore formulate Average Detection Rate (ADR) to be similar to the metric used in the COCO benchmark: IoUs are evaluated in the range $0.3$ to $0.7$ in steps of size $0.1$, as illustrated in Figure~\ref{fig:ap_rocs}, and averaged to produce an overall performance metric. This can be thought of as an approximation to an integral over both classifier sensitivity and required localization accuracy.} This range of overlap values corresponds to values that may be relevant for robotic applications in this domain, depending on the task: The low end (0.3) could be appropriate for tasks that only require detecting the presence of a person or counting people in an area, while the high end (0.7) should provide fine enough localization for interaction tasks. Reference points for ADR computation are at false positive rates in the range $10^{-3}$ to $10^{-1}$ per image, in steps of $10^{1/4}$, also picked as a range of reasonable operating points for an automated system with appropriate filtering. At our frame rate of 7.5 fps, This would correspond to a range of false positive rates roughly between one per second and one every two minutes.

For evaluation on the various data subsets, we always compare against the entire dataset, but use the same ``ignore'' flag method as the Caltech benchmark~\cite{Dollar2012PAMI:Caltech} to filter data out. Matches to ignored data are not considered false positives, but nor are misses of them considered false negatives. This allows the rest of the image, outside the ignored bouding box, to be used for evaluating false positives in every breakdown of the data.

\section{Algorithmic Approach}
\label{sec:alg}

\begin{figure*}[t]
    \centering
    \includegraphics[width=.95\linewidth]{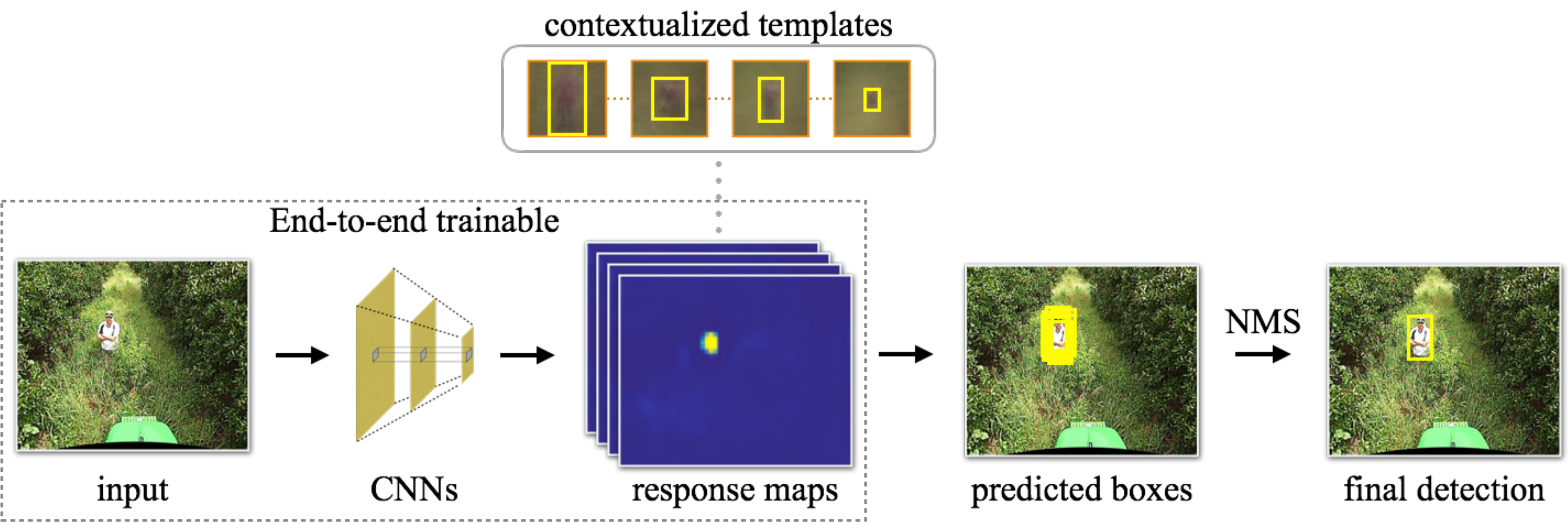}
    \caption{ \label{fig:arch} Overview of our detection pipeline: Given an input image, we first extract multiresolution ``foveal'' descriptors from CNNs, based on which we run our contextualized templates (tuned for different canonical shapes). Each template produces its own dense heatmap predictions (including both detection and regression), which may result in multiple detections on a single person from nearby locations and similar templates. To eliminate redundant hypotheses, we run standard Non-Maximum Suppression. }
\end{figure*}

Pedestrians appear in different bounding box shapes (including aspect ratios and sizes), due to large variety in scale, pose, occlusion etc., as shown in Sections~\ref{sec:labeling} and~\ref{sec:filtering}. To address such variance, we first define a number of canonical shapes and build a scanning-window detector tuned for each canonical shape. We treat the detection for each canonical shape (meaning fixed width $w$ and fixed height $h$) as a {\it binary heatmap prediction problem}, where the predicted heatmap at position $(x,y)$ specifies the confidence of a $h \times w$ detection appearing at $(x,y)$. \revised{To improve localization in the face of variance around each canonical shape, we also learn a linear regression for each of these shapes to locally refine initial bounding box predictions.} We limit the refinement to be an affine transformation and parameterize it in the same way as \cite{girshick2014rich}. We treat bounding box regression also as a {\it heatmap prediction problem} but with {\it continuous \revised{values}}. We train these heatmap predictions, including both detection and regression, using a fully convolutional network~\cite{long2015fully} defined over a state-of-the-art architecture ResNet~\cite{he2015deep}. \revised{While training}, we use logarithmic loss for the detection heatmap and Huber loss for the regression heatmap. \revised{Note that we use a monocular approach here, though stereo data are available for future approaches.} We illustrate our pipeline in Figure~\ref{fig:arch}.

We take the view that \revised{the key way to extend system-level performance of safety systems in these environments is to improve detection of pedestrians while they are still small in the image, as it gives a fast moving vehicle time to react.} We follow the approach in \cite{hu2017tiny} and use ``oversized'' templates\revised{,} whose spatial support includes background pixels surrounding the object of interest, shown as contextualized templates in Figure~\ref{fig:arch}. It turns out that \revised{including massive amounts of surrounding area} (such that 99\% of the template includes the background), which may capture additional contextual cues, such as shadows from a ground plane, is helpful for finding small objects. Such large contextual templates can be efficiently encoded in a multi\revised{-}resolution ``foveal'' descriptor, where background pixels toward the outer edge of the template are represented in lower (coarser) resolution. Such foveal descriptors can be efficiently processed by fully-convolutional networks~\cite{long2015fully} that extract multi-resolution features from multiple layers of a deep network~\cite{hariharan2015hypercolumns}. We call the method Multiscale Foveal Context (MFC) in our results and refer the reader to \cite{hu2017tiny} for \revised{more quantitative analysis, such as how different ways of encoding context affects performance}.

\revised{We use a clustering process to establish canonical bounding box shapes. We cluster in two dimensions, using the heights and widths of bounding boxes in the training data, defining pairwise dissimilarity using Jaccard distance (one minus intersection over union). Each cluster center is taken as a canonical size, and our model learns one template for each. Learning a large set of independent, scale-specific detectors may fail due to data scarcity though, due to the relatively small number of instances of each shape, and testing a large set of such scale-specific detectors can be inefficient.} To address both concerns, we model such scale-specific detectors in a multi-task framework, where they share features from a single hierarchy produced by deep networks. While this produces reasonable accuracy for finding large objects, finding small objects is fundamentally challenging, because few pixels on the object are available for processing. In fact, many prior benchmarks ignore such small instances as being too difficult~\cite{Dollar2012PAMI:Caltech}.

\begin{figure}
    \centering
    \begin{subfigure}[b]{0.4\textwidth}
    \includegraphics[width=\linewidth]{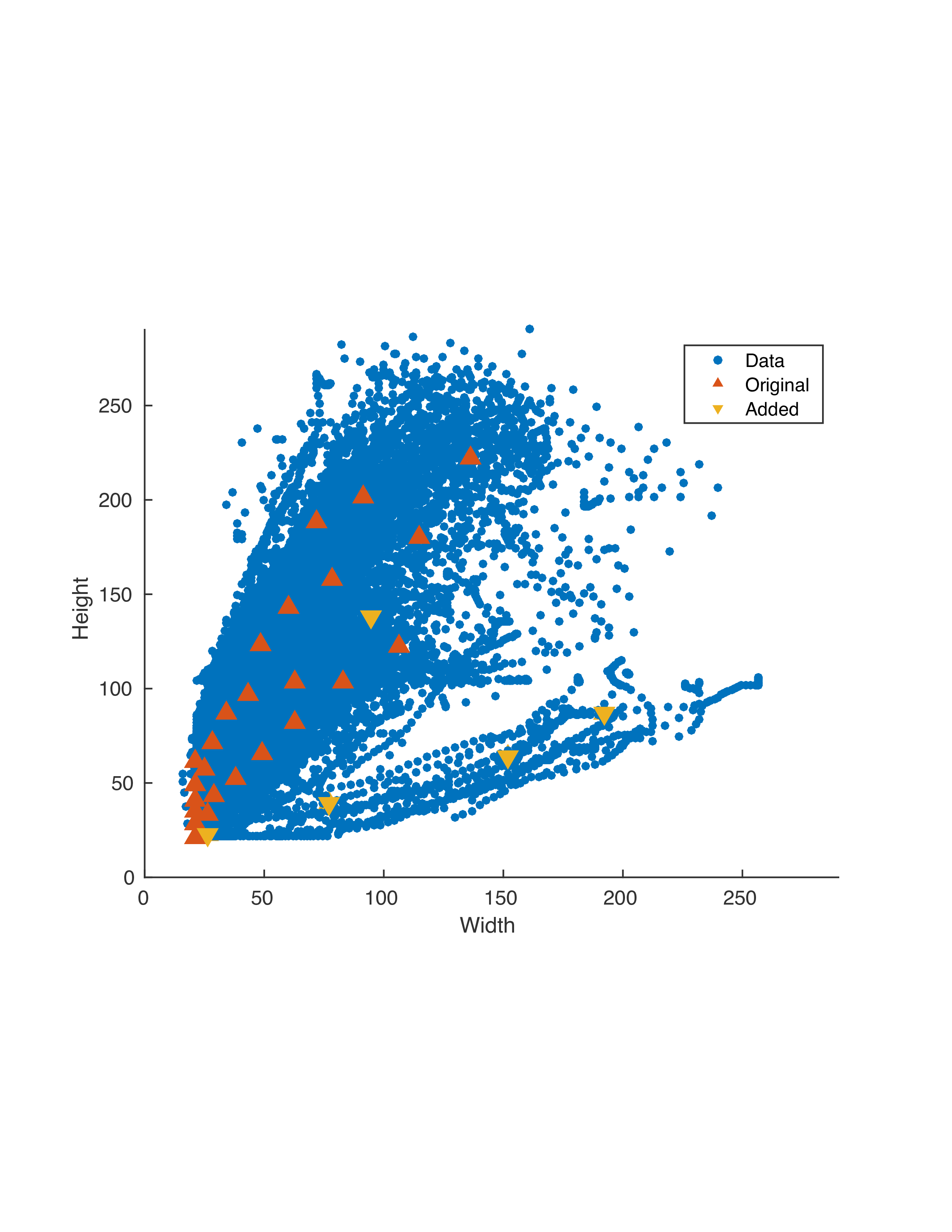}
    \caption{}
    \label{fig:templates}
    \end{subfigure}
    \begin{subfigure}[b]{0.4\textwidth}
    \centering
    \includegraphics[width=\linewidth]{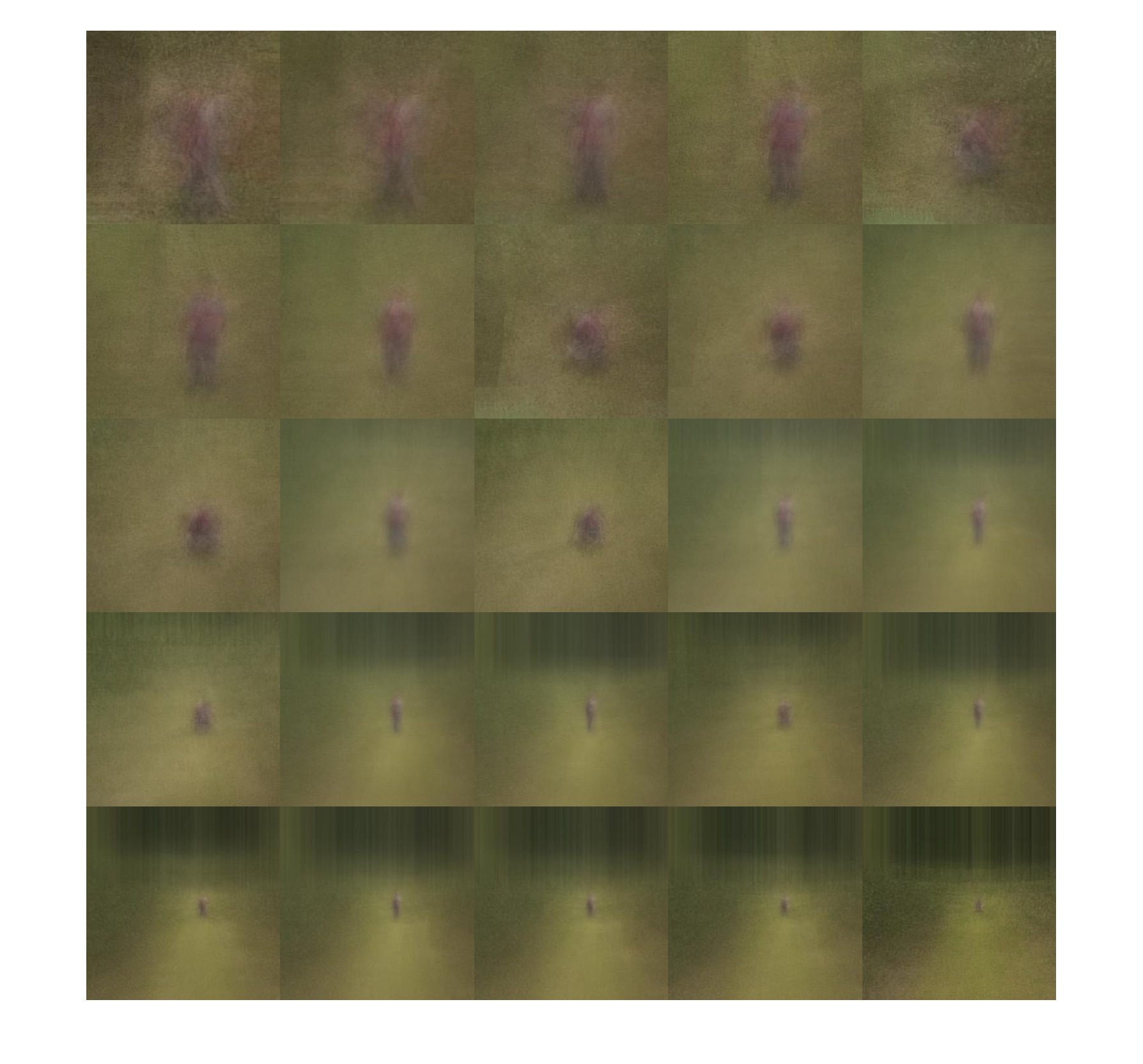}
    \caption{}
    \label{fig:avg}
    \end{subfigure}
    \caption{Templates derived through two-stage clustering process. These templates are used for both \ours{} and RPN+BF methods. (a) shows template sizes. The first 20 templates (orange) were selected from typical pose data, and then 5 more (yellow) were added along with the addition of unusual pose data. (b) shows the average of positive examples for each of 25 templates.}
\end{figure}

{\bf Implementation:} We train a model with 25 different templates, each with a fixed spatial support (300x300 pixels), to find pedestrians of different shapes and sizes (20 - 200 pixels in height). \revised{We choose 25 clusters, since it yields a nice coverage over the size distribution, as shown in Figure~\ref{fig:templates}. We also visualize the averages of positive labeled data assigned to each of the template clusters in Figure~\ref{fig:avg}. Note that templates for small pedestrians (20 pixels) include vastly more context than for large pedestrians (such that they can cover as little as about 1\% of the template area).} We will show that \revised{this} strategy is surprisingly effective for finding small pedestrians. Given training images with ground-truth annotations of objects and \revised{the templates described above, we follow the practice in \cite{ren2015faster} and} define positive locations on the detection heatmap to be those where IOU overlap exceeds 70\% and negative locations to be those where the overlap is below 30\%. We ignore all other locations by zero-ing out the gradient. Note that this implies that each large object instance generates many more positive training examples than small instances. Since this results in a highly imbalanced binary classification training set, we \revised{apply balanced sampling~\cite{girshick2014rich} to mitigate the effect.} We find performance increased with a post-processing linear regressor that fine-tuned reported bounding-box locations. To generate final detections, we apply standard \revised{non-maximal suppression} to the detected heatmap with an overlap threshold of 30\%. We start with a model pre-trained on ImageNet for image recognition (resnet-50), extract features from res3(\texttt{res3\_relu}) and res4(\texttt{res4\_relu}), and fine-tune on our dataset with \revised{stochastic gradient descent. For learning parameters, all experiments are trained with a fixed learning rate of $10^{-4}$, a weight decay of $0.0005$, momentum of 0.9, and a batch size of 20. Our parameters are chosen following previous work that optimizes a similar objective function. Though parameter tuning to the particular domain often leads to improvement in performance, we did not explore that in this work; we focused instead on how readily adaptable existing methods are to the new environment(s).}

\section{Evaluation Results}

\begin{table*}[th]
    \centering
    \begin{tabular}{c|cccc|ccc}
    Subset & \RPNBF{} & \MSCNN{} & \DetectNet{} & \oursOA{} & \oursO{} & \oursA{} & \oursOAC{}\\
    \hline
\Baseline{} & 44.2 & 54.6 & 9.0 & \textbf{59.4} & 46.6 & 47.2 & 57.0 \\
Env=Orange & 46.0 & 55.0 & 11.0 & \textbf{59.1} & 52.0 & 41.9 & 56.5 \\
Env=Apple & 48.5 & 56.6 & 6.9 & \textbf{65.8} & 43.4 & 61.1 & 62.9 \\
All & 43.5 & 53.8 & 8.9 & \textbf{58.4} & 45.7 & 46.4 & 56.1 \\
Occ=Clear & 52.1 & 60.7 & 10.5 & \textbf{67.3} & 52.9 & 54.9 & 64.4 \\
Occ=Partial & 31.8 & 40.9 & 6.1 & \textbf{47.2} & 37.4 & 32.4 & 44.3 \\
Occ=Heavy & 9.9 & 7.3 & 0.3 & \textbf{16.8} & 11.3 & 9.0 & 14.3 \\
Scale=Large & 59.7 & \textbf{72.7} & 17.3 & 69.0 & 58.4 & 57.7 & 62.6 \\
Scale=Medium & 67.7 & 59.5 & 4.4 & \textbf{70.7} & 58.9 & 58.5 & 70.1 \\
Scale=Small & 15.7 & 30.3 & 1.7 & 47.6 & 29.8 & 32.6 & \textbf{48.5} \\
Pose=Typical & 48.8 & 59.5 & 10.2 & \textbf{62.7} & 51.4 & 49.1 & 60.1 \\
Pose=Unusual & 14.0 & 19.1 & 0.7 & \textbf{41.1} & 13.0 & 37.6 & 38.1 \\
Motion=Static & 40.5 & 53.0 & 10.5 & \textbf{56.4} & 44.9 & 43.6 & 55.0 \\
Motion=Moving & 57.9 & 60.0 & 7.8 & \textbf{70.4} & 55.2 & 59.0 & 65.7 \\
\hline
    \end{tabular}
    \caption{ADR (calculated as described in Section~\ref{sec:eval} for data subsets. The left section uses the standard training set, and the right section shows the effect of varying training data on \ours{}.}
    \label{tab:results}
\end{table*}

\begin{figure*}[th]
    \centering
    \begin{subfigure}[b]{0.15\textwidth}\includegraphics[width=\linewidth]{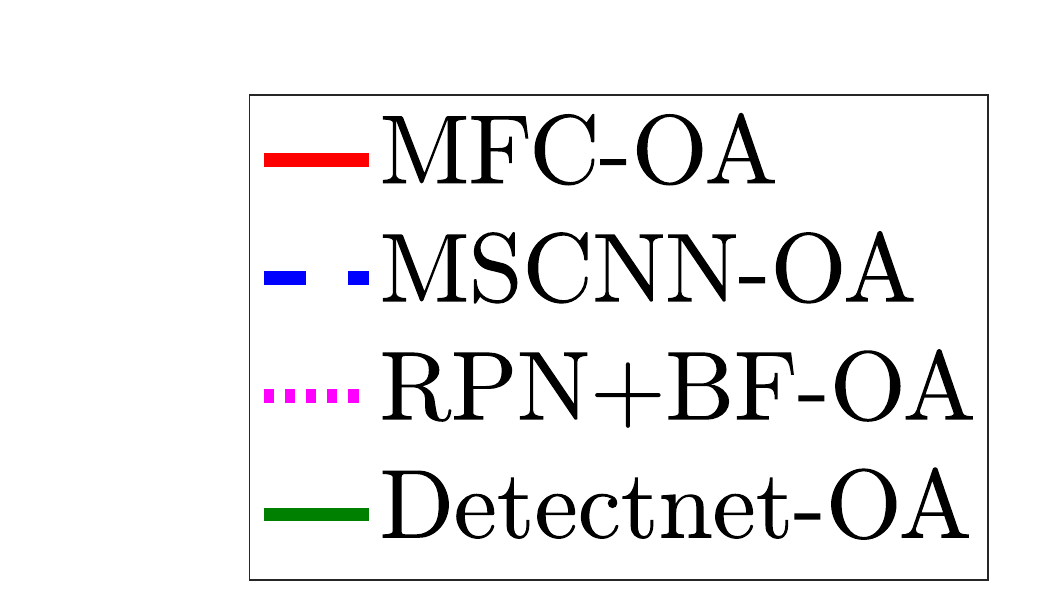}\vspace{1cm}\caption{Legend}\label{fig:results_legend}\end{subfigure}
    \begin{subfigure}[b]{0.225\textwidth}\includegraphics[width=\linewidth]{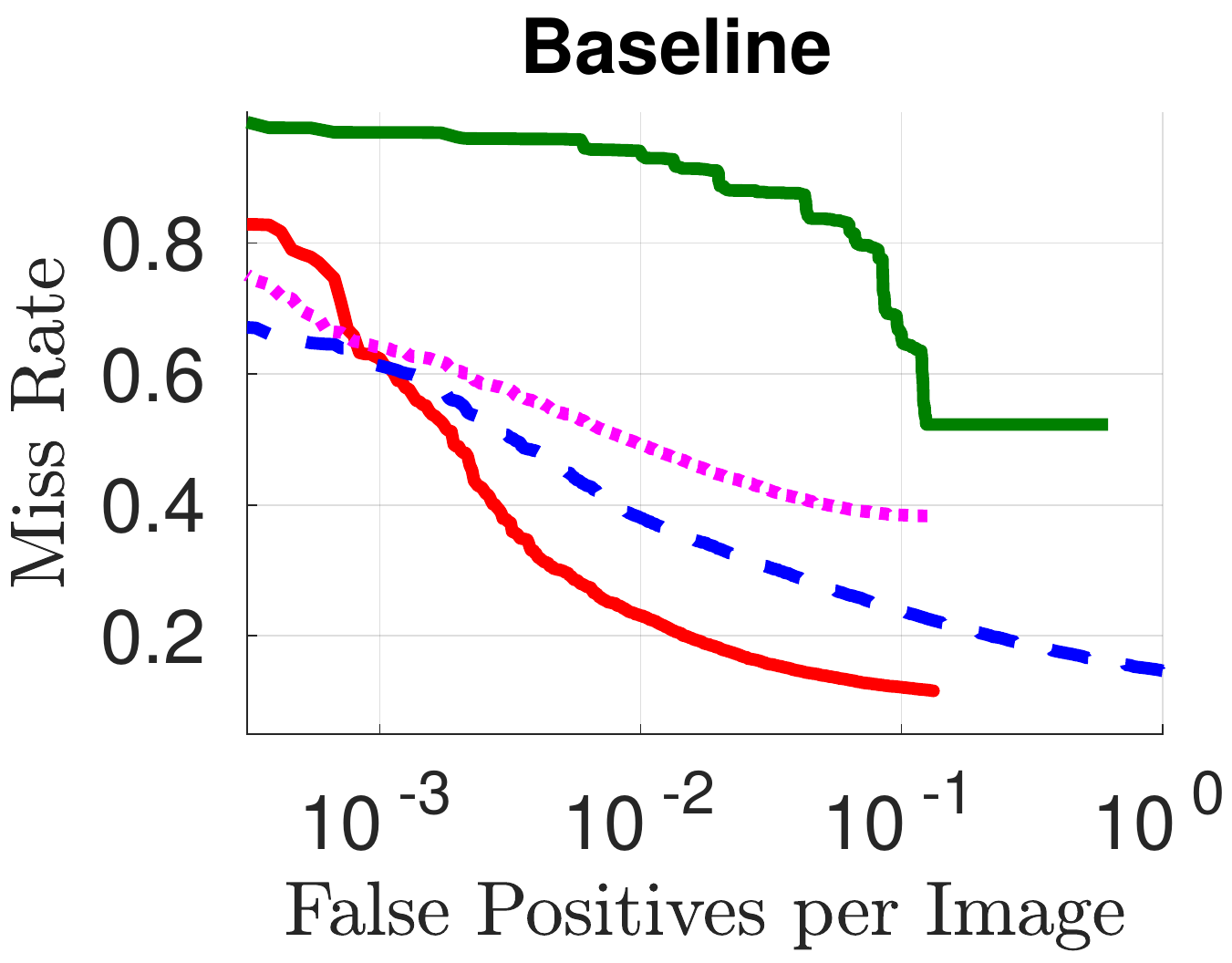}\caption{Test}\end{subfigure}
    \begin{subfigure}[b]{0.225\textwidth}\includegraphics[width=\linewidth]{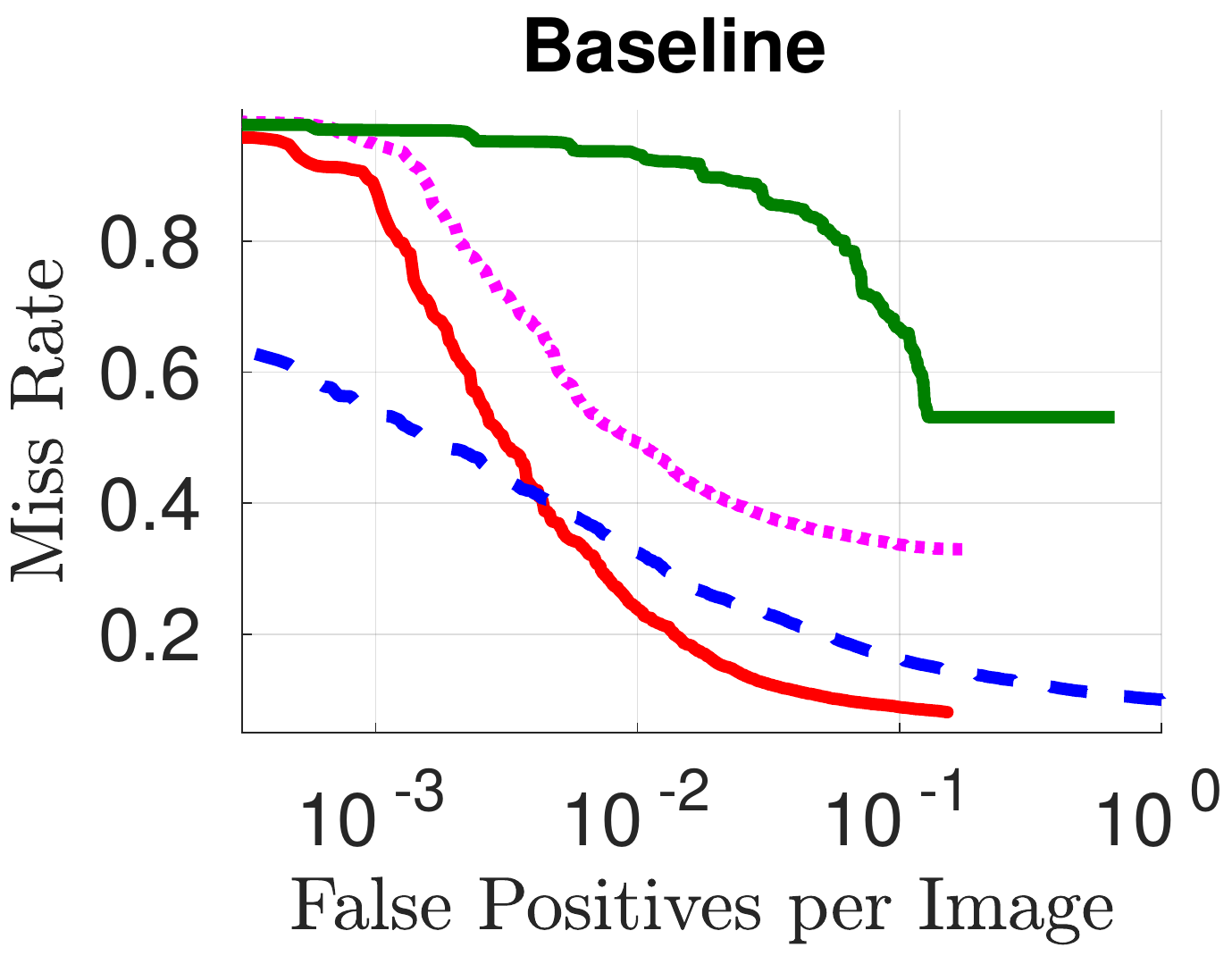}\caption{Validation}\end{subfigure}
    \caption{\Baseline{} performance on \revised{Test set vs. Validation set (IoU 0.5). General performance trends are the same on both sets, though there are some differences at the left end of the plots (where there is the smallest data support). All other figures show evaluations of only the Test set.} Legend applies to all following ROCs also.}
    \label{fig:results_val}
\end{figure*}

\begin{figure*}[th]
    \centering
    \begin{tabular}{ccc}
    \begin{subfigure}[b]{0.225\textwidth}\includegraphics[width=\linewidth]{figures/results/roc-test-exp1}\caption{}\label{fig:results_baseline}\end{subfigure}
    \begin{subfigure}[b]{0.225\textwidth}\includegraphics[width=\linewidth]{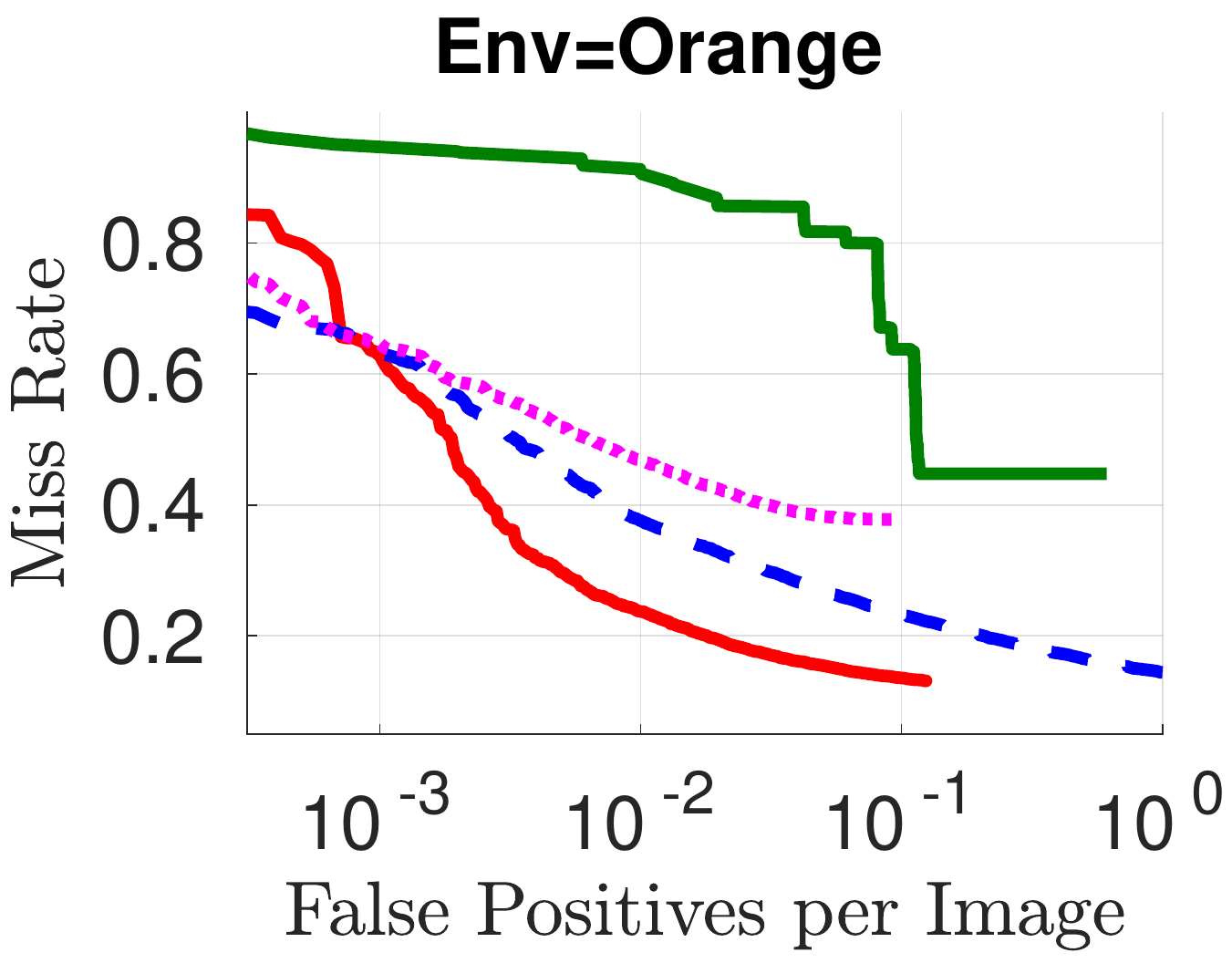}\caption{}\end{subfigure}
    \begin{subfigure}[b]{0.225\textwidth}\includegraphics[width=\linewidth]{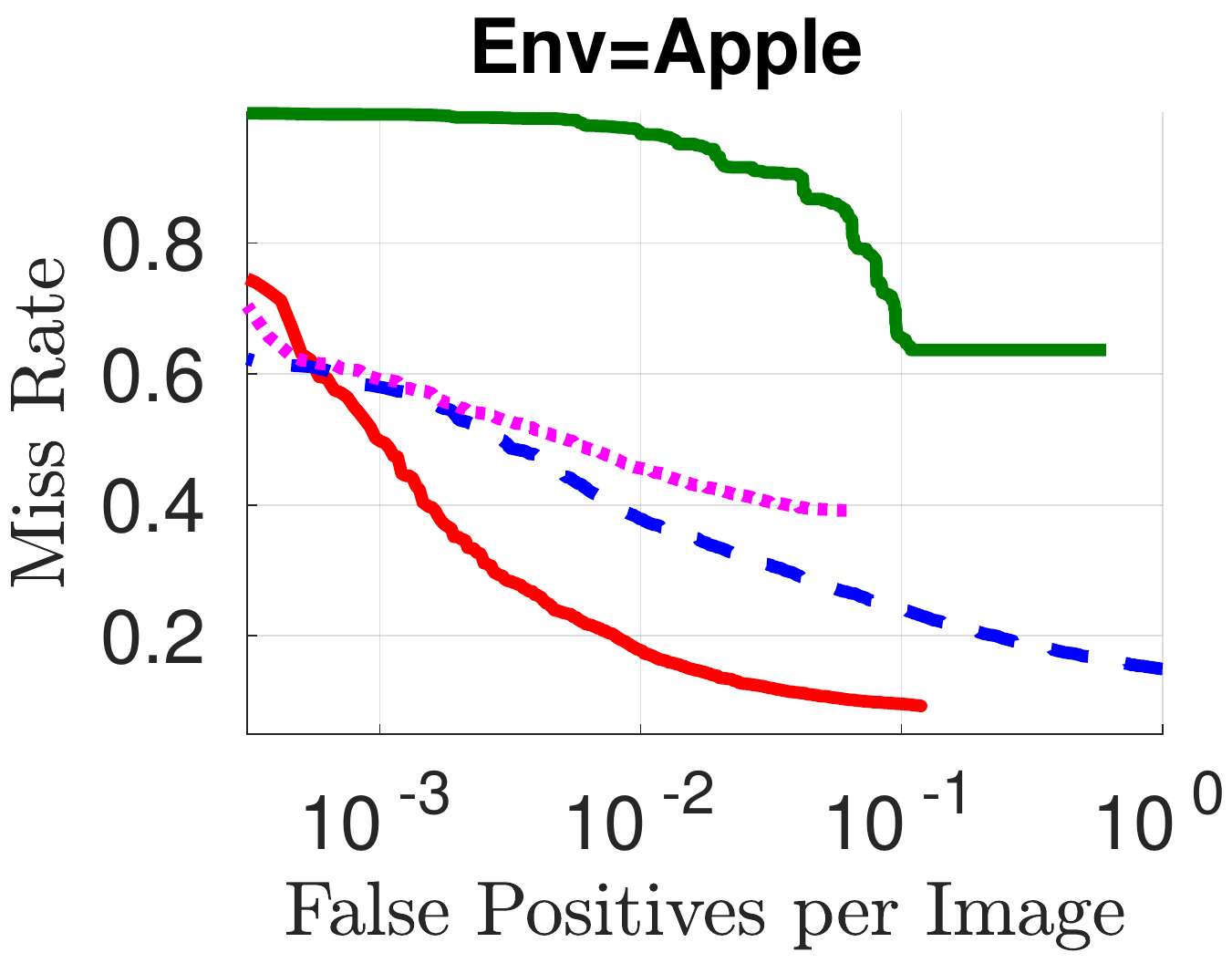}\caption{}\end{subfigure}
    \end{tabular}
    \caption{Performance per environment (IoU 0.5). \revised{(a) shows performance on the full \Baseline{} set, while (b) and (c) evaluate on subsets of the data from only the orange grove and the apple orchard respectively.} See Figure~\ref{fig:results_legend} for legend.}
    \label{fig:results_env}
\end{figure*}

\begin{figure*}[th]
    \centering
    \begin{tabular}{ccc}
    \begin{subfigure}[b]{0.225\textwidth}\includegraphics[width=\linewidth]{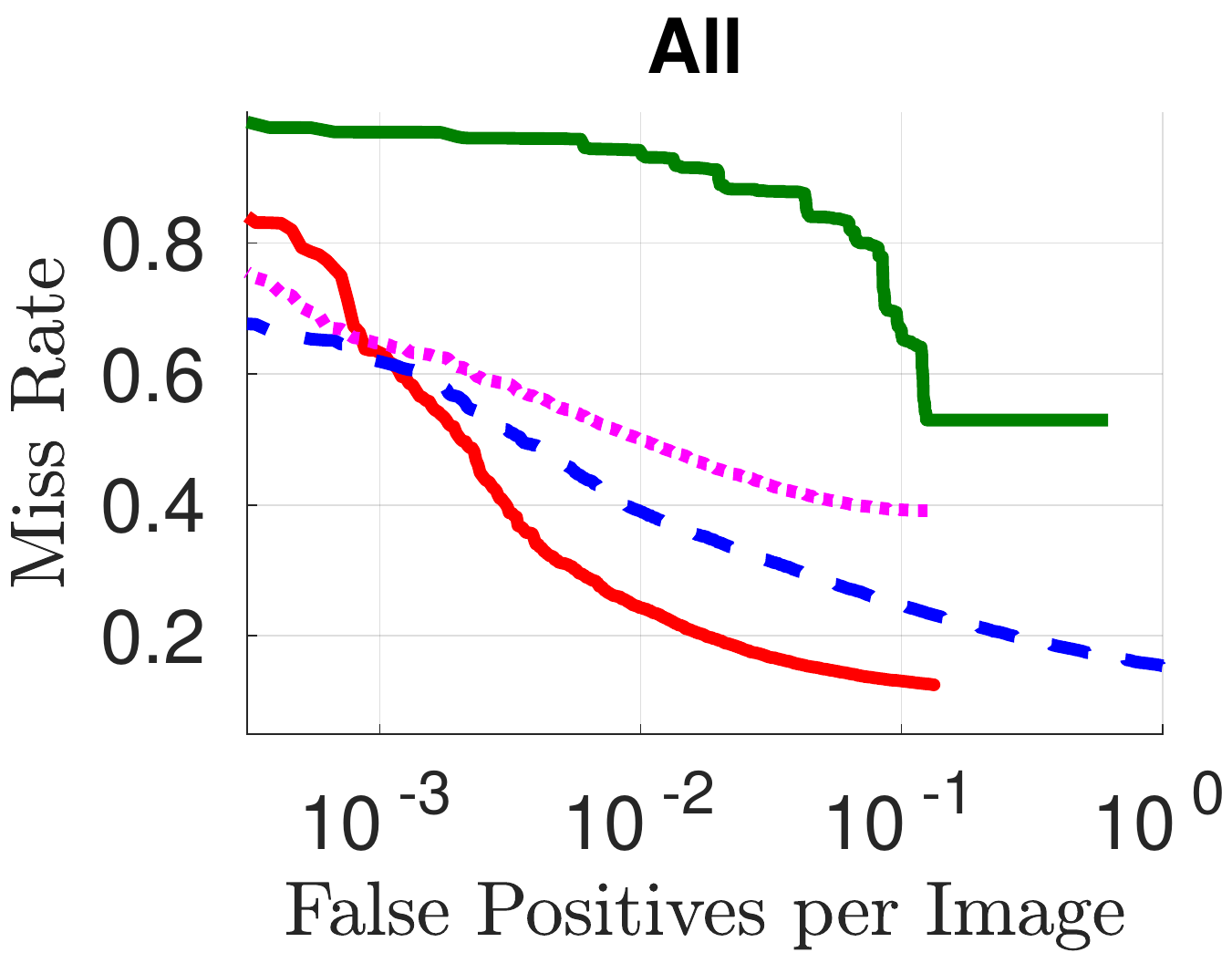}\caption{}\label{fig:resAll}\end{subfigure}
    \begin{subfigure}[b]{0.225\textwidth}\includegraphics[width=\linewidth]{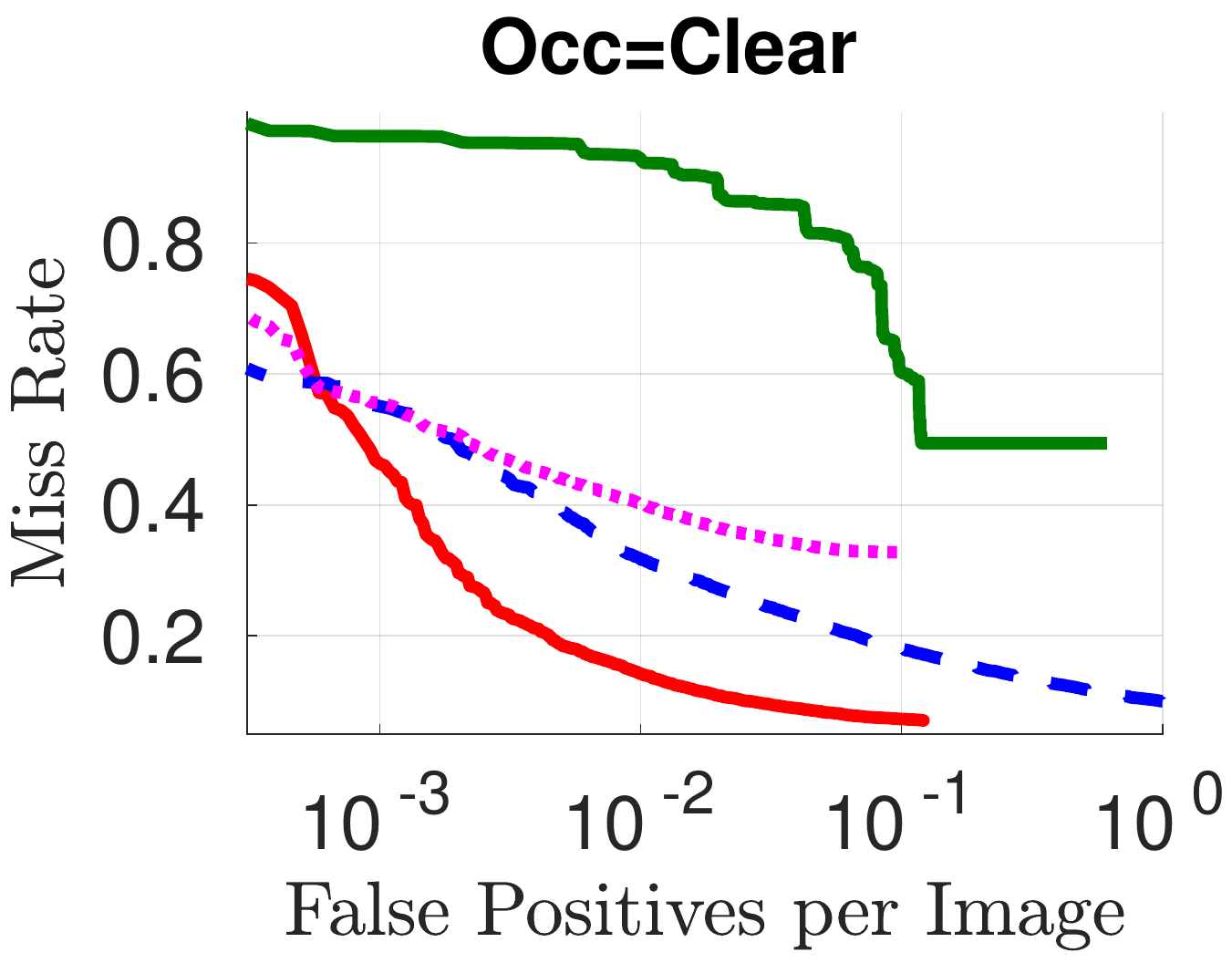}\caption{}\end{subfigure}
    \begin{subfigure}[b]{0.225\textwidth}\includegraphics[width=\linewidth]{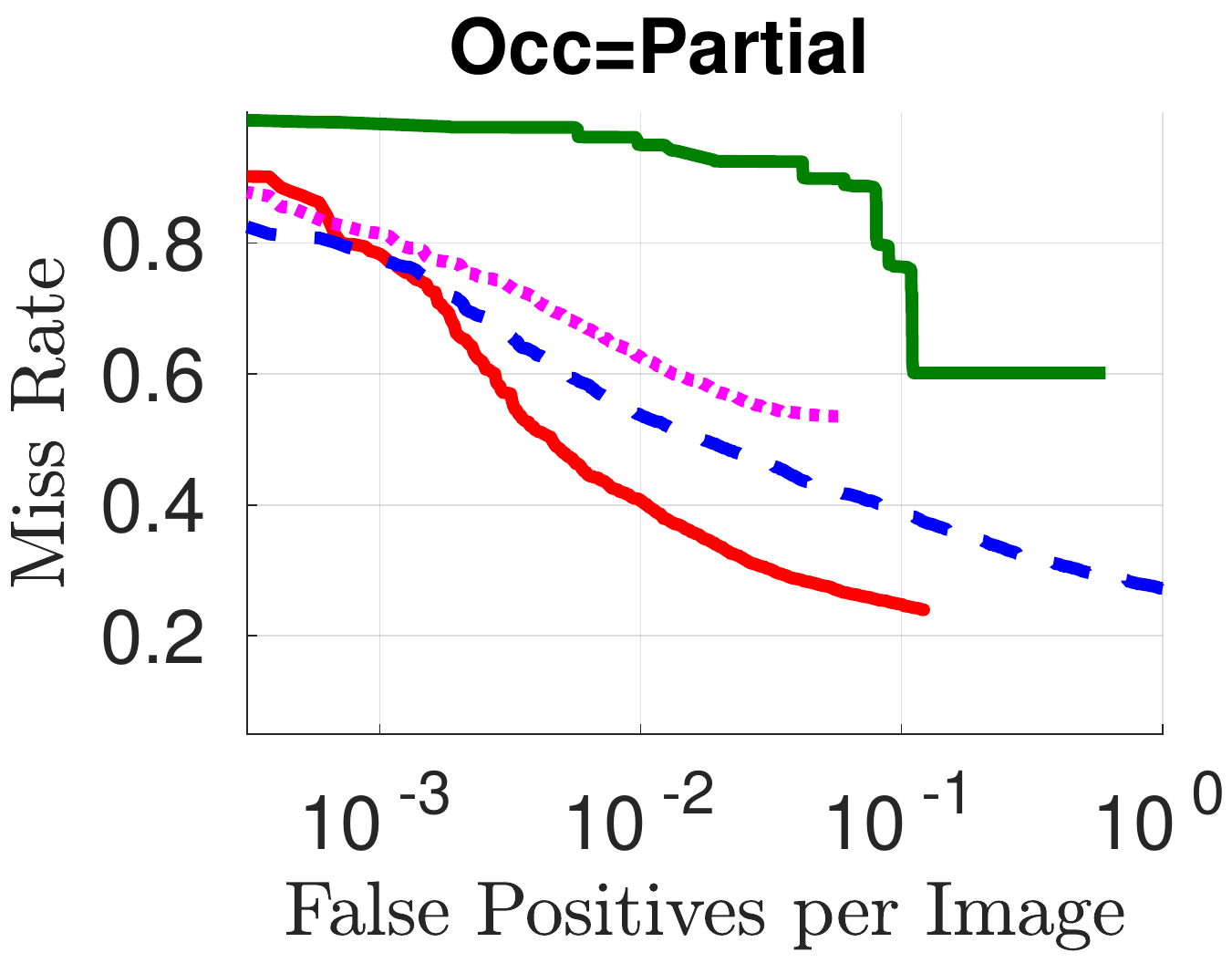}\caption{}\end{subfigure}
    \begin{subfigure}[b]{0.225\textwidth}\includegraphics[width=\linewidth]{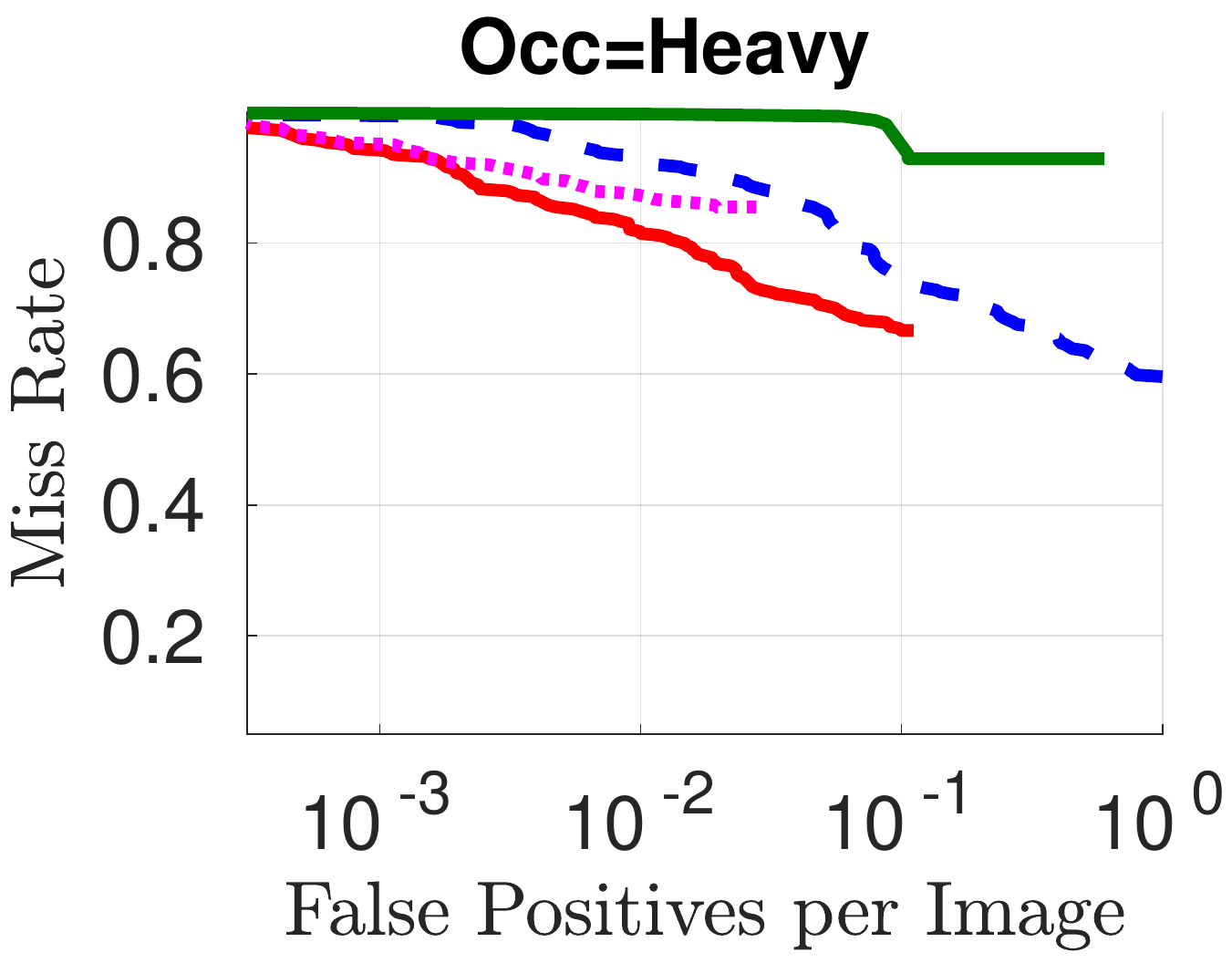}\caption{}\label{fig:resHeavy}\end{subfigure}
    \end{tabular}
    \caption{Performance on different occlusion levels (IoU 0.5). Figure~\ref{fig:resAll} (All) and Figure~\ref{fig:resHeavy} (Occ=Heavy) include labels that are ignored in \Baseline{} and all other results. See Figure~\ref{fig:results_legend} for legend.}
    \label{fig:results_occ}
\end{figure*}

\begin{figure*}[th]
    \centering
    \begin{tabular}{ccc}
    \begin{subfigure}[b]{0.225\textwidth}\includegraphics[width=\linewidth]{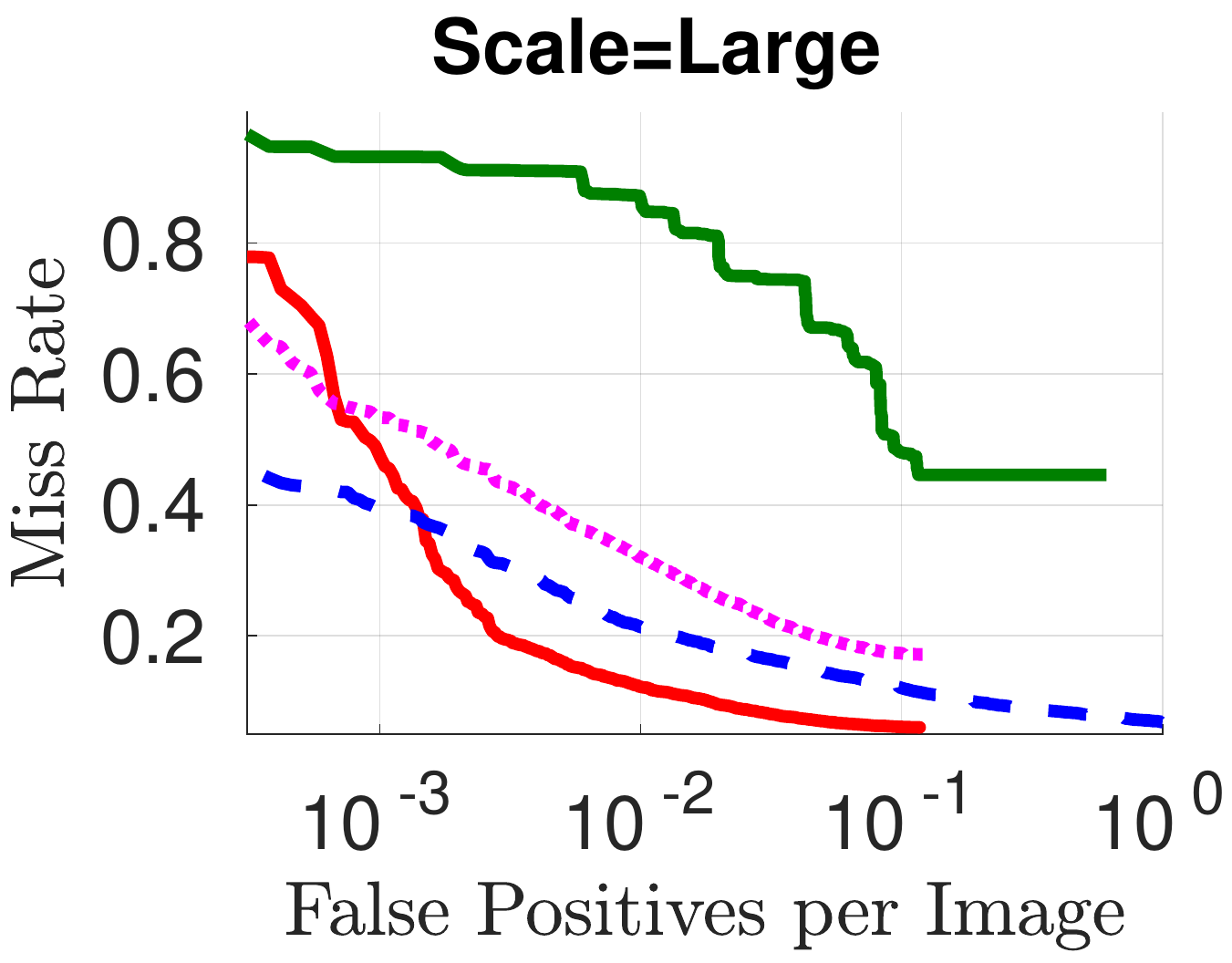}\caption{}\end{subfigure}
    \begin{subfigure}[b]{0.225\textwidth}\includegraphics[width=\linewidth]{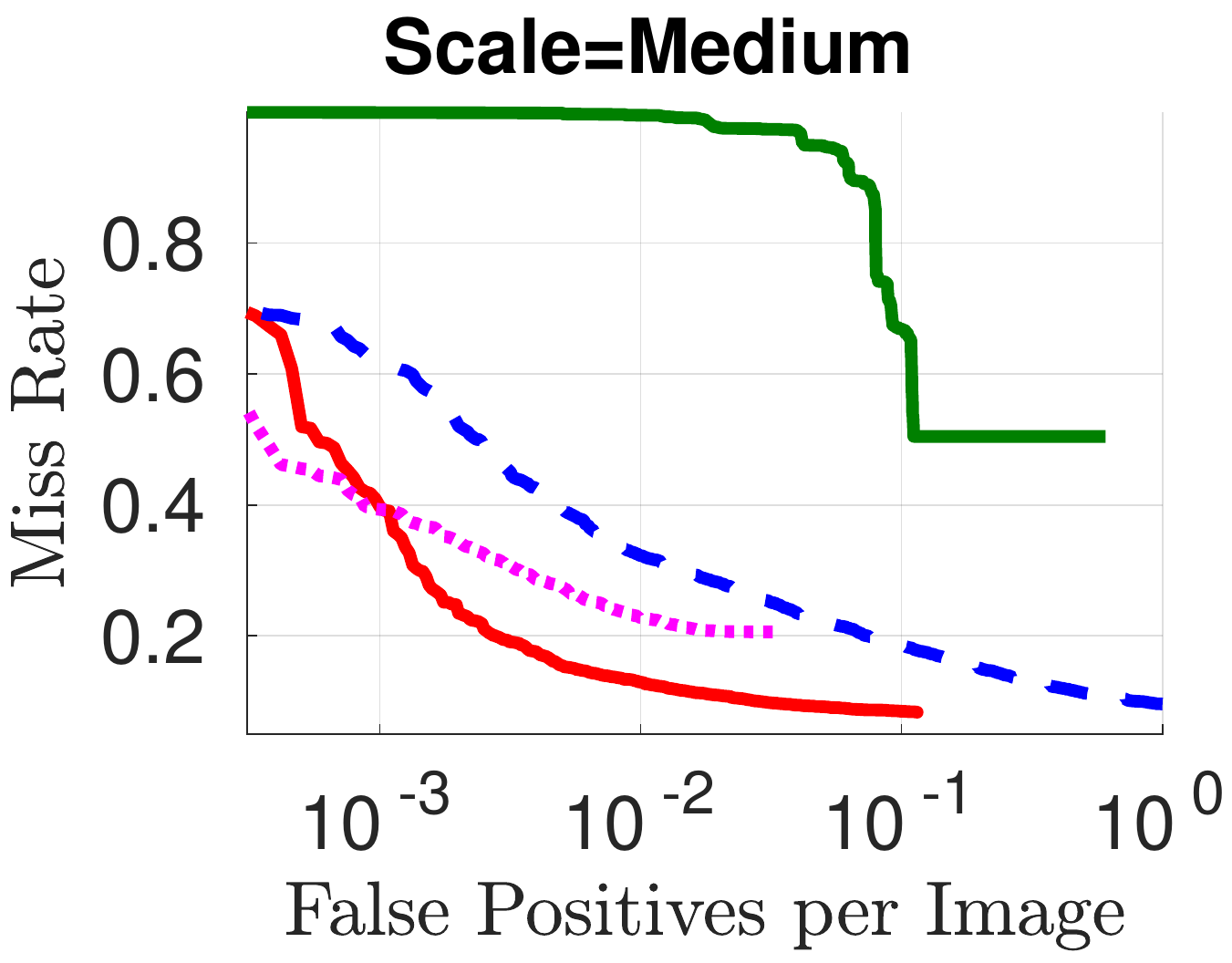}\caption{}\end{subfigure}
    \begin{subfigure}[b]{0.225\textwidth}\includegraphics[width=\linewidth]{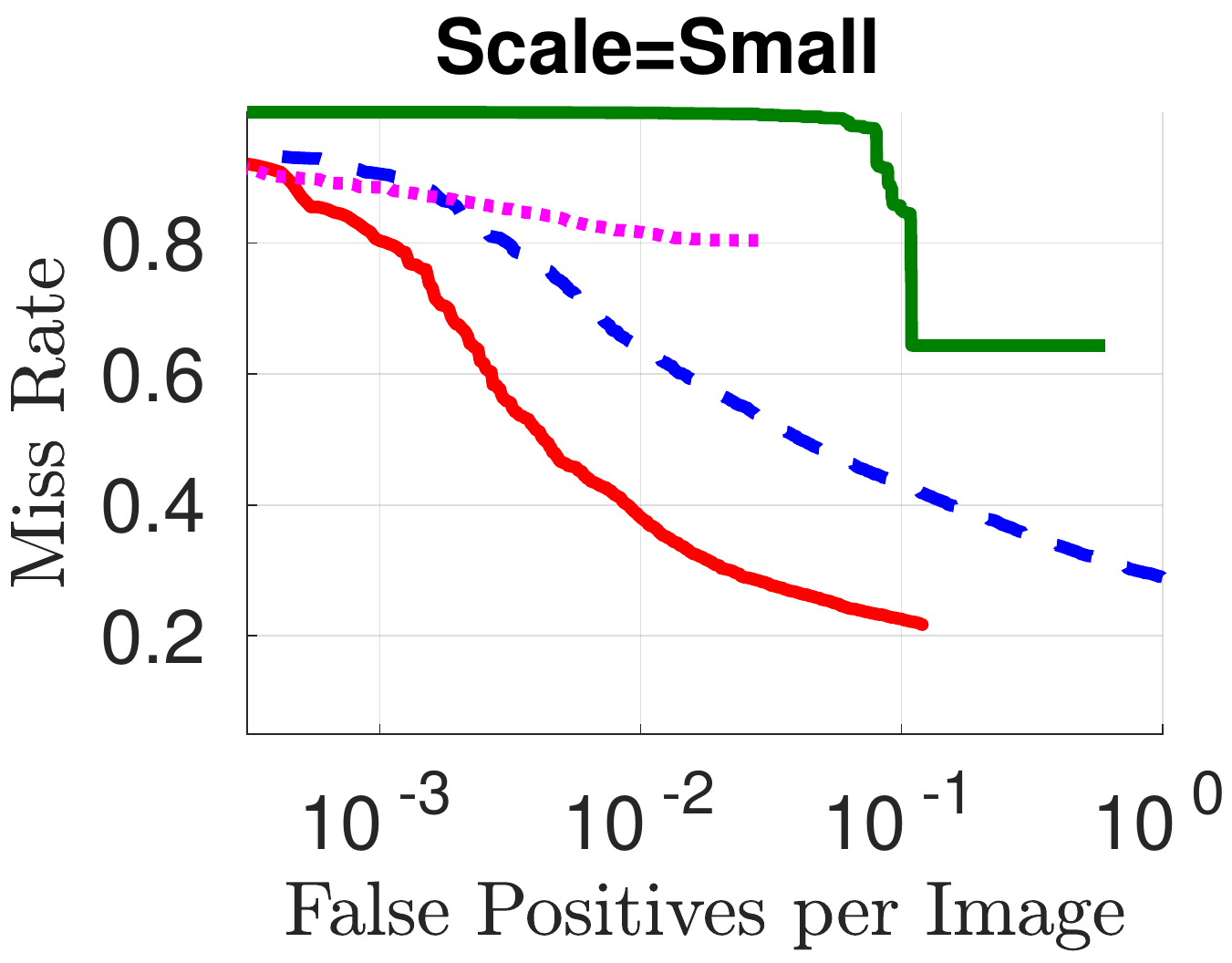}\caption{}\end{subfigure}
    \end{tabular}
    \caption{Performance across scales (IoU 0.5). \revised{Here we show performance on subsets of \Baseline{} where bounding boxes have area larger than 3500 pixels, smaller than 1300 pixels, or in between.} See Figure~\ref{fig:results_legend} for legend.}
    \label{fig:results_scale}
\end{figure*}

\begin{figure}[th]
    \centering
    \begin{tabular}{ccc}
    \begin{subfigure}[b]{0.225\textwidth}\includegraphics[width=\linewidth]{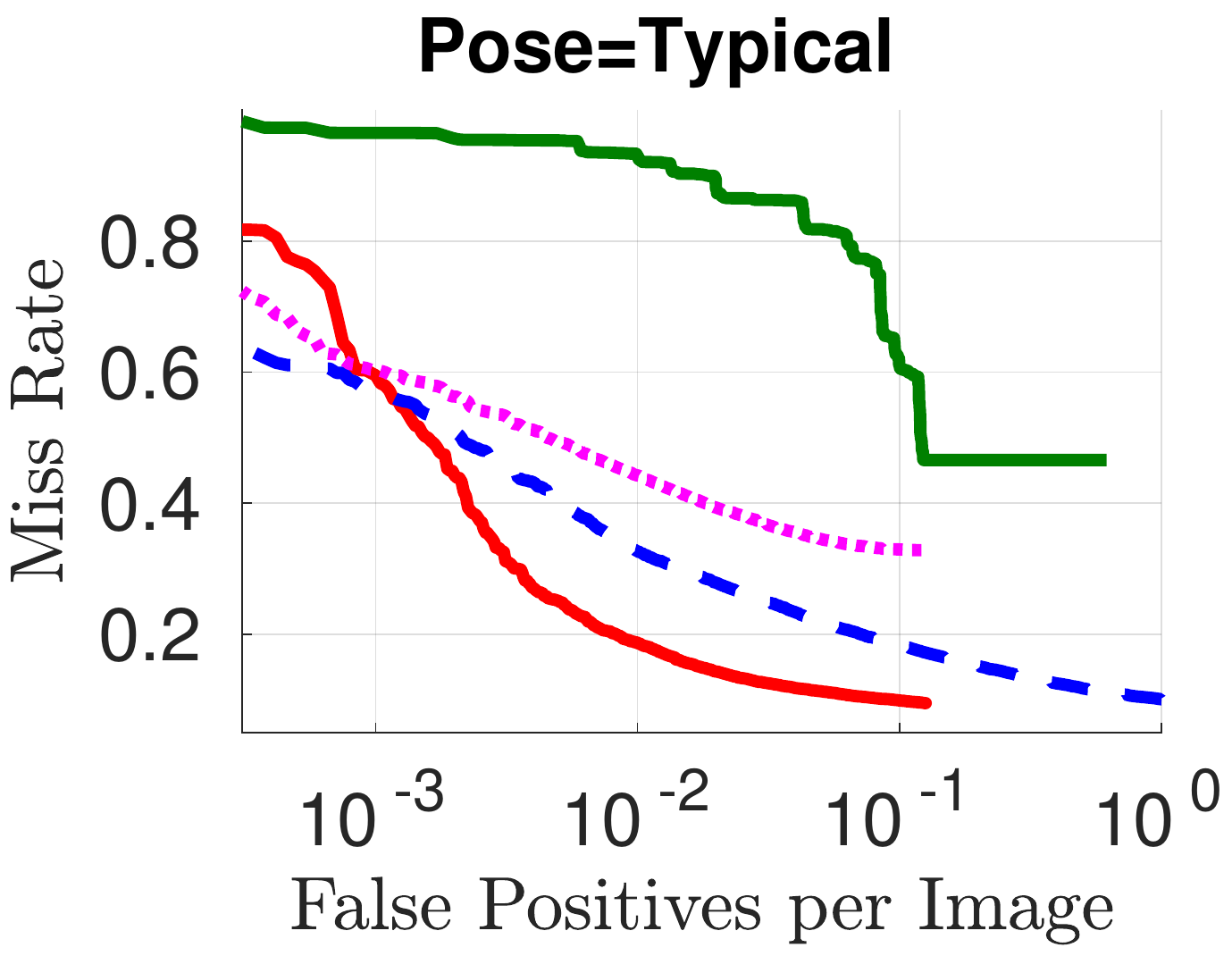}\caption{}\end{subfigure}
    \begin{subfigure}[b]{0.225\textwidth}\includegraphics[width=\linewidth]{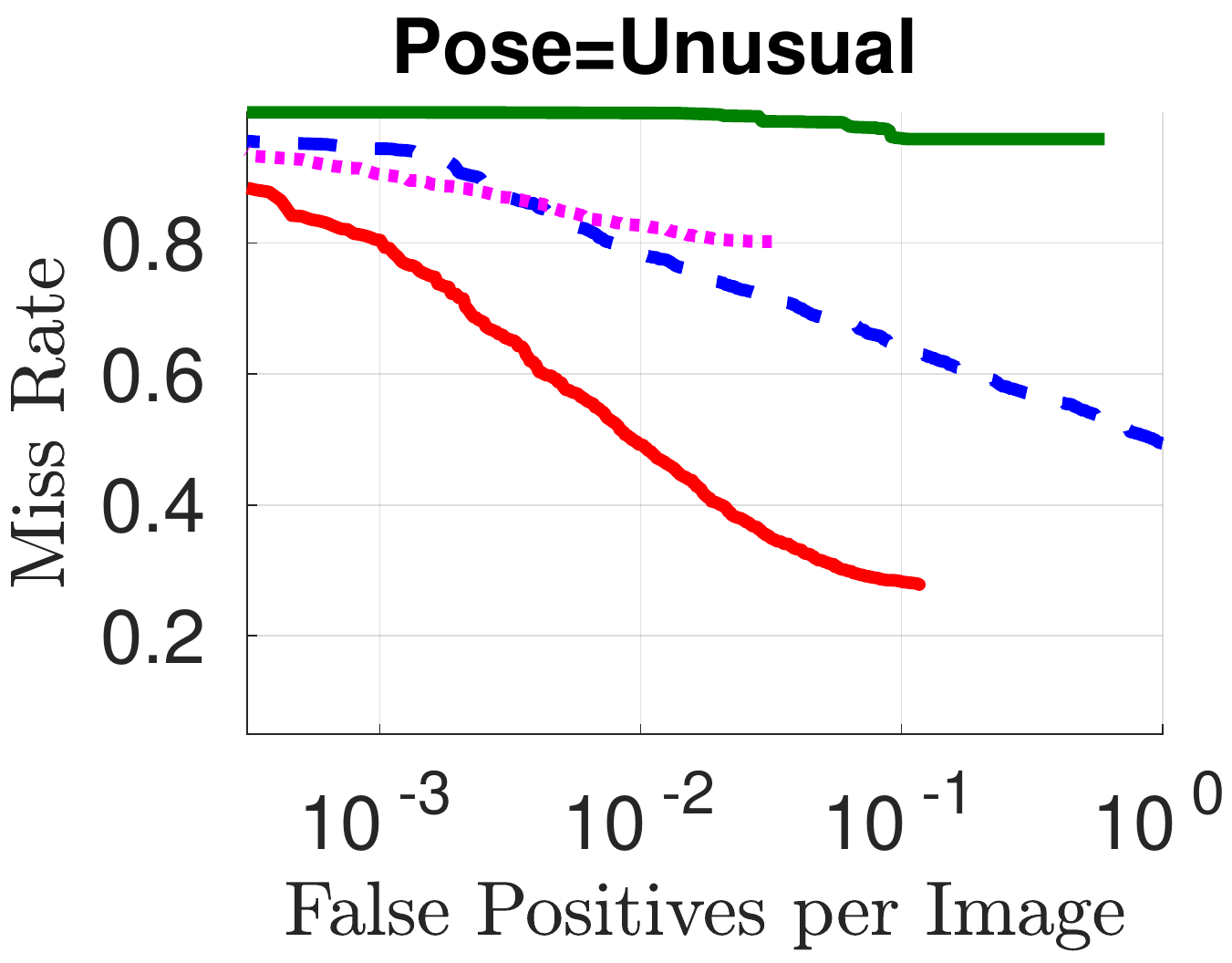}\caption{}\end{subfigure}
    \end{tabular}
    \caption{Performance on different poses (IoU 0.5). \revised{Unusual pose data consists of logs of people falling, lying down, or climbing on ladders, as described in Section~\ref{sec:collect_unusual_pose}. Typical poses comprise the rest of \Baseline{}.} See Figure~\ref{fig:results_legend} for legend.}
    \label{fig:results_pose}
\end{figure}

\begin{figure}[th]
    \centering   
    \begin{subfigure}[b]{0.225\textwidth}\includegraphics[width=\linewidth]{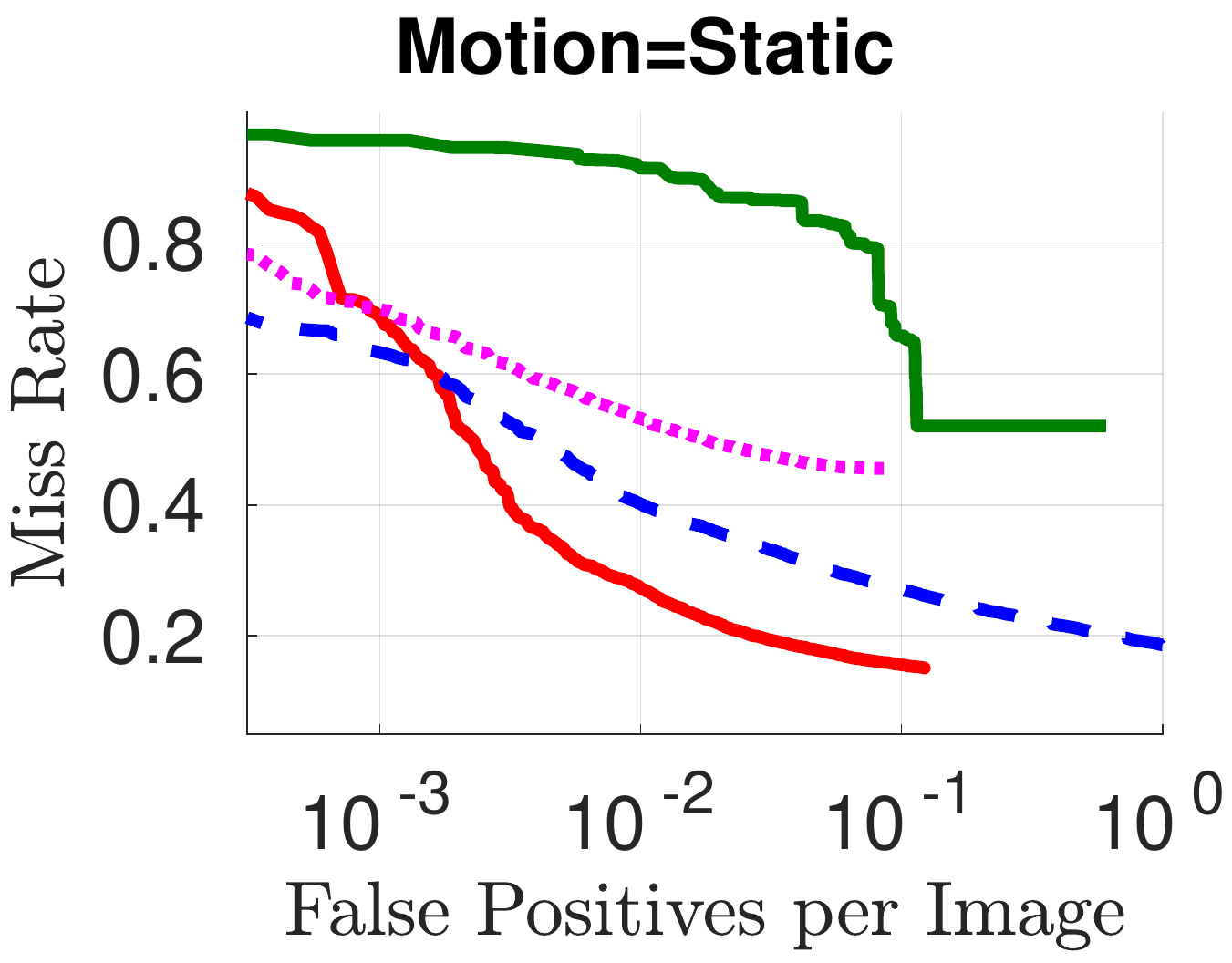}\caption{}\end{subfigure}
    \begin{subfigure}[b]{0.225\textwidth}\includegraphics[width=\linewidth]{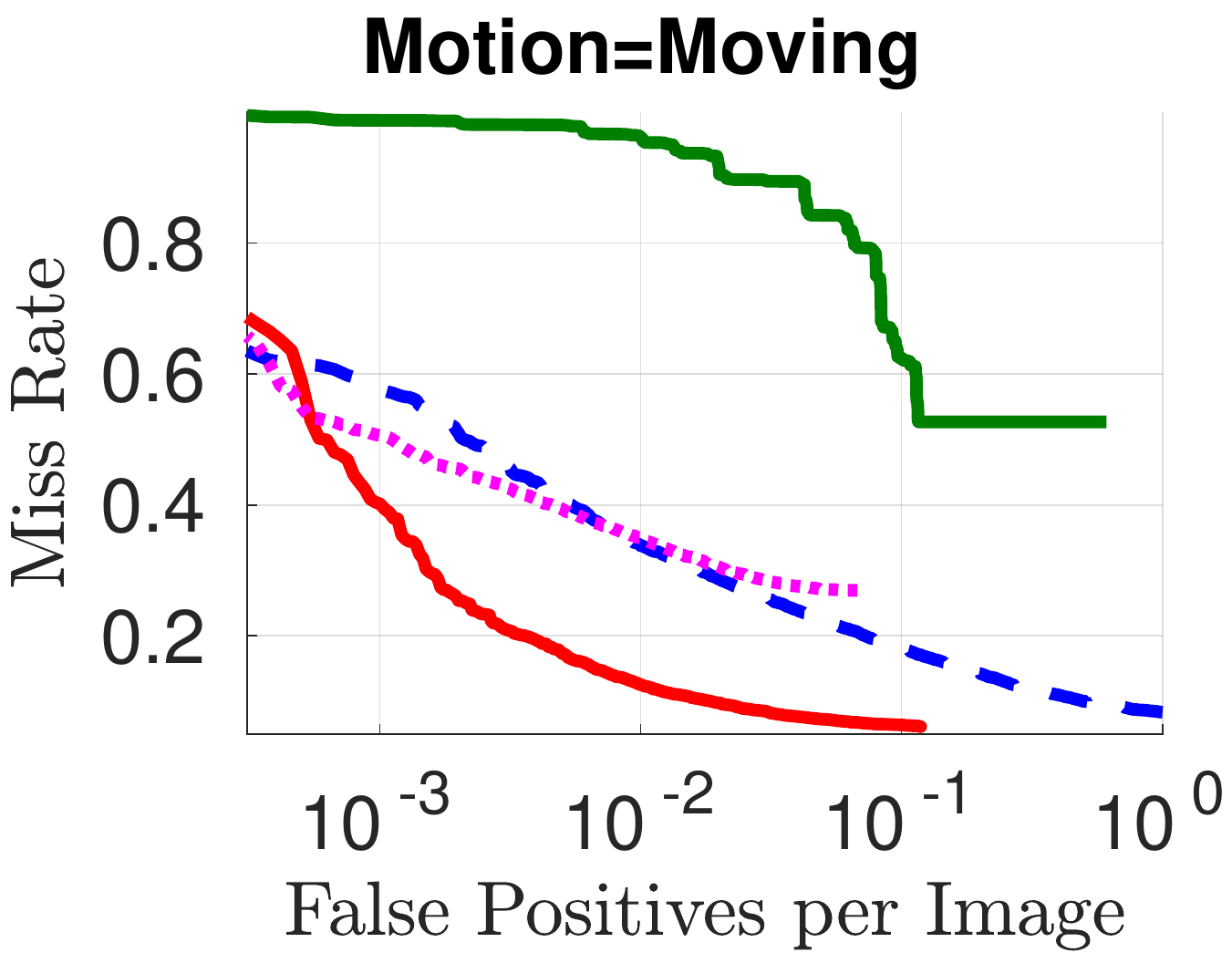}\caption{}\end{subfigure}
    \caption{Performance with and without motion (IoU 0.5). \revised{Moving logs show the person in motion, as described in Section~\ref{sec:collect_moving_person}.} See Figure~\ref{fig:results_legend} for legend.}
    \label{fig:results_motion}
\end{figure}

We evaluated our approach against \revised{\nApproachesText{} leading} CNN approaches from the urban pedestrian detection literature on the benchmark, broken down into all of the subsets defined in Section~\ref{sec:filtering}. 
RPN+BF~\cite{zhang2016faster} is the current leader on the Caltech benchmark, and was adapted to this dataset by using the same size and aspect ratio templates used in our approach.  
MS-CNN~\cite{cai2016unified} is the current leading method with available code on the KITTI benchmark, and it was adapted only for the differing image sizes in this dataset.
DetectNet~\cite{detectNetBlog} builds upon the ILSVRC 2014 winner, GoogLeNet. 
As described in Section~\ref{sec:eval}, ADR uses an average of IoUs in the range 0.3 to 0.7. The performance of all evaluated algorithms is shown in Table~\ref{tab:results}, and a nearly hour-long video of test set detections from the $Std$ training set of each algorithm is available online\footnote{https://www.youtube.com/watch?v=YNRY4y3vfFA}.

We summarize high-level conclusions here, but delve into diagnostic details with performance curves below. Apple and Orange appear similar in difficulty for all methods. Training on either dataset alone seems to result in overfitting (implied by poor cross-dataset performance), but training on both training sets noticeably improves performance on both test environments. This also suggests that performance would continue to increase with additional training data. \revised{DetectNet significantly under-performs the other methods, and in all experiments it reaches a minimum miss rate, where decreasing sensitivity further only adds more false positives.} The performance of the state-of-the-art (\RPNBF{}) struggles with heavy oclusions (13\%) and small pedestrians (18\%). Our model significantly improves accuracy in these regimes, doubling accuracy on heavy occlusions (26\%) and dramatically improving performance on small pedestrians (66\%). Unusual poses are significantly more challenging for the state-of-the-art (14\%), whereas we perform considerably better (41\%). Finally, an interesting observation is that moving pedestrians are easier for all detectors.

Figure~\ref{fig:results_val} shows a comparison of performance on the test set to the validation set. \revised{We took care to ensure that both the validation and the test set were as independent of the training set as possible, yet still had compositions such that validation performance would be predictive of test performance, as described in Section~\ref{sec:traintest}. Because of this independence, there is inevitably some difference in difficulty between test and validation, such that one cannot expect to get the same ADR scores on each. Nonetheless, qualitative trends should hold across these sets, and the general ranking of approaches is indeed consistent between the two. Note also that some algorithm meta-parameters were tuned on the validation set, so (setting aside differences in difficulty) one can expect a somewhat stronger performance there than on the test set that was never seen in the development process.}

We compare apples and oranges in Figure~\ref{fig:results_env}. As expected, algorithms trained only on one environment do well on that environment and perform more poorly on the other. Training on both apple and orange data consistently gives the best results.

The breakdown of occlusion categories is shown in Figure~\ref{fig:results_occ}. As expected, performance degrades as occlusion increases, to the point that performance is dramatically worse on heavy-occlusion cases. These cases are very difficult even for humans though, which is why they are omitted from every other subset evaluation (except All).

The trend across different scales in Figure~\ref{fig:results_scale} is slightly different. All approaches perform worst on small scales, though some are more robust than others; \RPNBF{} struggles particularly here. Most approaches perform slightly better on Medium than Large scale though. Although there are more examples of Large scale represented (see Figure~\revised{\ref{fig:subsets}}), these span a much larger range of scales that is likely more difficult to cover.

One additional effect of the large context windows used by \ours{} is that the receptive field more often extends noticeably beyond the edge of the image. This has implications for both occlusions and small people. Occlusions that are due to truncation by the edge of the image will have contribution from the padded regions outside the image, which could be interpreted as known occlusion flags. For the camera configuration in this dataset (and most of the prior work), small people tend to be near the top of the image (See Figure~\ref{fig:label_scales}). Since including more context will cause the template to extend into this padded region, it becomes possible to learn this prior.

Unusual poses, shown in Figure~\ref{fig:results_pose}, also offer a test of robustness. Once again, all approaches perform better on typical than unusual poses, but the drop in performance is much larger for some than others. \RPNBF{} struggles particularly here as well.

Finally, Figure~\ref{fig:results_motion} breaks down moving and stationary people. Performance is generally stronger on Moving than on Static people. Although Moving people exhibit a greater variety of poses (which might be challenging), they are also less often occluded and tend to spend less time at great distances. None of these methods make explicit use of the motion cues available from the continuous nature of the video sequences in the dataset though, so there is opportunity for further improvement here.

\subsection{Generalization Across Domains}\label{sec:generalization}

\begin{figure}[th]
    \centering
    \includegraphics[width=0.45\textwidth]{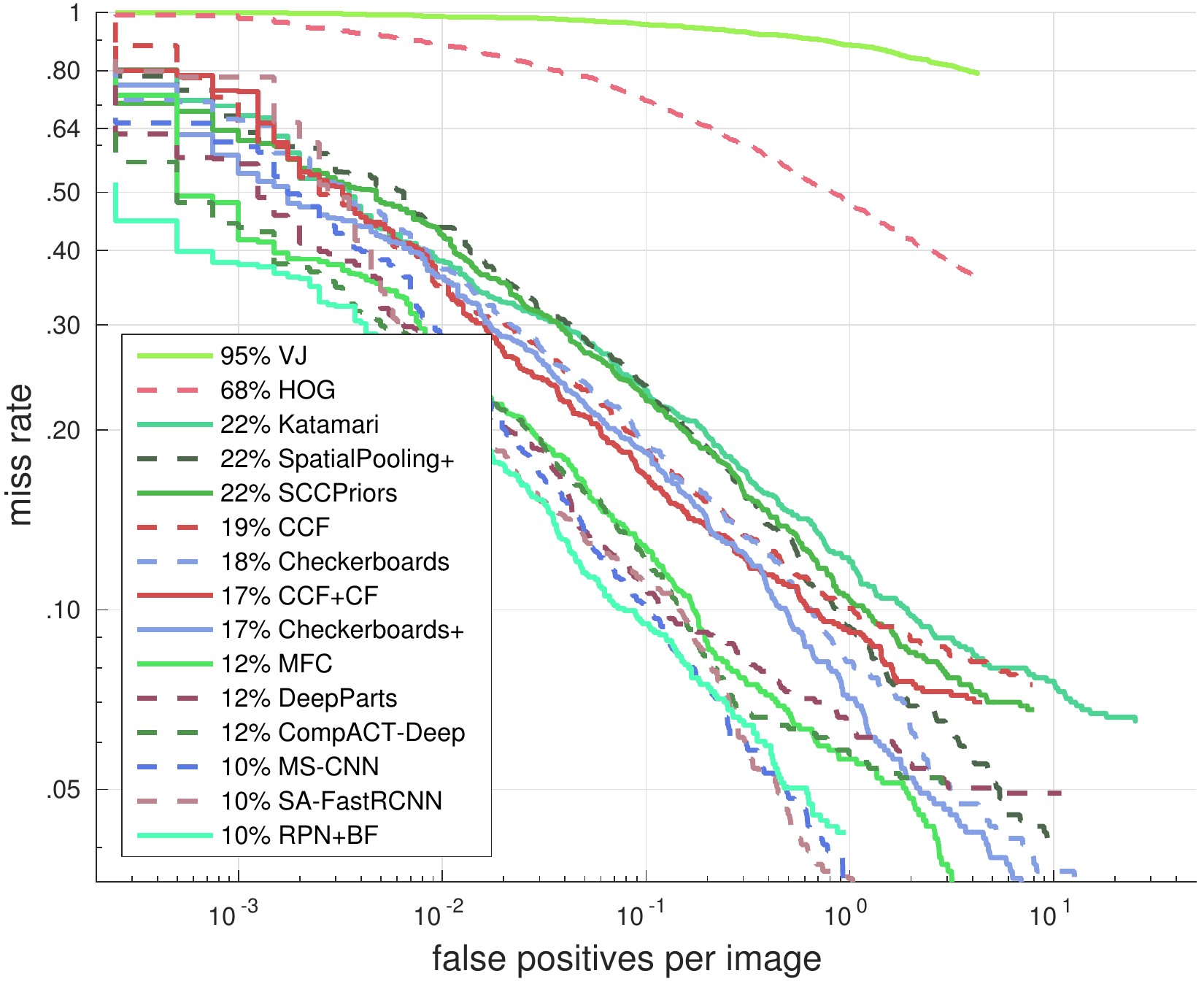}
    \caption{Performance on Caltech dataset ``Reasonable'' benchmark when training and evaluating on bounding boxes of the full extent of the person, compared to current leaderboard in the ``Caltech+ImageNet'' training scheme.}
    \label{fig:caltech-full}
\end{figure}

\begin{figure}[th]
    \centering
    \includegraphics[width=0.225\textwidth]{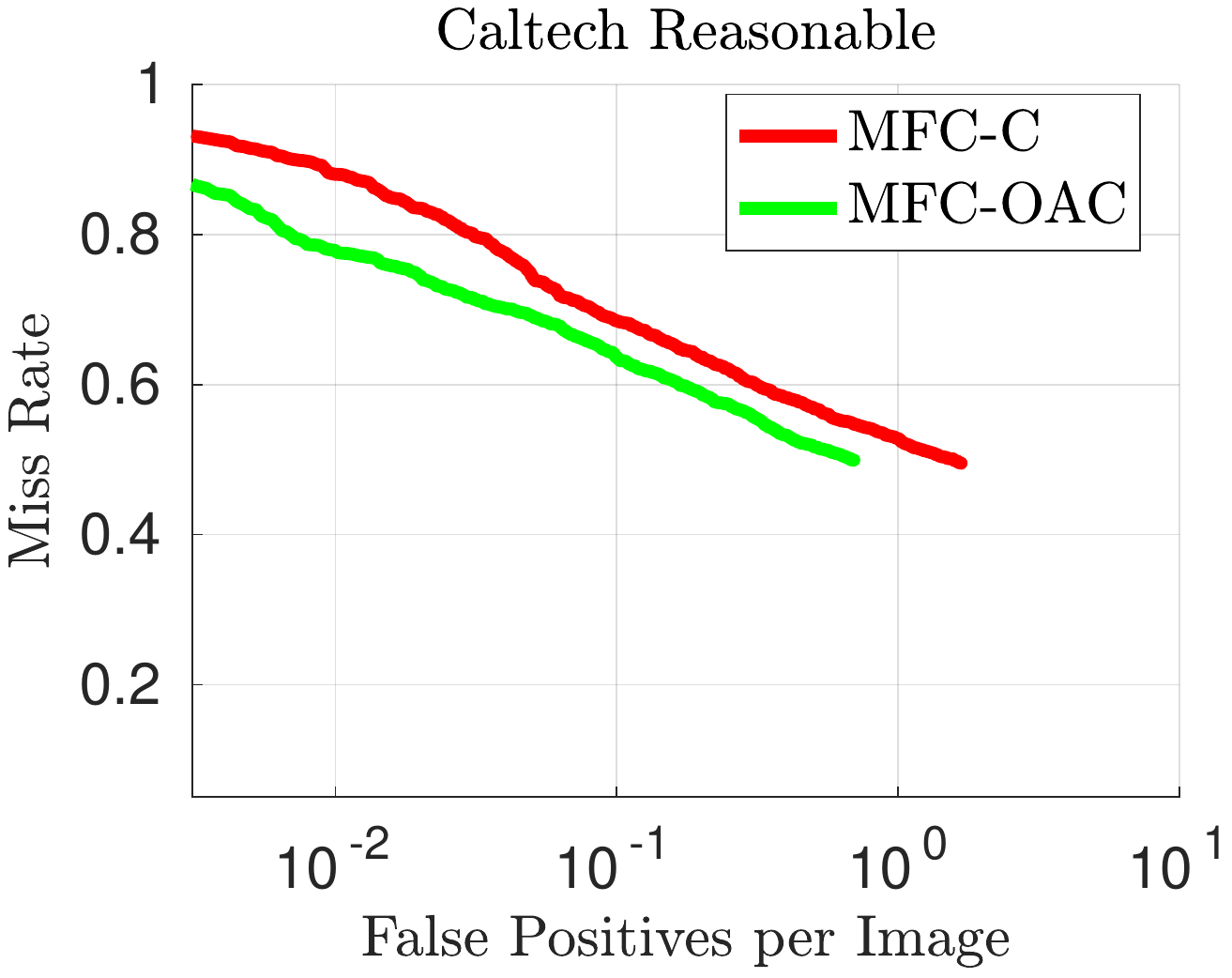}
    \caption{Performance on Caltech dataset ``Reasonable'' benchmark when training and evaluating on bounding boxes of the visible portion of the person.}
    \label{fig:caltech-vis}
\end{figure}

\begin{figure}[th]
    \centering
    \begin{subfigure}[b]{0.225\textwidth}                             \includegraphics[width=\linewidth]{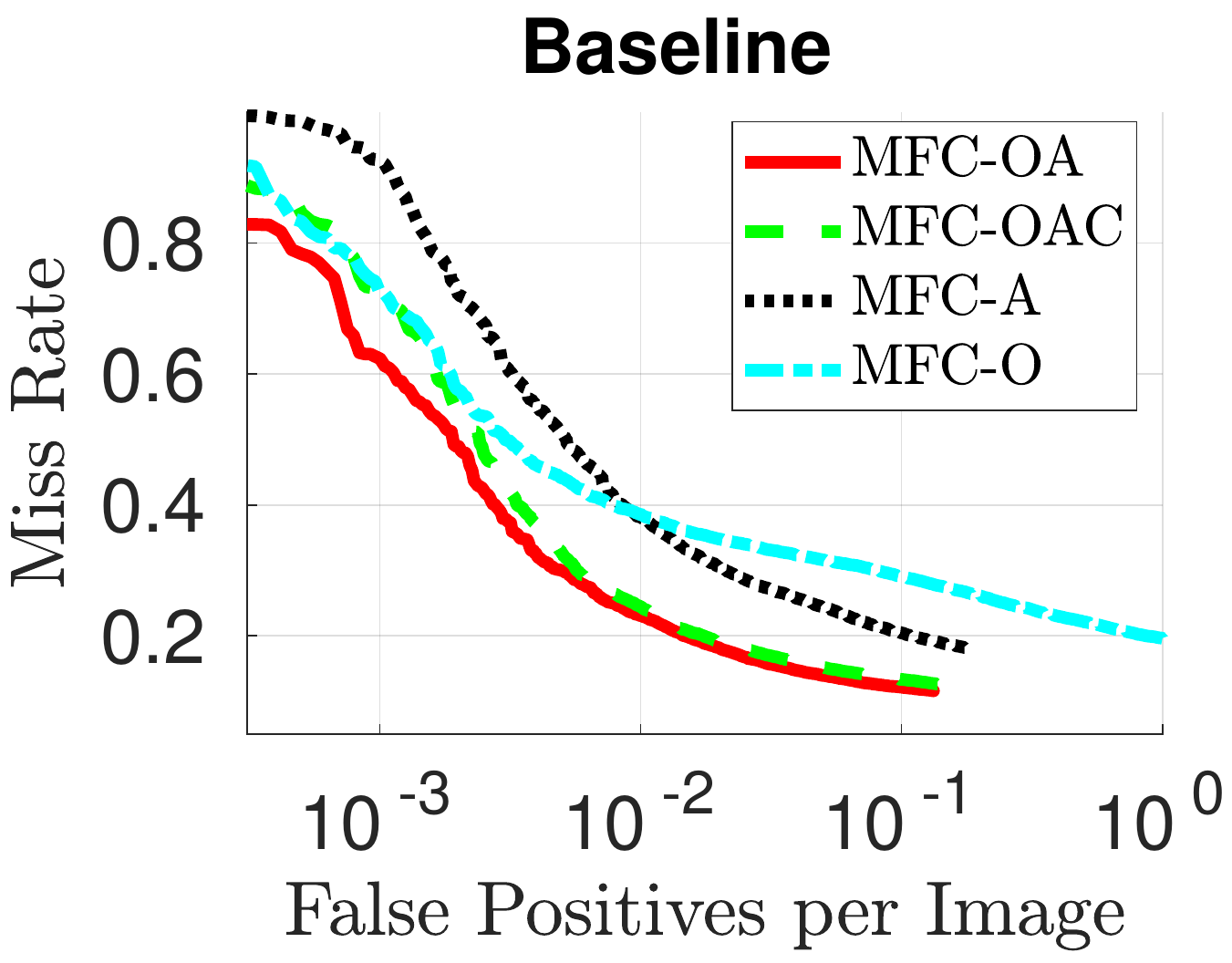}
        \caption{}\label{fig:varyTrain-baseline}
    \end{subfigure}
    \begin{subfigure}[b]{0.225\textwidth}
        \includegraphics[width=\linewidth]{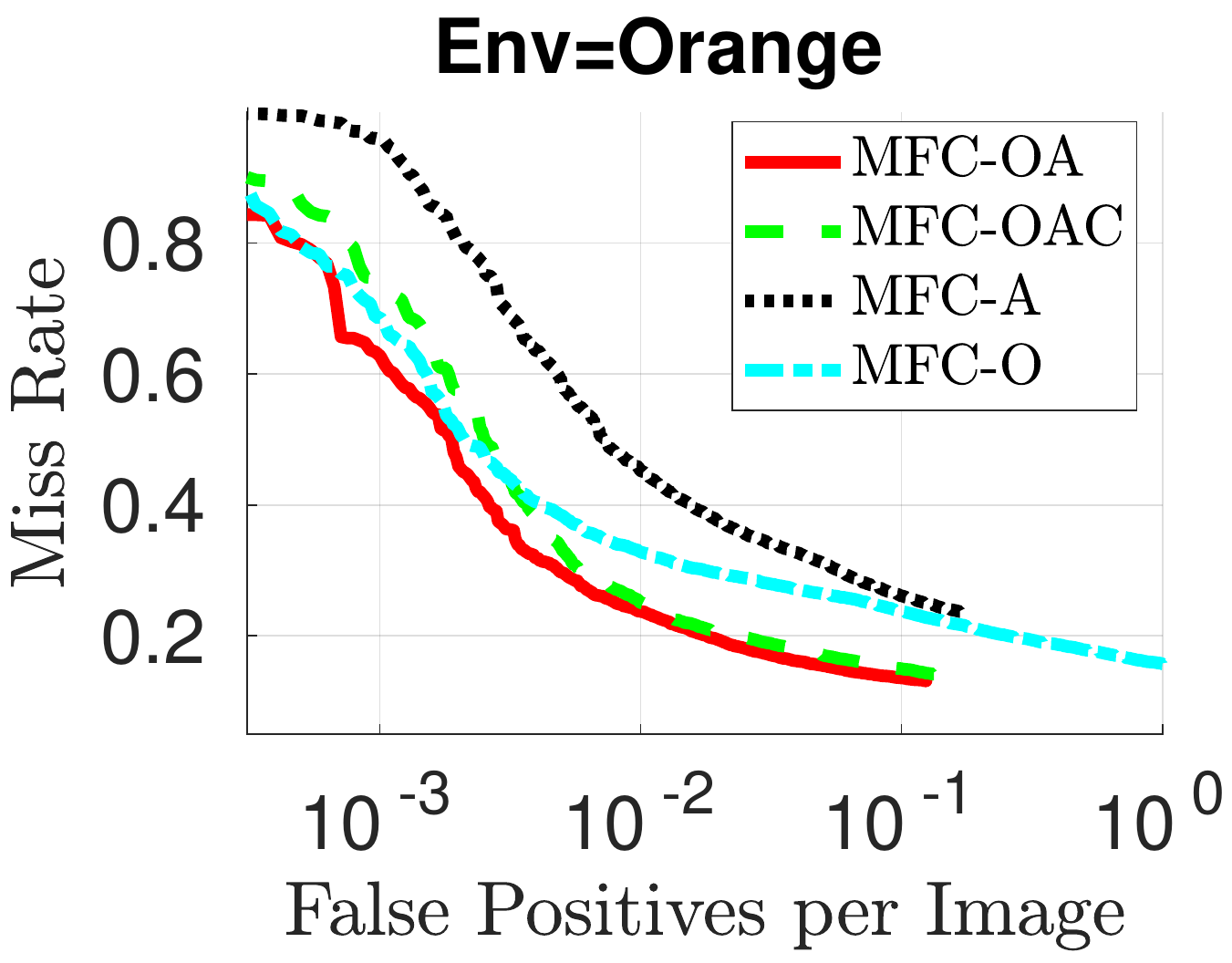}
        \caption{}\label{fig:varyTrain-orange}
    \end{subfigure}
    \begin{subfigure}[b]{0.225\textwidth}
        \includegraphics[width=\linewidth]{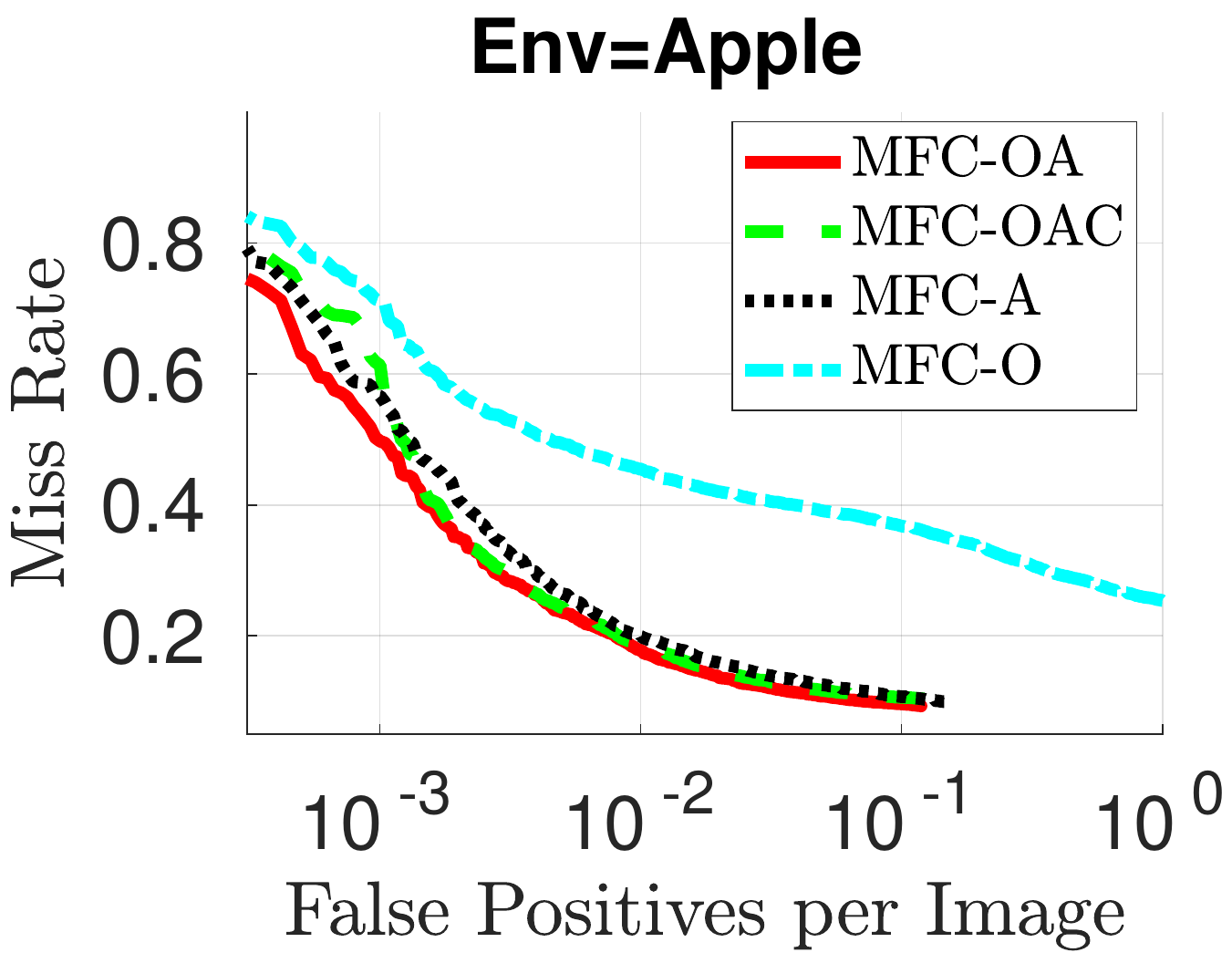}
        \caption{}\label{fig:varyTrain-apple}
    \end{subfigure}
    \begin{subfigure}[b]{0.225\textwidth}
        \includegraphics[width=\linewidth]{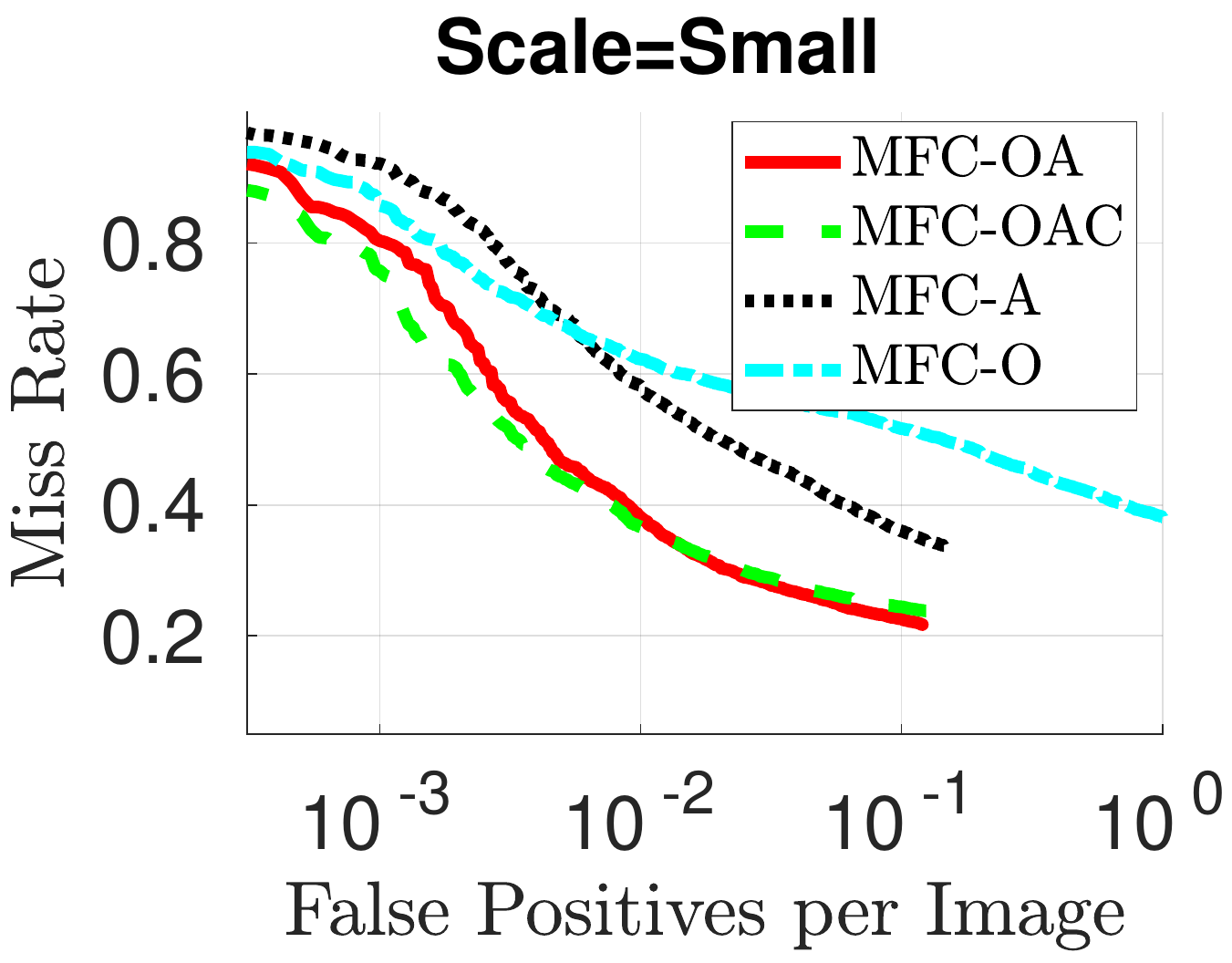}
        \caption{}\label{fig:varyTrain-small}
    \end{subfigure}
    \caption{Performance while varying training set. \revised{Each plot shows the performance of \ours{} using different training sets (including Orange, Apple, and/or Caltech data), evaluated on either the full \Baseline{} set or relevant subsets. The small scale subset is highlighted because of the substantial data representation at that scale in the Caltech dataset.}}
    \label{fig:varyTrain}
\end{figure}

In some experiments, we vary the training data used, in which case we denote the trained detector as $algorithm^{trainingData}$.
Four flavors of our approach are shown, varying the training data used:
\begin{itemize}
  \item \oursO{}: Orange data only
  \item \oursA{}: Apple data only
  \item $\ours{}^{OA}$ / \oursOA{}: Orange and apple data (full dataset) This is the ``standard'' corpus for evaluation.
  \item \textbf{\oursOAC{}}: Orange and apple (full dataset), plus Caltech dataset
\end{itemize}
The ADR for these experiments is shown in the right half of Table~\ref{tab:results}.

Methods with more data have a natural advantage, and including both apple and orange data always yields better performance than training on only one or the other. This trend does not continue with the incorporation of urban data though. The extra data from Caltech incorporated into \oursOAC{} do not seem to provide useful additional information that transfers to this domain; as highlighted in Table~\ref{tab:results}, \oursOA{} has the best performance in all categories except Scale=Small, always slightly better than \oursOAC{}. The distribution of scales in the Caltech data is skewed much more toward Small and away from Large (see Figure~\revised{\ref{fig:subsets}}), leading to some apparent biasing of the detector. Full ROCs for these domain generalization tests are in Figure~\ref{fig:varyTrain}.

We also evaluate our approach on the Caltech dataset~\cite{Dollar2012PAMI:Caltech} to see how it does on urban pedestrian detection. 
Figure~\ref{fig:caltech-full} shows results using the standard benchmark evaluation, which considers bounding boxes for the full extent of the person. We augment training set by 10x through sampling at 3fps and resize an image so that its shorter side has 720 pixels following \cite{zhang2016faster}. We also adopt the same 9 canonical shapes with a fixed aspect ratio of 0.41 as \cite{zhang2016faster} design's ``anchor boxes". We use an overlap (IoU) threshold of 50\% for deciding positive and negative locations on the ground truth heatmap and train with a batch size of 80, with 20 on each of 4 GPUs (Titan X). Other details remain the same as described in Section~\ref{sec:alg}. \ours{} shows performance competitive with the state of the art.
Figure~\ref{fig:caltech-vis} switches to the evaluation method used in this work, which uses only the visible portion of the person, to allow incorporation of our labeled data, and shows the difference between training on the Caltech data alone (\oursC{}) and also including orange and apple data in training (\oursOAC{}).
Interestingly, the addition of agricultural data seems to help performance on urban person detection, but not vice versa. This result further reinforces the importance of investigating this and other off-road domains.

\section{Discussion and Future Work}

This work represents the first large-scale analysis of person detection in an agricultural or off-road domain. We introduce a dataset that is larger than others for pedestrian detection, with richer data, including video, stereo, and vehicle position information. A major question at the outset of this work was how much of the algorithms and data from the urban domain would be transferable to off-road, agricultural detection. We evaluated leading approaches from urban pedestrian detection and found that they do not perform adequately well with recommended settings, and we propose a new approach that is better suited to the domain.

To recognize small instances, we need to consider more than the foreground (a random, natural pattern may visually look similar to a person in dark clothes far away without context). Our proposed approach builds scale-specific (also shape-specific) contextualized templates, outperforming prior-art (by +28\% in ADR) on small scale. Our approach also outperforms state-of-the-art approach by 28\% in ADR on \revised{unusual} poses. Our approach builds separate templates for usual and unusual poses (due to different bounding shapes). In comparison, other approaches try to fit one detector for all poses, which is likely to neglect unusual pose, due to its low proportion of representation (5.8\% on orange, as in Figure~\revised{\ref{fig:subsets}}).

RPN+BF's performance on small scales could be due to a number of factors.   While it passes multi-resolution features from different layers to the boosted classification stage, the proposal stage still only operates on the bottom low resolution layer.  As a result, small objects may not receive a proposal for classification.  Additionally, while the final classifier receives multi-resolution features, they are scale-normalized through RoI pooling, yielding a scale-invariant classifier.  Features learned for detecting large scale objects may be different than those for smaller scales in this dataset, due to the geometry of the orchard and camera.  Lastly, RPN+BF does not explicitly handle context in their detection stage, which can be important for small scale objects.

This work also demonstrates the complexity of transferring data from one domain to another. The hypothesis that \revised{large quantities of additional training data can be helpful even if they do not come from the final target domain} holds true for different environments within agriculture. The incorporation of orange and apple data improves detections in the opposite environment, as compared to training exclusively on that environment. However, the incorporation of urban data from the Caltech dataset did not show consistent improvement. At first blush, this suggests that the domains are too different for knowledge to transfer, yet adding agricultural data does yield an improvement when evaluating on Caltech. This is especially surprising, since this classifier receives more labels from the agricultural domains than the urban domain. Additionally, all of the leading methods need to be initialized by training on ImageNet (or some other similarly large corpus), which constitutes the transfer of knowledge from very different types of imagery. 

One possible explanation is that there is a greater variety represented in the agricultural data; urban data therefore do not add much new and useful information to agricultural detection, but agricultural data force an urban detector to generalize more, allowing detection of more rare cases. It's also possible that these patterns are specific to the particular datasets and learning approaches used. Drawing from curriculum learning to successively focus on data more closely related to the final target domain (as is already done in part with the ImageNet initialization) may allow for greater transfer.

Though our proposed approach improves upon the state of the art, there is still substantial room for improvement, and we hope the dataset will spur others to focus attention in this area. There are clear opportunities to make use of additional information. 
Specifically, though we have focused on single-frame methods and single-frame evaluation protocols, it is important to stress that our dataset contains {\bf stereo videos}. This allows for exploration of other cues based on geometry and temporal reasoning, discussed briefly below.

Geometric structure should be able to help disambiguate many cases that now prove elusive, and has previously been shown to be effective for stereo-based pedestrian detection~\cite{zhao2000stereo}. The most challenging cases that remain are those with significant occlusion, so approaches that effectively model occlusion could have significant impact in the field. Geometry can provide information about ordering of objects within the scene to infer occlusion.

Temporal information can also help in many situations. We see two particular mechanisms for improvement. The first is temporal context - a putative detection that lies below a threshold in a current frame could be boosted by associating it with a high-scoring detection in a neighboring frame through tracking~\cite{gavrila2007multi}. Secondly, motion or optical flow could be used as a cue for detection, which might be particularly helpful for camouflaged pedestrians that tend to be visible only after some movement~\cite{park2013exploring}. Such constraints could be incorporated into single-frame deep networks (such as the one we propose) by making use of optical-flow input channels~\cite{simonyan2014two} or recurrent processing~\cite{ondruska2016deep}.

\section*{Appendix}
\renewcommand{\thesubsection}{\Alph{subsection}}

\subsection{Evaluation Tools}\label{sec:appEvalTools}

Evaluation code is provided along with the dataset on the project website in two parts: The first is a fork~\cite{dollarMod} of the Dollar Matlab toolbox~\cite{dollar2013piotr}, consisting of a minor modification to add additional data filtering capabilities. The second is a repository of additional Matlab functions~\cite{labelFilteringRepo} that use those capabilities to carry out the same evaluation used to generate results in this paper.

\subsection{Additional Data}\label{sec:appExtraData}

The benchmark data consist of positive and negative labeled image of people, useful for monocular, single image detection approaches. However, the full release includes other data from our platforms that can be useful to approaches incorporating more context. This includes geometric information, additional unlabeled images from the benchmark logs, and additional labeled logs outside the standard set. These additional data can be used in the benchmark, but users are encouraged to note any modified training in their submission to the public results page.

\subsubsection{Vehicle Position and Calibration}\label{sec:appCalib}

\begin{figure}[ht]
    \centering
    \includegraphics[width=0.3\textwidth]{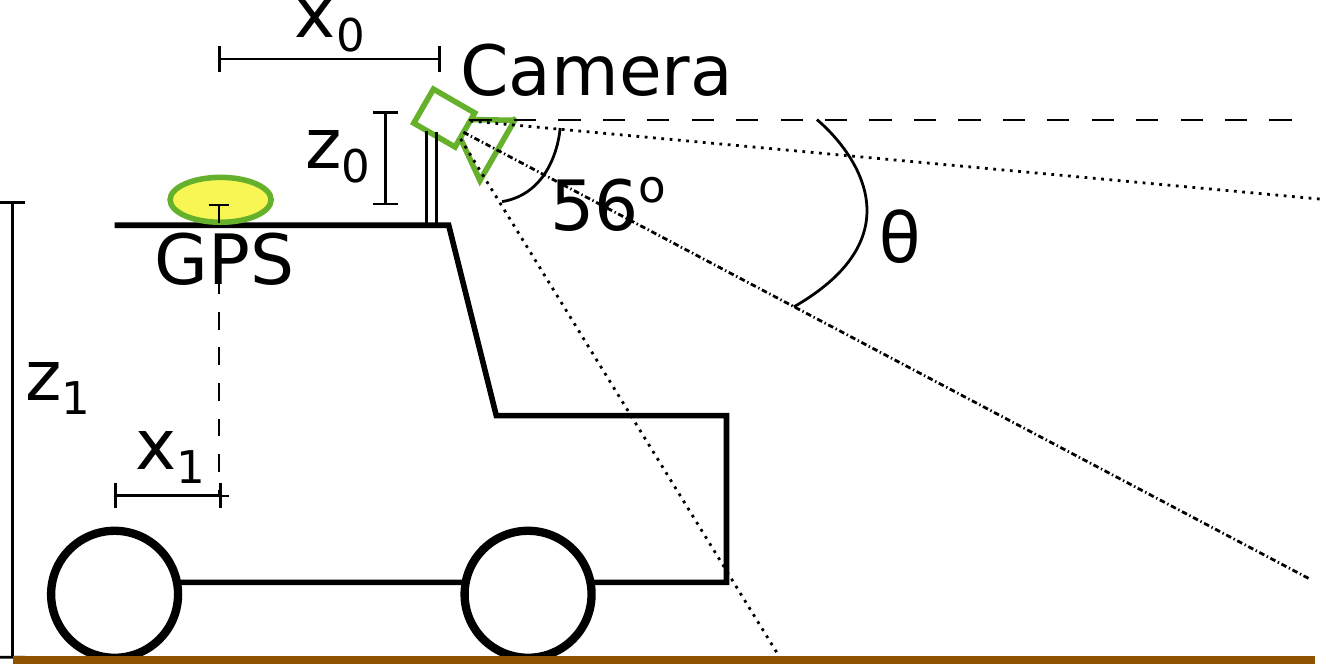}
    \caption{Diagram of camera GPS and camera position and field of view for data collection setups. See Table~\ref{tab:vehicleVals} for values on each vehicle. On both vehicles, the GPS was centered between the cameras of the 20cm-baseline stereo pair, so the lateral offset to each camera was 0.1m in either direction. $x_0$ denotes the distance from the GPS to the camera and $x_1$ the distance along the same axis from the GPS to the point above the rear axle. $z_0$ is the height of the camera above the GPS, and $z_1$ is the height of the GPS above the ground}
    \label{fig:vehicleDiagram}
\end{figure}

Calibrated vehicle position and camera intrinsics are included in the release for each sequence in the dataset to support validation of approaches that may need to model camera motion or scene geometry precisely, such as visual odometry or methods that make use of motion information or the ground plane. The camera intrinsics and rectification parameters are obtained using an industrial robotic arm for precise target positioning and performing in a joint optimization maximizing usable pixels~\cite{sturm2011camera}. The vehicle position is measured by a Starfire 3000 RTK GPS unit. For release, these measurements are interpolated and transformed to the camera frame. The mounting geometry of the GPS and camera on each vehicle is shown in Figure~\ref{fig:vehicleDiagram} and Table~\ref{tab:vehicleVals}. The translation from GPS to Camera was measured with a tape measure, and the pitch angle of the camera system was estimated from a total least squares fit of the ground plane. Other rotational components were considered close enough to zero to be negligible for purposes of detection. The vehicle position and camera information are released in KITTI \cite{geiger2013vision} odometry format for portability.

\begin{table}[th]
    \centering
    \begin{tabular}{c|ccccc}
        Environment & $x_0$ & $x_1$ & $z_0$ & $z_1$ & $\theta$ \\
        \hline
        Orange & 1.79m & 0m & 0m & 3m & $29\degree$ \\
        Apple & 1.42m & 1.05m & 0.35m & 2.15m & $22\degree$
    \end{tabular}
    \caption{Values for the quantities in Figure~\ref{fig:vehicleDiagram} for the vehicle from each environment}
    \label{tab:vehicleVals}
\end{table}

\subsubsection{Unlabeled Images}\label{sec:appUnlabeled}

The benchmark only requires the specified labeled images from the primary camera in the apple and orange orchards. This allows for new research on learning based single image detection methods. 
We also are releasing more data from the logs that may be useful for video research tasks, such as video object detection, visual odometry, mapping, and new view synthesis:

  \paragraph{Images before benchmark} Image labeling always begins at least 1 second (7 images in orange) into each log (and more if the person is not visible at the beginning). For some logs, these images can contain people, but they are not labeled or evaluated by the benchmark. Vehicle position data cover these images. These data can be used to initialize motion features or other temporal techniques or for experiments in visual odometry.
  \paragraph{Images after benchmark} Some logs continue after the person leaves view. These images are unlabeled for detection, but still include vehicle position, providing a small amount of additional data for visual odometry.
  \paragraph{Images subsampled from benchmark in Apple} When collecting the apple data, we changed platforms and system configuration. With this change, it became possible to log at a higher rate. These data are therefore released at 15 Hz, but labeled at 7.5 Hz for consistency with the orange data. The additional images that occur between the labeled images can be useful for visual odometry and other video research tasks.
  \paragraph{Images subsampled from benchmark in negative examples} All negative labeled images are randomly subsampled from logs that are known to contain no people. We are releasing the full negative logs. These logs are considerably longer than the positive logs and are likely the most useful for video research. They can also be used for computing motion features or in other video object detection frameworks.

\begin{figure}[t!]
    \centering
    \includegraphics[width=0.45\textwidth]{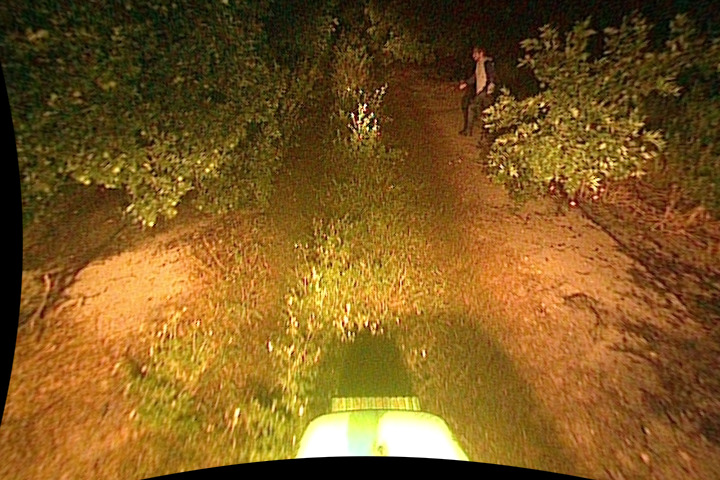}
    \caption{Example night \textbf{medium cross}.}
    \label{fig:appNight}
\end{figure}

\subsubsection{Unassigned Logs}\label{sec:appNight}
The dataset includes several logs that are labeled but that did not fit the benchmark. One orange log includes two people in view at once. In apple data, there are a few logs containing people that were not labeled. Most of the unassigned logs were orange logs collected at night under standard tractor headlights, an example of which is shown in Figure~\ref{fig:appNight}. These night logs comprise two large sets of static people and one set with a moving person, falls, and lying person. These were not considered sufficient to build three representative splits for training, validation, and final testing, but they could be used to evaluate generalization to new conditions.

\subsubsection*{Acknowledgments}
This work was supported by the USDA National Institute of Food and Agriculture as part of the National Robotics Initiative under award number 2014-67021-22171. The authors would like to thank John Deere for the use of the autonomous tractor used to collect data for this paper and Southern Gardens Citrus and Soergels Orchards for the use of their field sites. The authors would also like to thank William Drozd, and Sam Yim for assistance in this project.

\bibliographystyle{apalike}
\bibliography{jfrExampleRefs}

\end{document}